\pdfoutput=1

\documentclass[11pt]{article}
\usepackage[]{acl}

\usepackage{times}
\usepackage{latexsym}

\usepackage[T1]{fontenc}
\usepackage[utf8]{inputenc}

\usepackage{microtype}

\usepackage{inconsolata}

\usepackage{graphicx}
\usepackage{booktabs}
\usepackage{multirow}
\usepackage{color}
\usepackage{hyperref}
\usepackage{amsmath}
\usepackage{amsfonts}
\usepackage{amsthm}
\usepackage{scalefnt}
\usepackage{xspace}
\usepackage{rotating}
\usepackage{multirow}
\usepackage{listings}
\usepackage{authblk}

\theoremstyle{plain}

\def\secref#1{\S\ref{sec:#1}}
\def\seclabel#1{\label{sec:#1}}

\newcounter{notecounter}
\newcommand{\enotesoff}{\long\gdef\enote##1##2{}}
\newcommand{\enoteson}{\long\gdef\enote##1##2{{
\stepcounter{notecounter}
{\large\bf
\hspace{0cm}\arabic{notecounter} $<<<$ ##1: ##2
$>>>$\hspace{1cm}}}}}

\enoteson
\enotesoff

\title{A Comprehensive Evaluation of Multilingual Chain-of-Thought Reasoning: Performance, Consistency, and Faithfulness Across Languages}

\author[]{\bf{Raoyuan Zhao}$^{\text *}$}
\author[]{\bf{Yihong Liu}$^{\text *}$}
\author[]{{\bf Hinrich Sch\"utze}}
\author[]{\bf Michael A. Hedderich}

\affil[]{Center for Information and Language Processing, LMU Munich \\Munich Center for Machine Learning (MCML)
 \protect\\ \texttt{\{rzhao, yihong, hedderich\}@cis.lmu.de}} 

\begin{document}
\maketitle

\def\thefootnote{*}\footnotetext{Equal contribution.}\def\thefootnote{\arabic{footnote}}

\begin{abstract}

Large reasoning models (LRMs) increasingly rely on step-by-step Chain-of-Thought (CoT) reasoning to improve task performance, particularly in high-resource languages such as English. 
While recent work has examined final-answer accuracy in multilingual settings, the \emph{thinking traces} themselves, i.e., the intermediate steps that lead to the final answer,  remain underexplored.
In this paper, we present the first comprehensive study of multilingual CoT reasoning, evaluating three key dimensions: \emph{performance}, \emph{consistency}, and \emph{faithfulness}. 
We begin by measuring language compliance, answer accuracy, and answer consistency when LRMs are explicitly instructed or prompt-hacked to think in a target language, revealing strong language preferences and divergent performance across languages.
Next, we assess \emph{crosslingual consistency} of thinking traces by interchanging them between languages.
We find that the quality and effectiveness of thinking traces vary substantially depending on the prompt language.
Finally, we adapt perturbation-based techniques -- i.e., \emph{truncation} and \emph{error injection} -- to probe the \emph{faithfulness} of thinking traces across languages, showing that models rely on traces to varying degrees.
We release our code and data to support future research.\footnote{\href{https://github.com/mainlp/Multilingual-CoT-Evaluation}{https://github.com/mainlp/Multilingual-CoT-Evaluation}}



\end{abstract}

\section{Introduction}

CoT prompting has emerged as a widely adopted technique for eliciting step-by-step \emph{thinking traces} from LRMs~\citep{wei2022cot,kojima2022reasoners,zhou2023least}.  
These traces have been shown to substantially improve model performance on complex reasoning tasks, while also offering an \emph{interpretable} window for understanding the model's internal decision-making process~\citep{grattafiori2024llama3herdmodels,openai2024openaio1card,yang2025qwen3technicalreport,deepseekai2025deepseekr1incentivizingreasoningcapability,xu2025largereasoningmodelssurvey}.  


While most research on CoT reasoning has focused on English, the behavior of thinking traces in \emph{multilingual} settings remains underexplored.  
A very recent line of studies has begun to examine LRM \emph{performance} across languages, including scenarios where models are explicitly instructed or ``forced'' to reason in a specific language~\citep{yong2025crosslingualreasoningtesttimescaling,wang2025languagemixingreasoninglanguage,qi2025modelsreasonlanguagecontrolling}.  
However, these efforts largely concentrate on final-answer accuracy, leaving open critical questions about the reasoning process itself, particularly:  
(1) \emph{How \textbf{consistent} are the thinking traces across languages when answering semantically equivalent questions?}  
and  
(2) \emph{To what extent are thinking traces \textbf{faithful} in languages other than English, especially the low-resource ones?}

To address these gaps, we conduct the first comprehensive evaluation of multilingual CoT reasoning across a diverse set of LRMs.  
Our study focuses on three core dimensions: \emph{performance}, \emph{consistency}, and \emph{faithfulness}.  
In \secref{cot_performance}, we analyze language compliance, final-answer accuracy, and final-answer consistency when models are either explicitly instructed or prompt-hacked to think in a language aligned with the input.  
We find that LRMs exhibit strong language preferences during reasoning, and that performance varies substantially depending on the thinking language.
To better understand these disparities, \secref{cot_consistency} introduces a novel method for \emph{interchanging} thinking traces across languages.  
By substituting a thinking trace from one language into another, we assess whether reasoning is semantically aligned and transferable.  
Our results show that thinking traces are often inconsistent across languages, with quality varying largely by language.  
Surprisingly, we also find that final-answer accuracy is influenced not only by the thinking trace itself but also by the prompt language and thinking language.
In \secref{cot_faithfulness}, we evaluate the faithfulness of thinking traces.  
Extending prior monolingual work~\citep{lanham2023measuringfaithfulnesschainofthoughtreasoning}, we apply perturbation-based interventions -- such as truncation and error injection -- and measure how these changes impact model predictions.  
We find that models rely on their thinking traces to varying degrees across languages, suggesting that faithfulness is not uniformly preserved in multilingual contexts.

Overall, we make the following contributions:
\textbf{(i)} We present the first comprehensive evaluation of multilingual CoT reasoning, covering three core dimensions -- \emph{performance}, \emph{consistency}, and \emph{faithfulness}.
\textbf{(ii)} We propose a novel strategy: crosslingual thinking trace interchanging, to measure the semantic consistency of thinking traces across languages.
\textbf{(iii)} We find that consistency of thinking traces varies across languages, and even with identical traces, accuracy is influenced by the language of the prompt.
\textbf{(iv)} We show that languages other than English exhibit greater reliance on thinking traces, and that this reliance decreases as model scale increases.
\textbf{(v)} We will release our code to facilitate future research on the evaluation of consistency and faithfulness in multilingual reasoning.


\section{Related Work}

\paragraph{Faithfulness in CoT Reasoning}

CoT prompting~\citep{wei2022cot} has been shown to substantially improve the performance of LRMs across a variety of complex tasks~\citep{openai2024openaio1card,snell2024scalingllmtesttimecompute,muennighoff2025s1simpletesttimescaling,deepseekai2025deepseekr1incentivizingreasoningcapability}.  
Despite these gains, recent studies have raised concerns about the \emph{faithfulness} of the generated thinking traces, i.e., whether the model's stated CoT truly reflects its internal decision-making process~\citep{lyu-etal-2023-faithful,Turpin2023language,lanham2023measuringfaithfulnesschainofthoughtreasoning,tanneru2024hardnessfaithfulchainofthoughtreasoning,arcuschin2025chainofthoughtreasoningwildfaithful}.
One line of work evaluates faithfulness by introducing biases into the prompt, such as reordering multiple-choice options or injecting misleading arguments, and examining whether the model's answer changes accordingly~\citep{Turpin2023language,wang-etal-2024-answer-c,chua2025biasaugmentedconsistencytrainingreduces}.  
Another line of work manipulates the thinking trace itself, e.g., by truncating it or inserting errors, and observes how such changes affect the model's prediction~\citep{lanham2023measuringfaithfulnesschainofthoughtreasoning,yee2024dissociationfaithfulunfaithfulreasoning,xiong2025measuringfaithfulnessthinkingdrafts}.  
These studies generally reveal that models may produce a thinking trace that is disconnected from the actual decision path leading to the answer.
Our work builds on this latter line by extending it to \emph{multilingual} settings.
We manipulate thinking traces across languages, addressing the gap that existing studies evaluate faithfulness almost exclusively in English.

\paragraph{Evaluation of Multilingual Reasoning}

A growing body of work has evaluated CoT reasoning across languages~\citep{shi2023multilingual,huang-etal-2023-languages,qin-etal-2023-cross,ahuja-etal-2023-mega}, showing that CoT prompting improves performance on a variety of multilingual tasks.  
More recent studies explore how manipulating the thinking trace, such as increasing the generation budget at test time or enforcing language-specific reasoning, can further affect model performance~\citep{yong2025crosslingualreasoningtesttimescaling,wang2025languagemixingreasoninglanguage,qi2025modelsreasonlanguagecontrolling}.  
These works highlight that models often benefit from reasoning in high-resource languages like English or Chinese, or from being given more space to reason.
However, existing multilingual reasoning evaluation studies almost exclusively focus on \emph{performance}, overlooking whether models behave consistently across languages -- that is, whether they produce correct answers consistently and whether the thinking traces themselves are semantically equivalent across languages.
Such questions are especially important as LRMs are increasingly deployed in multilingual contexts~\citep{ghosh2025multilingualmindsurvey}.
Our work moves beyond performance to offer a systematic evaluation of multilingual CoT reasoning along two additional dimensions: \emph{consistency} and \emph{faithfulness}, providing a complementary perspective on how reasoning behavior generalizes across languages.

\section{Experimental Setup}

\subsection{Models}
We evaluate a wide range of open-source LRMs of different model sizes.
We consider the distilled versions of \texttt{DeepSeek-R1} \citep{deepseekai2025deepseekr1incentivizingreasoningcapability}: \texttt{DeepSeek-R1-Distill-Qwen-\{1.5B, 7B, 14B, 32B\}} whose base models are from \texttt{Qwen2.5} family \citep{qwen2025qwen25technicalreport} and \texttt{DeepSeek-R1-Distill-Llama-\{8B, 70B\}} whose base models are from \texttt{Llama3} family \citep{grattafiori2024llama3herdmodels}.
Additionally, we consider two models from \texttt{Qwen3} family \citep{yang2025qwen3technicalreport}: \texttt{Qwen3-\{8B, 32B\}}.

\subsection{Dataset}

\paragraph{MMMLU} Multilingual MMLU \citep{mmmlu} is a large-scale benchmark of general knowledge across various domains, such as  Humanities and STEM, in a multiple-choice-question format, covering 15 typologically different languages.

\paragraph{MGSM} Multilingual Grade School Math \citep{mgsm} is a benchmark that contains 250 grade-school math problems from the GSM8K \citep{cobbe2021trainingverifierssolvemath} (originally in English) that are manually translated into 10 additional languages. 

\subsection{Controlling Thinking Languages}\seclabel{control_language}

Our motivation is that the language used for reasoning should match the language of the question, as users are very likely to prefer inspecting the reasoning process in the same language they use to pose the query.
Accordingly, we consider two strategies for controlling the \emph{thinking language}, i.e., the language used in the thinking trace (text generated between the special tokens \texttt{<think>} and \texttt{</think>}), to ensure it aligns with the \emph{prompt language}, i.e., the language used in the original question.

\paragraph{Explicit Instruction}
The first strategy appends an \emph{explicit} instruction to the prompt, directly asking the model to think in a particular language.  
For example, to elicit German reasoning, we insert the phrase ``\texttt{Bitte denken Sie immer auf Deutsch.}'' [``Please always think in German.''] into the prompt. 
While intuitive, this approach seems less reliable: models may still default to their preferred thinking language, typically English, regardless of the instruction~\citep{yong2025crosslingualreasoningtesttimescaling,wang2025languagemixingreasoninglanguage}.

\paragraph{Prompt Hacking}
The second strategy uses \emph{prompt hacking}, a more targeted method to steer the model's language use~\citep{schulhoff-etal-2023-ignore,benjamin2024systematicallyanalyzingpromptinjection,qi2025modelsreasonlanguagecontrolling}.  
Here, a short \emph{prefix} in the desired language (e.g., ``\texttt{Auf Anfrage werde ich anfangen, auf Deutsch zu denken.}'' [``By Request, I will begin to think in German''] is inserted directly after the \texttt{<think>} token.  
The model is then expected to generate the remainder of the thinking trace,
until the \texttt{</think>} token.  
This approach has been shown to be more effective than explicit instructions, often leading to language-consistent CoT generation that aligns with the prefix~\citep{yong2025crosslingualreasoningtesttimescaling,qi2025modelsreasonlanguagecontrolling}.

\section{Language Compliance, Answer Accuracy, and Consistency}\seclabel{cot_performance}

\begin{table*}[]
\centering
    \setlength{\belowcaptionskip}{-0.3cm}
\resizebox{2\columnwidth}{!}{%
\begin{tabular}{clccccccccccccccc}
\hline
\textbf{Method} &
  \textbf{Model} &
  \textbf{ar} &
  \textbf{bn} &
  \textbf{de} &
  \textbf{en} &
  \textbf{es} &
  \textbf{fr} &
  \textbf{hi} &
  \textbf{id} &
  \textbf{it} &
  \textbf{ja} &
  \textbf{ko} &
  \textbf{pt} &
  \textbf{sw} &
  \textbf{yo} &
  \textbf{zh} \\ \hline
\multirow{8}{*}{\textbf{\begin{tabular}[c]{@{}c@{}}Explicit \\ Instruction\end{tabular}}} &
  R1-Qwen-1.5B &
  .25 (.07) &
  .24 (.03) &
  .35 (.24) &
  .47 (.95) &
  .32 (.89) &
  .35 (.43) &
  .28 (.05) &
  .21 (.20) &
  .30 (.51) &
  .24 (.14) &
  .32 (.02) &
  .29 (.81) &
  .19 (.40) &
  .21 (.12) &
  .37 (.85) \\
 &
  R1-Qwen-7B &
  .24 (.69) &
  .34 (.20) &
  .41 (.95) &
  .58 (.97) &
  .42 (.96) &
  .42 (.91) &
  .31 (.78) &
  .52 (.92) &
  .44 (.90) &
  .34 (.83) &
  .38 (.17) &
  .41 (.97) &
  .21 (.07) &
  .26 (.16) &
  .54 (.87) \\
 &
  R1-Qwen-14B &
  .66 (.78) &
  .56 (.02) &
  .66 (.17) &
  .76 (.97) &
  .67 (.48) &
  .62 (.17) &
  .47 (.04) &
  .67 (.17) &
  .68 (.24) &
  .65 (.32) &
  .66 (.05) &
  .71 (.12) &
  .32 (.08) &
  .31 (.09) &
  .67 (.85) \\
 &
  R1-Qwen-32B &
  .65 (.70) &
  .54 (.17) &
  .70 (.87) &
  .78 (.96) &
  .71 (.38) &
  .73 (.05) &
  .53 (.19) &
  .72 (.05) &
  .73 (.50) &
  .68 (.37) &
  .70 (.15) &
  .77 (.08) &
  .38 (.40) &
  .28 (.20) &
  .74 (.88) \\
 &
  Qwen-14B &
  .70 (.78) &
  .69 (.00) &
  .74 (.01) &
  .77 (.96) &
  .75 (.01) &
  .72 (.01) &
  .70 (.00) &
  .74 (.01) &
  .72 (.00) &
  .71 (.00) &
  .69 (.00) &
  .77 (.01) &
  .48 (.02) &
  .42 (.07) &
  .71 (.67) \\
 &
  Qwen-32B &
  .63 (.70) &
  .51 (.00) &
  .76 (.01) &
  .81 (.95) &
  .63 (.01) &
  .68 (.01) &
  .58 (.00) &
  .58 (.02) &
  .66 (.01) &
  .64 (.00) &
  .64 (.00) &
  .64 (.02) &
  .43 (.02) &
  .33 (.05) &
  .78 (.81) \\
 &
  R1-Llama-8B &
  .39 (.88) &
  .35 (.01) &
  .50 (.29) &
  .69 (.96) &
  .49 (.53) &
  .50 (.80) &
  .52 (.02) &
  .51 (.20) &
  .51 (.25) &
  .41 (.57) &
  .55 (.06) &
  .58 (.60) &
  .20 (.21) &
  .30 (.15) &
  .54 (.92) \\
 &
  R1-Llama-70B &
  .76 (.83) &
  .51 (.01) &
  .78 (.05) &
  .84 (.95) &
  .68 (.09) &
  .76 (.10) &
  .65 (.04) &
  .66 (.12) &
  .72 (.02) &
  .66 (.29) &
  .74 (.03) &
  .74 (.04) &
  .63 (.06) &
  .40 (.09) &
  .76 (.86) \\ \hline
\multirow{8}{*}{\textbf{\begin{tabular}[c]{@{}c@{}}Prompt\\  Hacking\end{tabular}}} &
  R1-Qwen-1.5B &
  .08 (.75) &
  .14 (.97) &
  .23 (.82) &
  .40 (.97) &
  .30 (.90) &
  .25 (.96) &
  .15 (.86) &
  .22 (.69) &
  .31 (.94) &
  .24 (.56) &
  .10 (.40) &
  .31 (.65) &
  .05 (.66) &
  .09 (.73) &
  .36 (.89) \\
 &
  R1-Qwen-7B &
  .29 (.66) &
  .27 (.92) &
  .37 (.95) &
  .58 (.97) &
  .48 (.96) &
  .48 (.93) &
  .29 (.84) &
  .54 (.91) &
  .42 (.96) &
  .34 (.74) &
  .23 (.61) &
  .43 (.97) &
  .04 (.76) &
  .08 (.97) &
  .55 (.91) \\
 &
  R1-Qwen-14B &
  .61 (.73) &
  .34 (.94) &
  .60 (.94) &
  .72 (.97) &
  .65 (.95) &
  .68 (.97) &
  .36 (.77) &
  .58 (.98) &
  .68 (.97) &
  .61 (.93) &
  .65 (.97) &
  .65 (.98) &
  .27 (.95) &
  .23 (.75) &
  .65 (.92) \\
 &
  R1-Qwen-32B &
  .66 (.78) &
  .49 (.91) &
  .71 (.97) &
  .80 (.96) &
  .74 (.96) &
  .75 (.97) &
  .52 (.80) &
  .73 (.98) &
  .70 (.96) &
  .67 (.73) &
  .70 (.97) &
  .73 (.98) &
  .35 (.94) &
  .25 (.94) &
  .75 (.85) \\
 &
  Qwen3-14B &
  .62 (.73) &
  .52 (.91) &
  .64 (.89) &
  .71 (.96) &
  .68 (.96) &
  .64 (.94) &
  .54 (.76) &
  .64 (.97) &
  .67 (.96) &
  .63 (.86) &
  .62 (.98) &
  .67 (.97) &
  .24 (.96) &
  .20 (.91) &
  .70 (.90) \\
 &
  Qwen3-32B &
  .62 (.78) &
  .65 (.85) &
  .67 (.59) &
  .80 (.97) &
  .55 (.42) &
  .55 (.50) &
  .73 (.67) &
  .70 (.80) &
  .68 (.40) &
  .63 (.73) &
  .71 (.86) &
  .77 (.61) &
  .40 (.91) &
  .31 (.91) &
  .74 (.88) \\
 &
  R1-Llama-8B &
  .32 (.88) &
  .25 (.81) &
  .48 (.87) &
  .69 (.94) &
  .44 (.95) &
  .54 (.94) &
  .43 (.87) &
  .40 (.98) &
  .48 (.95) &
  .33 (.89) &
  .42 (.87) &
  .51 (.95) &
  .16 (.89) &
  .20 (.85) &
  .49 (.78) \\
 &
  R1-Llama-70B &
  .70 (.83) &
  .62 (.91) &
  .76 (.81) &
  .86 (.95) &
  .79 (.78) &
  .78 (.93) &
  .75 (.73) &
  .71 (.94) &
  .74 (.88) &
  .70 (.85) &
  .74 (.77) &
  .80 (.94) &
  .62 (.90) &
  .36 (.94) &
  .76 (.87) \\ \hline
\end{tabular}%
}
\caption{Final-answer accuracy with sentence-level language compliance rates (in parentheses) for different LRMs across languages on the MMMLU task under two language-control strategies: \emph{explicit instruction} and \emph{prompt hacking}.
Results for MGSM and token-level compliance rates are provided in \secref{complete_performance}.}
\label{tab:mmmlu_performance}
\end{table*}

\begin{figure*}[htbp]
    \centering
        \setlength{\abovecaptionskip}{-0.05cm}
    \setlength{\belowcaptionskip}{-0.5cm}
    \includegraphics[width=0.235\textwidth]{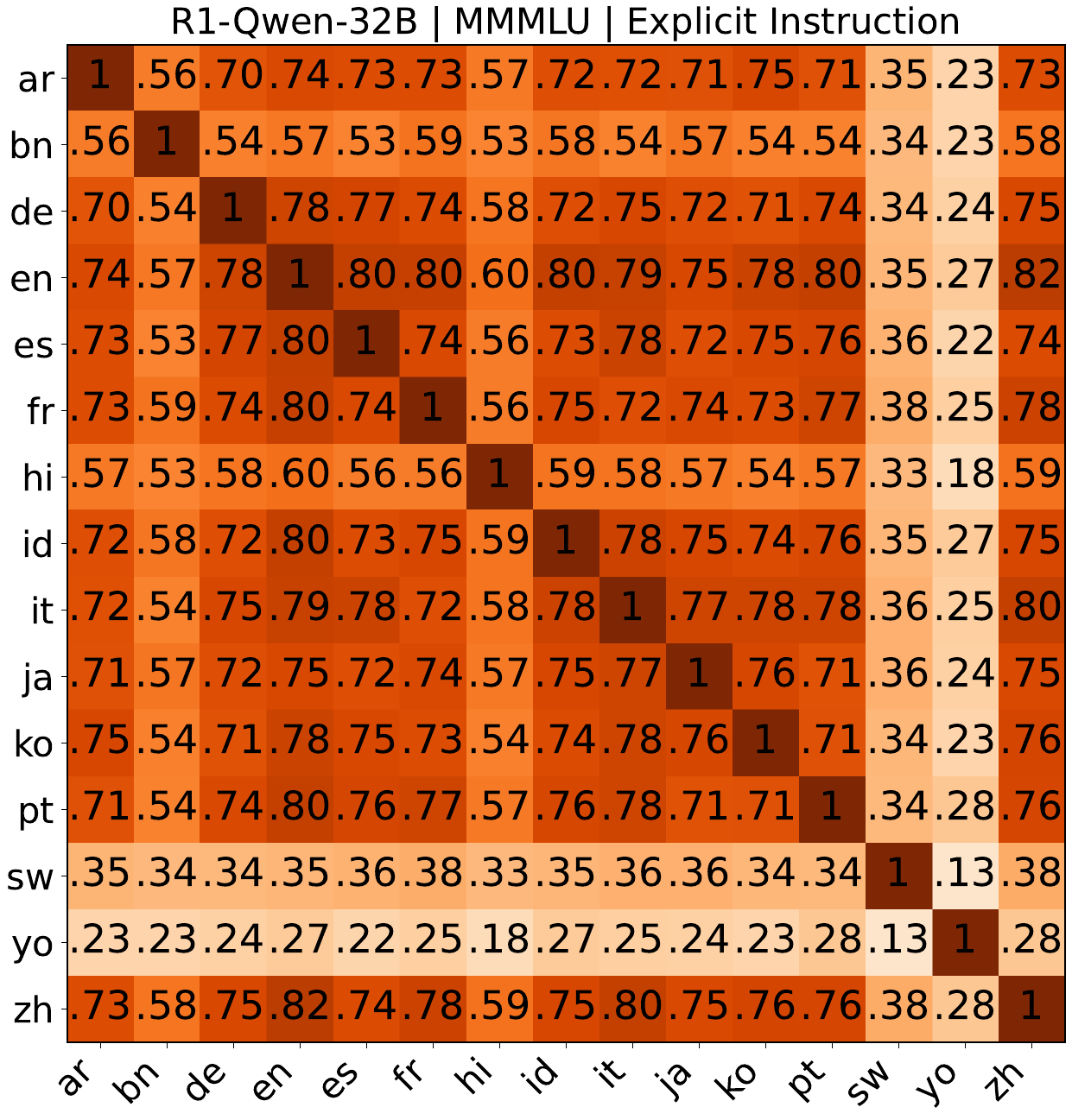}
    \includegraphics[width=0.235\textwidth]{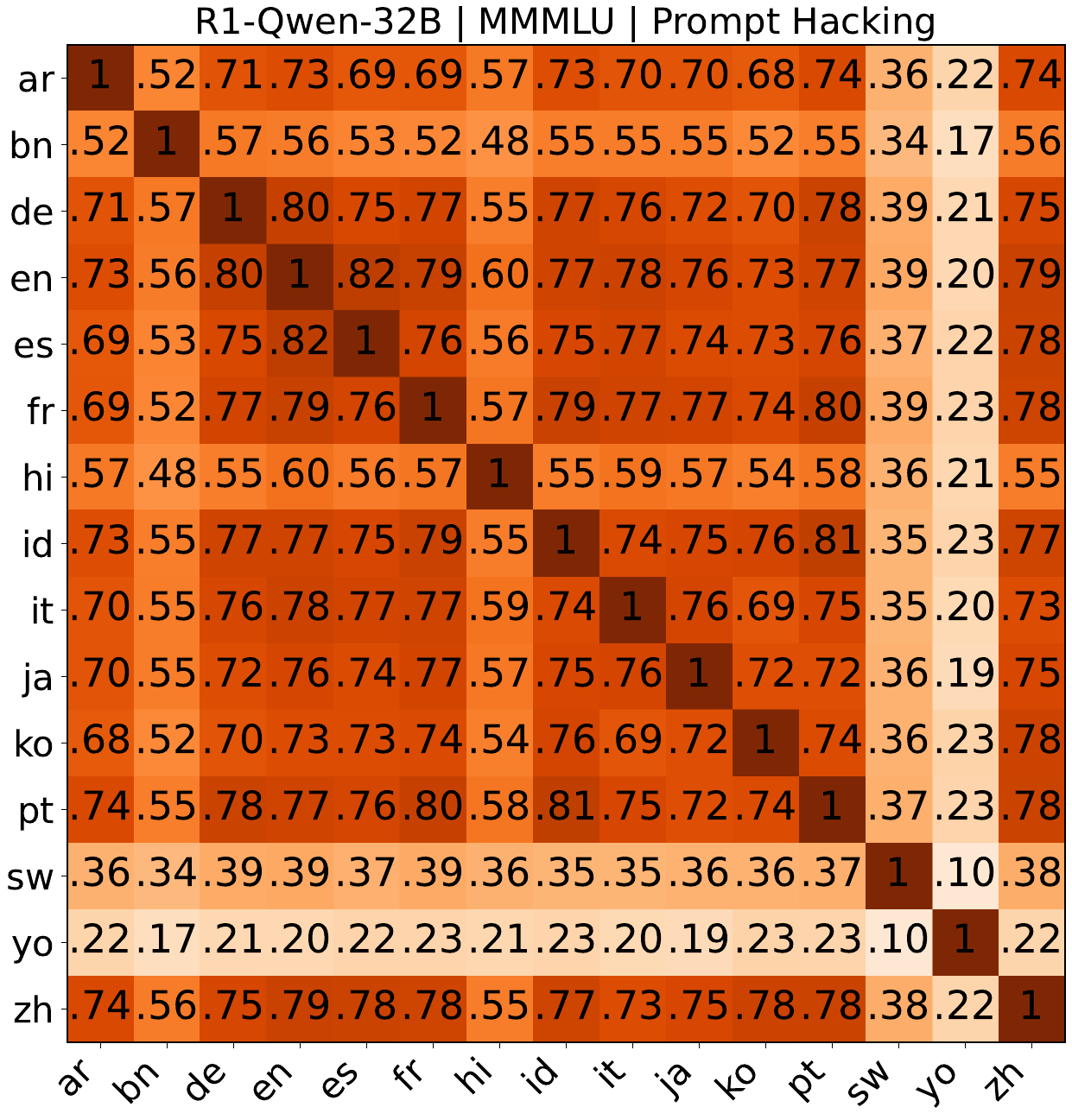}
    \includegraphics[width=0.235\textwidth]{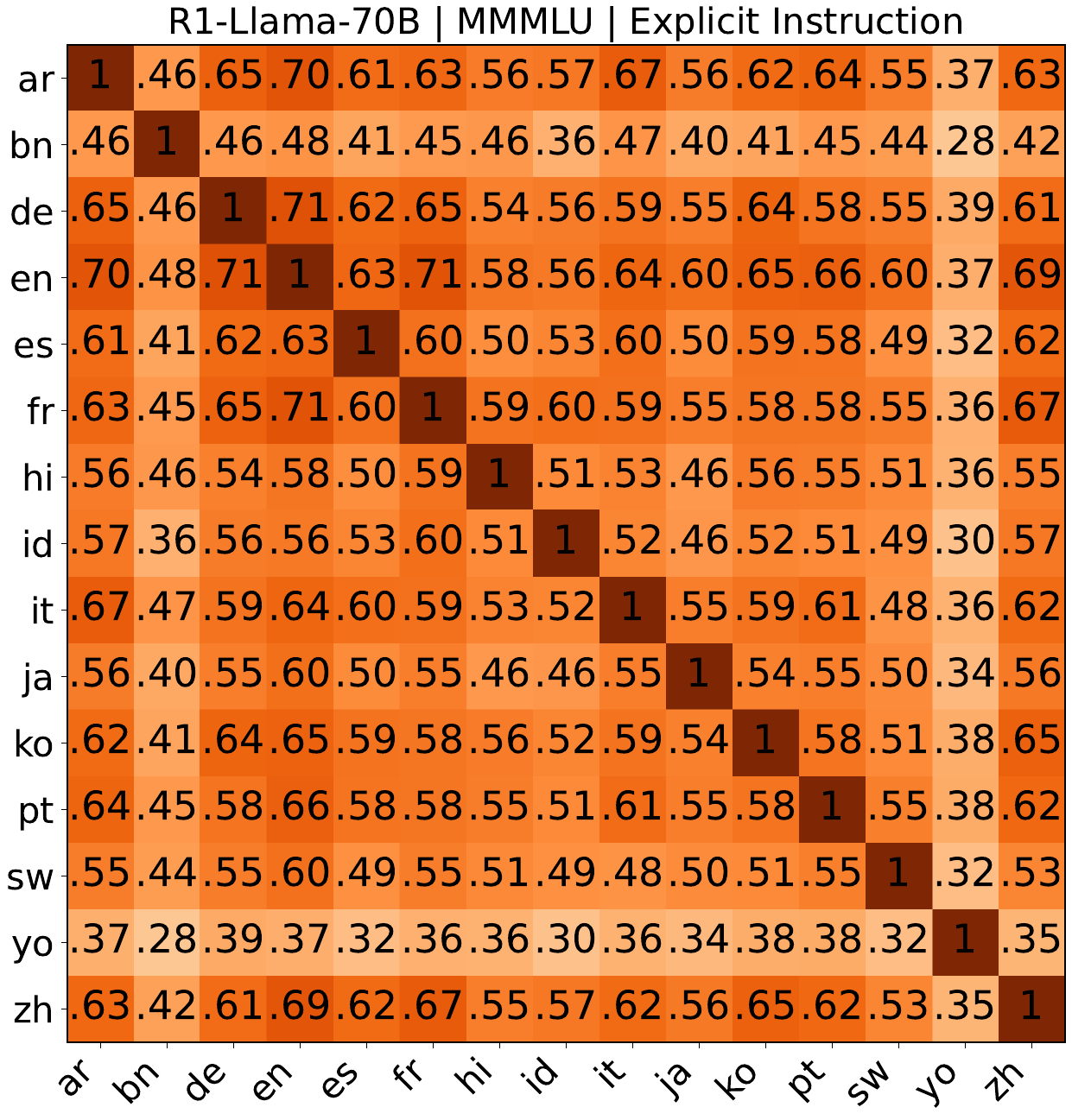}
    \includegraphics[width=0.26\textwidth]{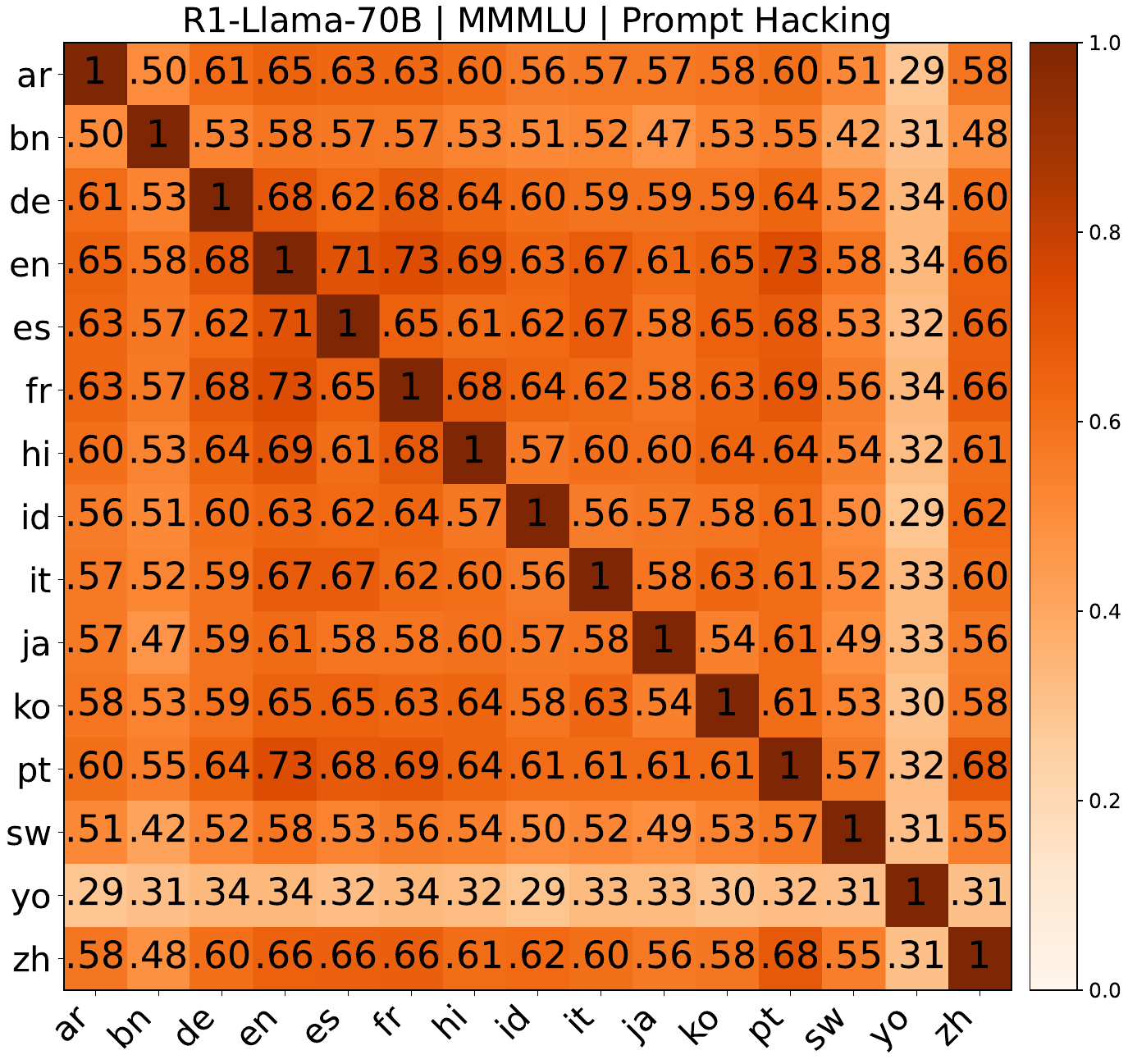}

    \caption{Final-answer consistency for R1-Qwen-32B and R1-Llama-70B under \emph{explicit instruction} and \emph{prompt hacking}.
    Similar language pairs, such as German and English, show higher consistency.
    Each cell shows the final-answer consistency between the language on the x-axis and the language on the y-axis.
    }
    \label{fig:performance_heatmaps}
\end{figure*}

In this section, we evaluate the multilingual reasoning performance of LRMs under the two language control strategies introduced in \secref{control_language}.  
Using the metrics defined in \secref{performance_metric}, we assess each model's language compliance, final-answer accuracy, and crosslingual answer consistency.  
These results, presented and discussed in \secref{performance_discussion}, allow us to examine the effectiveness of language control mechanisms and how the choice of thinking language influences model behavior.
This multi-dimensional evaluation provides a foundation for our deeper analyses in later sections, particularly regarding reasoning consistency and faithfulness across languages.

\subsection{Evaluation Metrics}\seclabel{performance_metric}


\paragraph{Language Compliance Rate}  
This metric measures the proportion of text within the thinking trace -- i.e., between the special tokens \texttt{<think>} and \texttt{</think>} -- that is generated in the intended target language (the prompt language).
To compute this, we first split each thinking trace into individual sentences and then identify the language of each sentence using GlotLID~\citep{kargaran-etal-2023-glotlid}.  
We then compute the overall proportion of reasoning content generated in the corresponding prompt language, following prior work~\citep{yong2025crosslingualreasoningtesttimescaling,wang2025languagemixingreasoninglanguage,qi2025modelsreasonlanguagecontrolling}.\footnote{In addition, we report the language usage distributions for English and Chinese across prompts in other languages, as well as token-level language distributions, in \secref{complete_performance}.}

\paragraph{Final Answer Accuracy}

This metric evaluates the correctness of the model's final prediction:
$
\mathrm{ACC}(l) = \frac{1}{|\mathcal{D}|} \sum_{i=1}^{|\mathcal{D}|} \mathbf{1}\left[\mathcal{M}(q_i^l) =  o_i^l) \right]
$
where $\mathcal{D}$ is the dataset, $\mathcal{M}(q_i^l)$ the model prediction, and $o_i^l$ the gold answer for question $i$.
We compute accuracy \emph{independently} for each language $l$.

\paragraph{Final Answer Consistency}

This metric quantifies the \emph{crosslingual consistency} of model predictions.  
Given the same question posed in two languages, we evaluate whether the model produces the same and correct answer in both languages:
\[
\begin{aligned}
\mathrm{CO}(l_1, l_2) 
&= \textstyle\frac{
\sum_{i=1}^{|\mathcal{D}|} \mathbf{1} [ \mathcal{M}(q_i^{l_1}) = o_i^{l_1} \land \mathcal{M}(q_i^{l_2}) = o_i^{l_2}]
}
{
\sum_{i=1}^{|\mathcal{D}|} \mathbf{1} [ \mathcal{M}(q_i^{l_1}) = o_i^{l_1} \lor \mathcal{M}(q_i^{l_2}) = o_i^{l_2}]
}
\end{aligned}
\]
Consistency is widely used as a metric in knowledge probing, factual knowledge recall, and cultural awareness evaluation
\citep{jiang-etal-2020-x,wang2025lostmultilingualitydissectingcrosslingual,zhao2025we,liu2025tracingmultilingualfactualknowledge,zhao2025makievalmultilingualautomaticwikidatabased}.

\subsection{Results and Discussion}\seclabel{performance_discussion}

Table~\ref{tab:mmmlu_performance} reports the accuracy and language compliance rates of the evaluated LRMs on \textbf{MMMLU} across languages.
Figure~\ref{fig:performance_heatmaps} illustrates the consistency across languages for R1-Qwen-32B and R1-Llama-70B (see \secref{complete_performance} for additional results).

\paragraph{Enforcing target-language reasoning improves compliance but may harm performance.}  
When models are explicitly instructed to think in the same language as the prompt language, many fail to follow the instruction -- especially in lower-resource languages such as Bengali (bn) and Yoruba (yo), which show low language compliance rates (less than 0.2 across all models).\footnote{Models typically default to English reasoning. See \secref{complete_performance} for the English proportion in different thinking traces.}   
Prompt hacking, by contrast, substantially improves language compliance, leading to very high alignment between the prompt and thinking language.  
However, this improved compliance often comes at the cost of reduced final-answer accuracy.  
For instance, while all models achieve remarkable compliance when forced to reason in Yoruba via prompt hacking, their accuracy drops substantially compared to explicit instruction.  
This reveals a trade-off between language control and task performance, consistent with findings in prior work~\citep{qi2025modelsreasonlanguagecontrolling}.

\paragraph{Answer consistency reflects typological proximity across languages.}  
Under both prompt hacking and explicit instruction setup, we observe that models exhibit high answer consistency across typologically similar languages.
For instance, consistencies among Indo-European languages, e.g., English (en), German (de), and French (fr), tend to be high.
To verify this, we compute average consistency for Indo-European language pairs versus mixed pairs (i.e., one Indo-European and one non–Indo-European), and find that the former is significantly higher (cf. Table~\ref{tab:p_value_final_answer} in \secref{final_answer_consistency}).
This suggests that \emph{models reason similarly in these related languages}.
In R1-Llama-70B, though the scaling improves the answer accuracy (cf. Table~\ref{tab:mmmlu_performance}), the answer consistency is lower than that of its counterpart R1-Qwen-32B (cf. Figure~\ref{fig:performance_heatmaps}), possibly due to different underlying base models.
Nevertheless, the same trend holds: consistency is generally high among typologically similar languages.

\paragraph{Performance disparities persist across language control strategies, reflecting data exposure during training.}  
Models consistently perform better on high-resource languages such as English and Chinese, which are overrepresented in pretraining and instruction tuning.  
In contrast, low-resource languages like Swahili and Yoruba yield lower accuracy in both explicit instruction and prompt hacking setups.  
These persistent gaps raise one core research question:
\emph{Why do models show different performance when the actual thinking languages vary, even with semantically equivalent prompts?} 
We hypothesize that this effect stems from inconsistencies in the quality and semantics of the thinking traces across languages -- which we explore in \secref{cot_consistency}.

\paragraph{Summary.}  
Our analysis reveals three key patterns.  
First, models do not follow explicit instructions well,
and while prompt hacking effectively enforces target-language reasoning, it often reduces accuracy.
Second, consistency is high between
similar languages.  
Finally, substantial performance gaps persist across languages regardless of control strategy,
suggesting that inconsistencies in thinking trace quality may drive these gaps.

\section{Consistency of Thinking Traces}\seclabel{cot_consistency}

We hypothesize that the disparities in final-answer accuracy and consistency observed in \secref{cot_performance} stem primarily from the \emph{quality} and \emph{semantic inconsistency} of the generated thinking traces across languages.
To investigate this, we introduce several novel substitution methods to evaluate the consistency of thinking traces between languages (\secref{interchanging}).
We further propose a new metric, \emph{substitution consistency}, which quantifies how model predictions change before and after thinking trace substitution (\secref{substitution_consistency}).
We then present and interpret our findings in \secref{consistency_discussion}.

\subsection{Thinking Trace Interchanging}\seclabel{interchanging}

To better understand the disparities in thinking traces across languages, we consider three crosslingual substitution methods, each revealing different aspects of multilingual reasoning behavior.
For the substitution, we reuse the thinking traces obtained for different languages in \secref{cot_performance} and ask the model to directly generate the final answer based on the prompts and substituted thinking traces.

\paragraph{\texttt{BaseSub}}
\begin{figure*}[ht]
    \centering
    \setlength{\abovecaptionskip}{-0.05cm}
    \setlength{\belowcaptionskip}{-0.2cm}
    \includegraphics[width=0.25\textwidth]{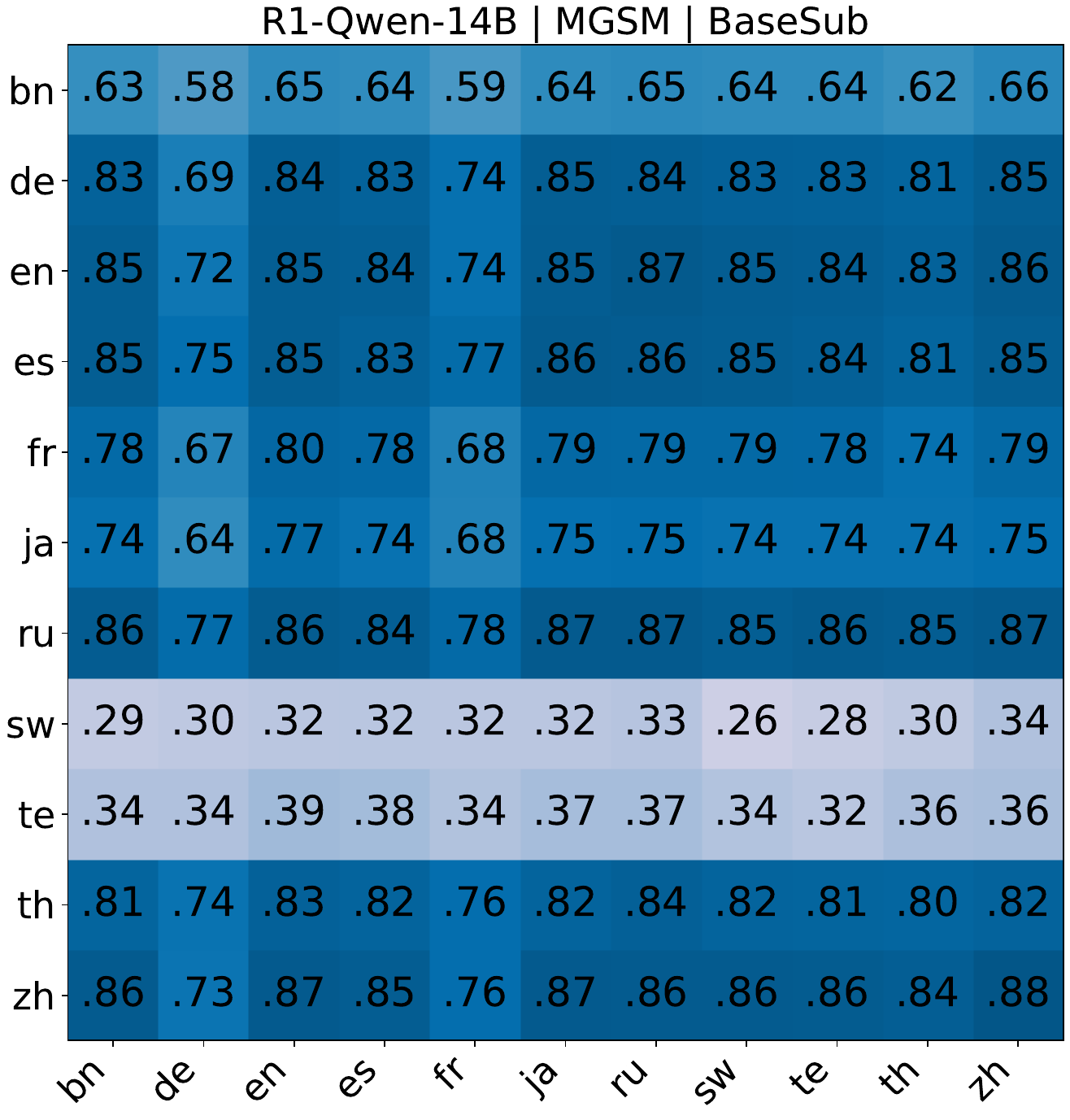}
    \hspace{0.42cm}
    \includegraphics[width=0.25\textwidth]{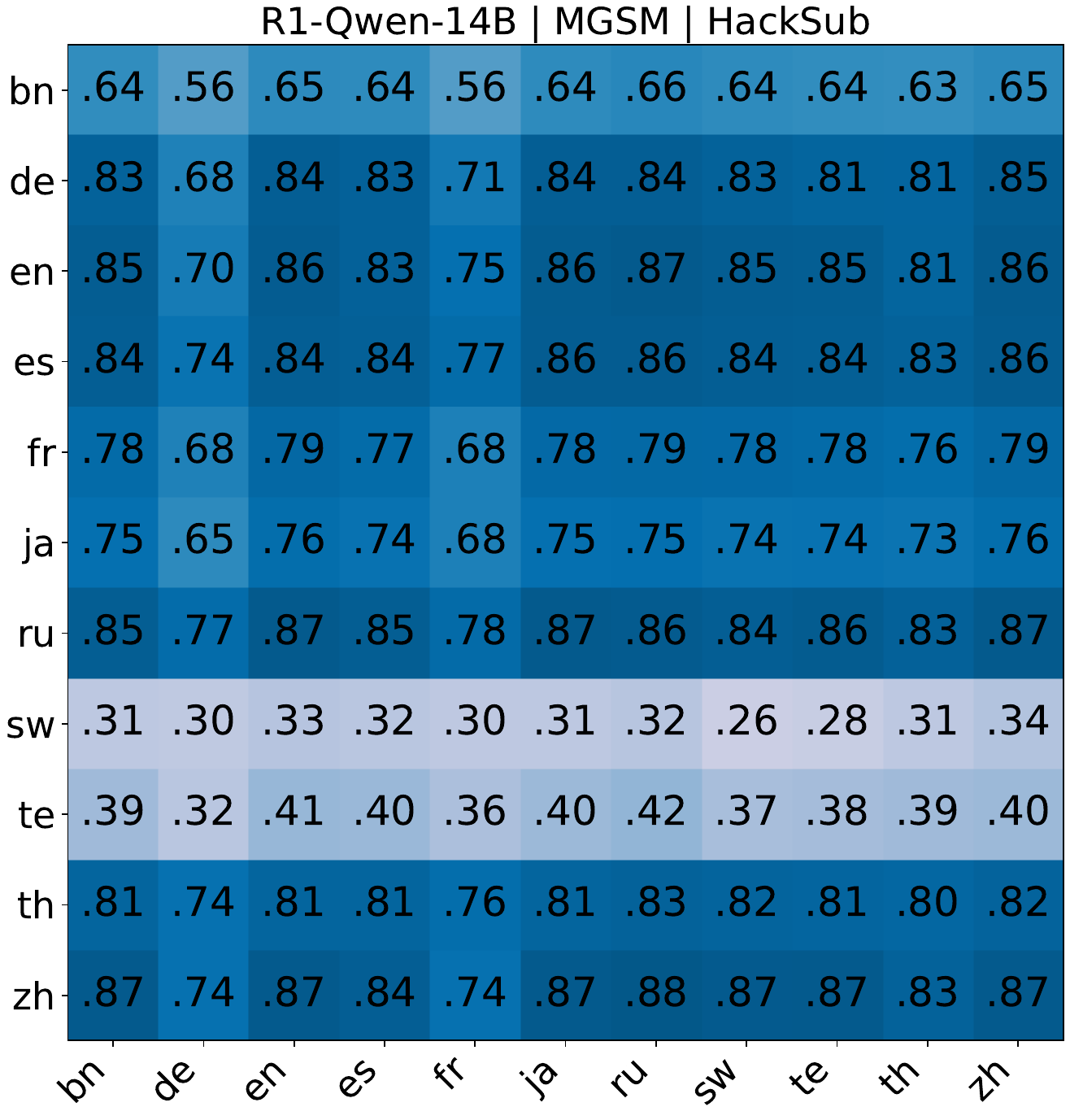}
    \hspace{0.42cm}
    \includegraphics[width=0.277\textwidth]{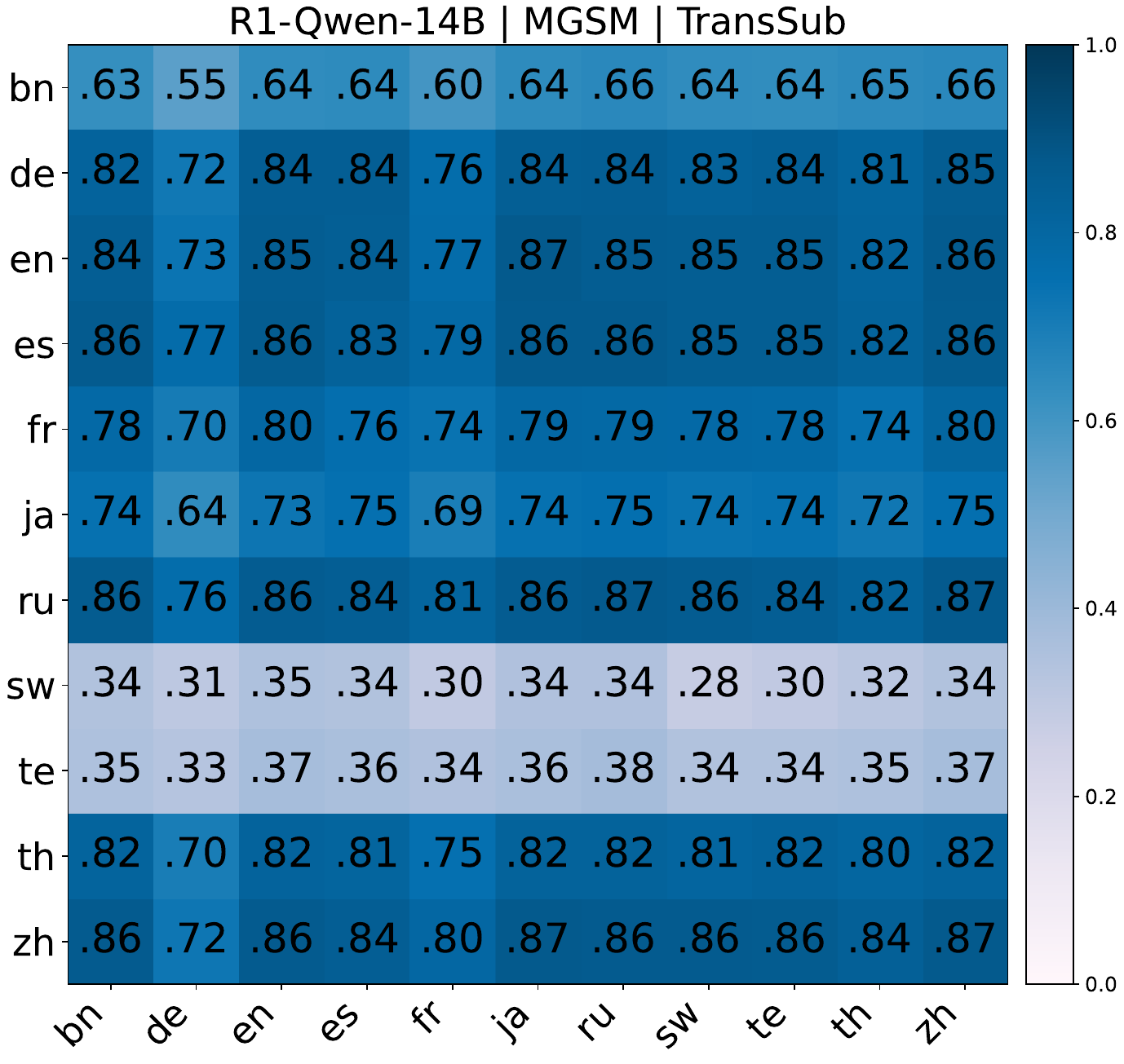}
    \caption{Final-answer accuracy of R1-Qwen-14B model under three thinking trace substitutions: \texttt{BaseSub}, \texttt{HackSub}, and \texttt{TransSub}.
    Each cell shows the accuracy when injecting thinking traces from a language on the y-axis into a language on the y-axis.
    Performance disparities indicate that thinking trace quality varies across languages.
    }
    \label{fig:sub_acc}
\end{figure*}

\begin{figure*}[ht]
    \centering
    \setlength{\abovecaptionskip}{-0.05cm}
    \setlength{\belowcaptionskip}{-0.5cm}
    \includegraphics[width=0.25\textwidth]{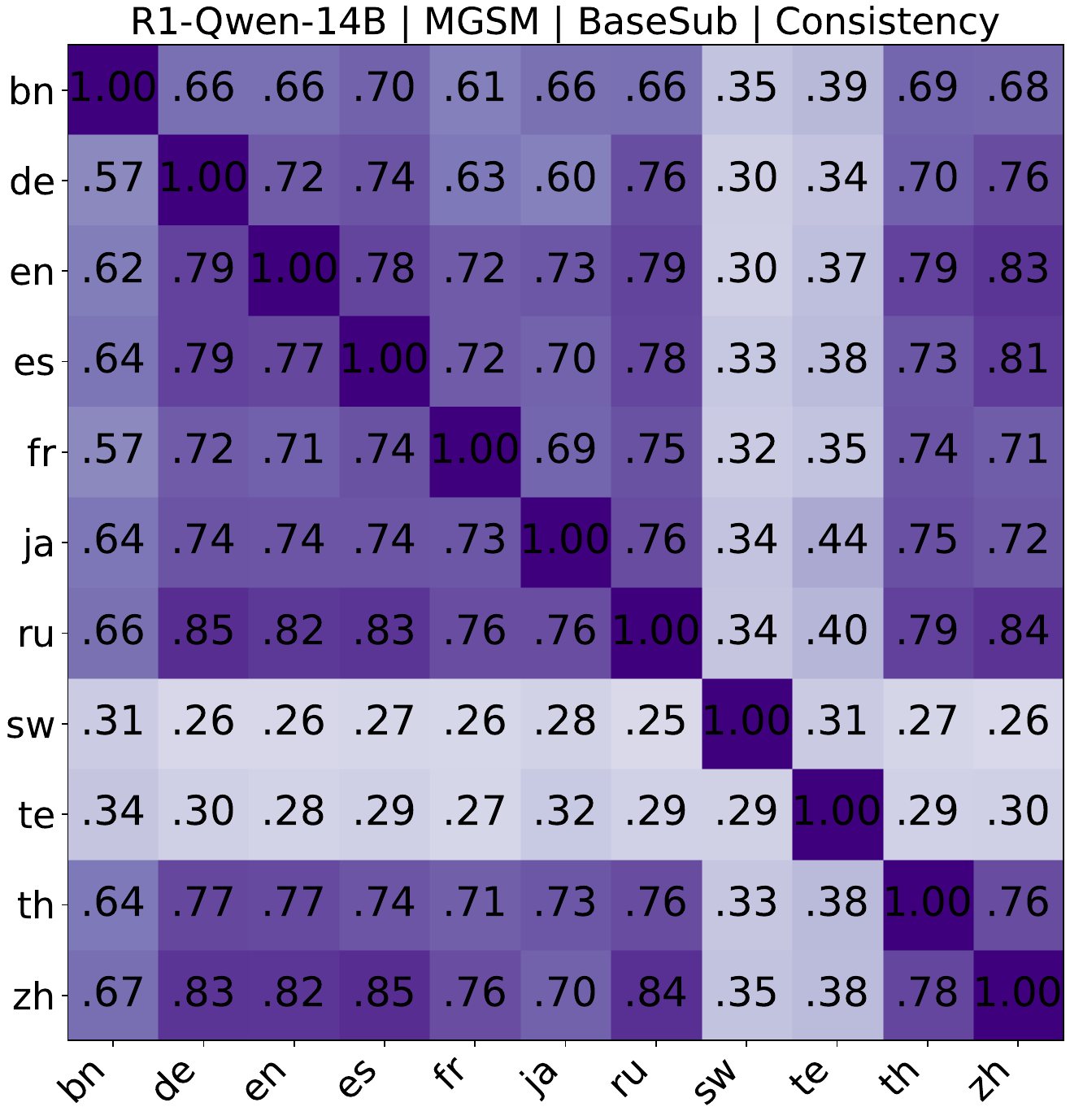}
    \hspace{0.42cm}
    \includegraphics[width=0.25\textwidth]{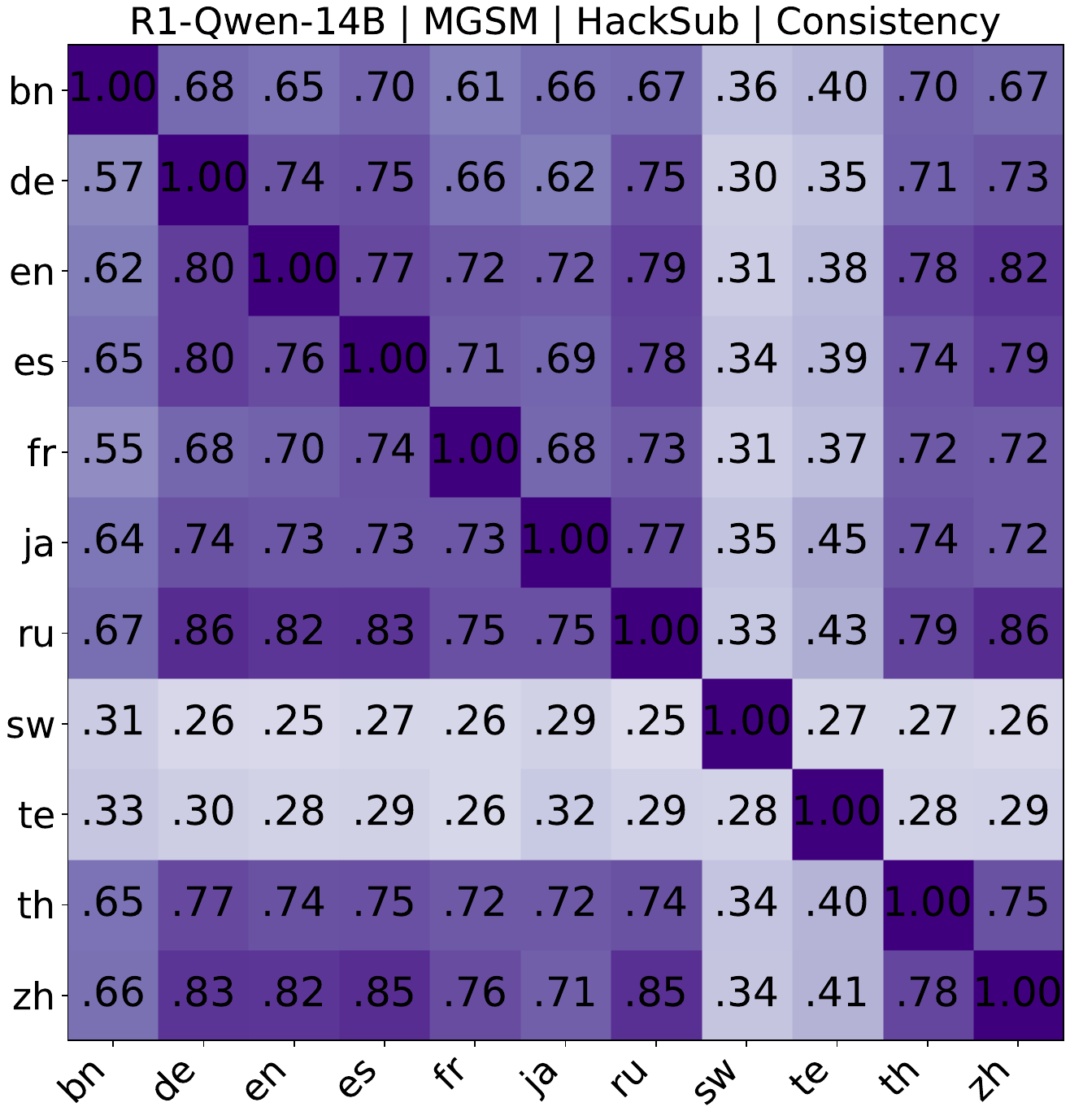}
    \hspace{0.42cm}
    \includegraphics[width=0.277\textwidth]{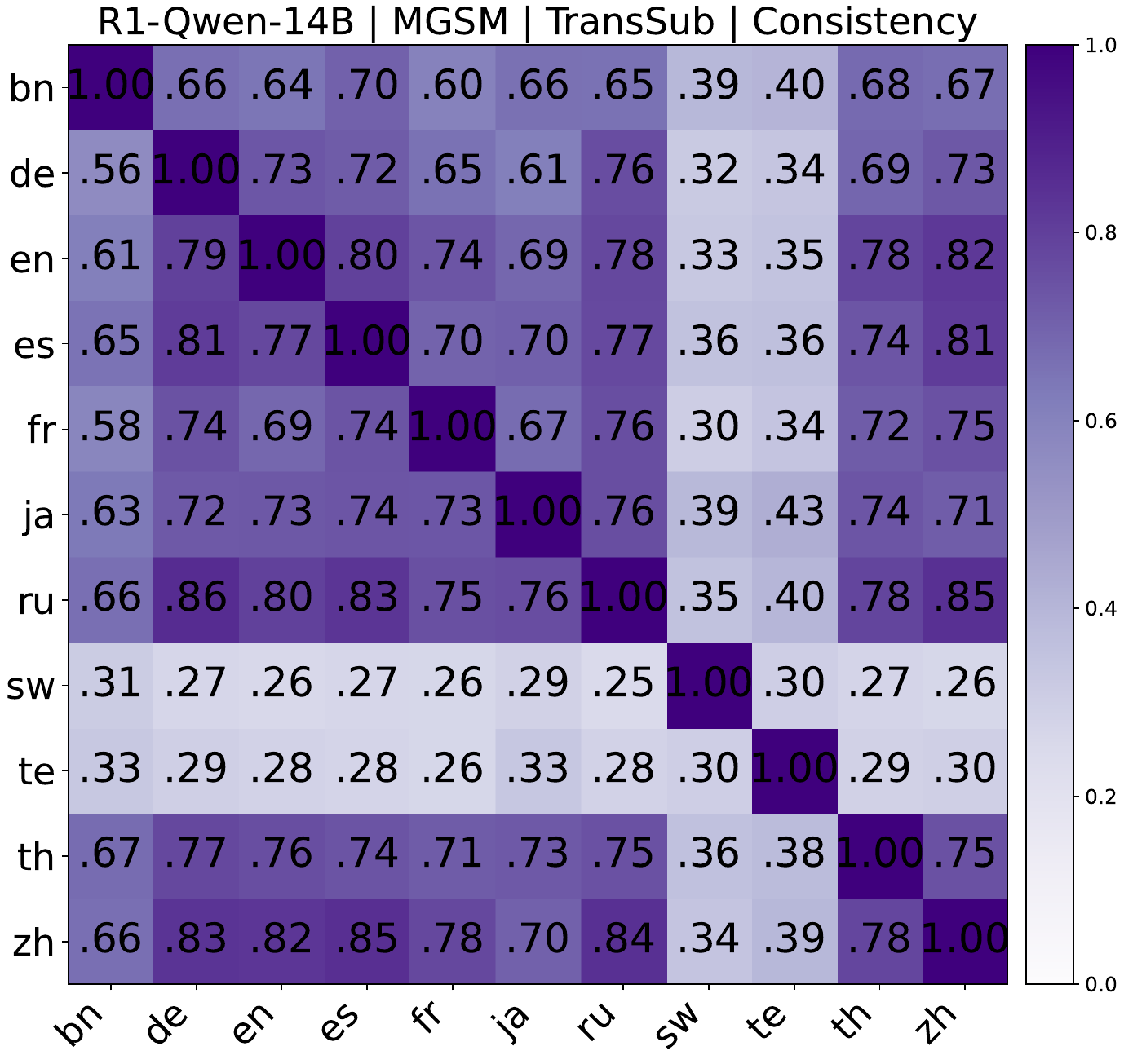}
    \caption{Substitution consistency of R1-Qwen-14B model under three thinking trace substitutions: \texttt{BaseSub}, \texttt{HackSub}, and \texttt{TransSub}.
    Each cell indicates the consistency between the original predictions in the language on the x-axis and the predictions after injecting thinking traces from the language on the y-axis.
    Higher consistency is observed when traces are substituted between similar languages.
    }
    \label{fig:sub_consistency}
\end{figure*}
In this setup, we interchange the thinking traces between languages $l_1$ and $l_2$ that are generated under \emph{explicit instruction}.  
Since models often default to high-resource languages, even when instructed otherwise, many of these traces are in English regardless of the prompt language.  
This method allows us to understand why models present different performance even though the \emph{thinking language} remains roughly flexible,
but the \emph{prompt language} varies.

\paragraph{\texttt{HackSub}}  
Here, we interchange thinking traces generated under the \emph{prompt hacking} setup.  
In this case, the thinking traces typically align with the prompt language due to the strong language control enforced by hacking prefixes. 
By interchanging these language-specific thinking traces between languages $l_1$ and $l_2$, we can examine how \emph{consistent} the thinking traces are.

\paragraph{\texttt{TransSub}}  
We first translate the thinking traces obtained under the \emph{prompt hacking} setup into \textbf{English} using the Google Translate API.\footnote{\url{https://cloud.google.com/translate}}  
We then interchange the translated English traces between language pairs $l_1$ and $l_2$.  
This setup removes the confounding variable of thinking language by standardizing all thinking traces to English.  
It provides a controlled environment to assess the quality of the generated thinking traces, independent of their original thinking languages.

\subsection{Substitution Consistency}\seclabel{substitution_consistency}

Beyond the final-answer accuracy defined in \secref{performance_metric}, we introduce \emph{substitution consistency} to quantify how a model's predictions in language $l$ change after its thinking trace is substituted with one from another language $l'$.  
Formally, let $C_{l}$ denote the set of indices of questions for which the model produces a \emph{correct} prediction in language $l$ under the original thinking trace, and let $C_{l' \rightarrow l}$ denote the set of indices for which the model produces a correct prediction in $l$ after the thinking trace from $l'$ is substituted into $l$.  
We compute substitution consistency as the intersection-over-union (IoU) between these two sets:
$
CO(l', l) = 
\frac{|C_{l} \cap C_{l' \rightarrow l}|}{|C_{l} \cup C_{l' \rightarrow l}|}
$.
Intuitively, $CO(l', l)$ measures how stable the model's correct predictions in $l$ remain after replacing its thinking trace with one from $l'$.  
Note that this metric is not symmetric -- i.e., $CO(l', l) \neq CO(l, l')$ -- because it specifically evaluates how the predictions in $l$ change when thinking traces from another language are introduced.

\subsection{Results and Discussion}\seclabel{consistency_discussion}

Figure~\ref{fig:sub_acc} and Figure~\ref{fig:sub_consistency} show the \emph{final-answer accuracy} and \emph{substitution consistency}, respectively, of R1-Qwen-14B under the three substitution strategies introduced in \secref{interchanging} for \textbf{MGSM}, where thinking traces are interchanged across languages.

\paragraph{Interchanging thinking traces substantially affects performance, revealing quality disparities across languages.}  
We find that substituting thinking traces from one language into another often leads to large performance shifts.  
Low-resource languages generally benefit from substitution with high-resource thinking traces, while high-resource languages tend to suffer performance degradation when traces from low-resource languages are injected.  
For example, under the \texttt{HackSub} strategy, the accuracy of Chinese (zh) drops to 0.40 when thinking traces from Telugu (te) are used.  
Conversely, Telugu's accuracy rises to 0.87 when using traces from Chinese.
This pattern is consistent across all three substitution strategies, suggesting that the quality of thinking traces varies dramatically across languages.

\paragraph{Substitution consistency is high between typologically-similar or resource-rich language pairs.}
We find that interchanging thinking traces between typologically-similar languages -- such as English and German -- yields relatively high substitution consistency.
In contrast, substitution between more distant language pairs (e.g., Bengali and French) results in lower consistency (e.g., 0.61 in \texttt{BaseSub}).
We further verify this by comparing the substitution consistency among different language pairs in Table~\ref{tab:p_value_swap} in \secref{complete_consistency}.
We also observe that language pairs where both languages are high-resource -- i.e., well-represented in the model's pretraining data -- tend to exhibit higher consistency.
These findings suggest that the semantic consistency of thinking traces is easier to preserve when languages share language/geographic similarity or strong pretraining exposure.


\paragraph{Thinking traces alone do not fully determine final-answer accuracy.}  
While thinking trace quality plays a major role in performance, it is not the only factor.  
We observe cases where models perform better in high-resource languages even when using identical thinking traces from low-resource languages.  
For example, when Swahili traces are injected into English prompts, the model achieves 0.33 accuracy -- higher than Swahili's own original accuracy of 0.26 in \texttt{HackSub} (cf. Figure~\ref{fig:sub_acc}).  
This indicates that the \emph{prompt language} also influences performance.

\paragraph{Models sometimes leverage English thinking traces better, even when semantically equivalent.}  
An interesting pattern emerges when comparing \texttt{HackSub} and \texttt{TransSub} for Swahili accuracy (cf. Figure~\ref{fig:sub_acc}).  
When Swahili or Telugu thinking traces are first translated into English and then injected into other languages, the model usually achieves higher accuracy than when using the original Swahili or Telugu traces directly. 
This suggests that models are better at utilizing thinking traces expressed in English, even when the underlying semantics remain unchanged. 
This bias toward English may stem from both pretraining exposure and instruction tuning in English-heavy corpora.

\paragraph{Summary.}  
Our analysis shows that the quality of thinking traces is highly uneven across languages, possibly shaped by both resource availability and inherent model biases.  
Semantic consistency of thinking traces across languages is also suboptimal, indicating that models do not generate equally aligned reasoning
in different languages.  
Lastly, our results highlight that final-answer accuracy is jointly influenced by the \emph{prompt language}, \emph{thinking language}, and the \emph{thinking trace}.

\section{Faithfulness of Thinking Traces}\seclabel{cot_faithfulness}

\begin{table*}[!ht]
\centering
\setlength{\belowcaptionskip}{-0.4cm}
\resizebox{2\columnwidth}{!}{
\begin{tabular}{cl c c c c c c c c c c c c}
\toprule
\textbf{Operation} & \textbf{Model} & \textbf{de} & \textbf{en} & \textbf{es} & \textbf{fr} & \textbf{ja} & \textbf{ru} & \textbf{sw} & \textbf{th} & \textbf{zh} & \textbf{bn} & \textbf{te} \\
\midrule
\multirow{8}{*}{\shortstack{Truncation \\ (Last)}} 
&Qwen-14B     & .68 (.75) & .75 (.79) & .69 (.79) & .64 (.77) & .69 (.82) & .68 (.75) & .40 (.84) & .65 (.72) & .67 (.79) & .66 (.83) & .63 (.87) \\
&Qwen-32B     & .66 (.83) & .76 (.86) & .55 (.81) & .39 (.72) & .62 (.87) & .60 (.81) & .56 (.89) & .77 (.86) & .60 (.75) & .77 (.89) & .60 (.88) \\
&R1-Qwen-7B   & .28 (.45) & .43 (.51) & .41 (.54) & .32 (.53) & .22 (.42) & .36 (.50) & .03 (.67) & .22 (.43) & .29 (.36) & .06 (.12) & .00 (.00) \\
&R1-Qwen-14B  & .40 (.52) & .32 (.38) & .29 (.36) & .35 (.44) & .24 (.31) & .39 (.44) & .12 (.48) & .30 (.37) & .24 (.27) & .24 (.37) & .06 (.20) \\
&R1-Qwen-32B  & .30 (.35) & .26 (.27) & .35 (.40) & .21 (.26) & .30 (.34) & .29 (.32) & .20 (.43) & .22 (.25) & .14 (.16) & .20 (.25) & -.22 (-1.22) \\
&R1-Llama-8B  & .38 (.71) & .59 (.70) & .54 (.77) & .45 (.71) & .28 (.61) & .48 (.72) & .01 (.18) & .16 (.43) & .44 (.60) & .01 (.18) & .04 (.38) \\
&R1-Llama-70B & .36 (.44) & .22 (.23) & .38 (.42) & .35 (.41) & .28 (.34) & .39 (.43) & .32 (.37) & .26 (.31) & .29 (.33) & .11 (.15) & -.09 (-.19) \\
\midrule
\multirow{8}{*}{Error Inject.}
&Qwen-14B     & .20 (.22) & .17 (.18) & .15 (.17) & .14 (.17) & .22 (.27) & .32 (.35) & .07 (.15) & .20 (.23) & .20 (.24) & .15 (.19) & .16 (.22) \\
&Qwen-32B     & .10 (.12) & .22 (.25) & .03 (.04) & -.09 (-.16) & .08 (.12) & .10 (.13) & .08 (.13) & .13 (.14) & .18 (.23) & .17 (.20) & .32 (.47) \\
&R1-Qwen-7B   & .56 (.92) & .74 (.87) & .68 (.89) & .54 (.89) & .44 (.83) & .66 (.92) & .03 (.58) & .43 (.82) & .69 (.86) & .40 (.85) & .09 (.59) \\
&R1-Qwen-14B  & .61 (.80) & .22 (.26) & .66 (.80) & .64 (.82) & .58 (.77) & .56 (.63) & .20 (.81) & .67 (.82) & .59 (.67) & .47 (.71) & .08 (.30) \\
&R1-Qwen-32B  & .57 (.65) & .16 (.16) & .60 (.68) & .60 (.73) & .62 (.72) & .57 (.63) & .33 (.73) & .67 (.75) & .64 (.71) & .56 (.70) & -.01 (-.07) \\
&R1-Llama-8B  & .47 (.88) & .71 (.84) & .62 (.89) & .55 (.88) & .38 (.82) & .53 (.80) & .03 (.73) & .27 (.72) & .63 (.87) & -.02 (-.29) & .04 (.35) \\
&R1-Llama-70B & .48 (.59) & .41 (.44) & .61 (.68) & .56 (.65) & .41 (.49) & .38 (.42) & .59 (.70) & .65 (.76) & .12 (.13) & .42 (.59) & .24 (.54) \\
\bottomrule
\end{tabular}
}
\caption{Performance absolute drop after perturbation -- \emph{last-part truncation} and \emph{error injection} -- compared to original accuracy. 
Relative drops (in percentage) are shown in parentheses.
Higher drops indicate greater sensitivity to the thinking trace perturbation, and therefore can be interpreted as stronger faithfulness.}
\label{tab:diff_last_corr}
\end{table*}

\begin{table}[!ht]
\centering
\setlength{\belowcaptionskip}{-0.4cm}
\small
\resizebox{1\columnwidth}{!}{%
\begin{tabular}{lrrrrrrrrrrr}
\toprule
Model & de & en & es & fr & ja & ru & sw & th & zh & bn & te \\
\midrule
R1-Qwen-1.5B & .61 & .46 & .51 & .58 & .49 & .53 & .28 & .16 & .69 & .38 & .22 \\
R1-Qwen-7B  & .57 & .56 & .62 & .46 & .62 & .69 & .53 & .59 & .71 & .62 & .51 \\
R1-Qwen-14B & .50 & .26 & .58 & .54 & .63 & .53 & .59 & .62 & .54 & .53 & .28 \\
R1-Qwen-32B & .43 & .12 & .51 & .46 & .57 & .54 & .60 & .61 & .59 & .56 & .40 \\
Qwen-14B    & .06 & .07 & .06 & .04 & .08 & .06 & .10 & .07 & .03 & .05 & .06 \\
Qwen-32B    & .07 & .02 & .06 & .06 & .02 & .05 & .04 & .03 & .02 & .04 & .20 \\
R1-Llama-8B & .48 & .52 & .50 & .42 & .65 & .56 & .33 & .44 & .63 & .55 & .38 \\
R1-Llama-70B & .26 &  .24 &  .41 &  .34 &  .31 &  .33 &  .53 &  .56 &  .07 &  .38 &  .49 \\
\bottomrule
\end{tabular}
}
\caption{Per-language matching ratio for each model, indicating the proportion of predictions that match the incorrect number injected into the final sentence of the thinking trace. 
Higher values suggest stronger reliance on the surface form of the reasoning.}
\label{tab:align_by_lang}
\end{table}

\begin{figure}
    \centering
    \setlength{\abovecaptionskip}{-0.01cm}
    \setlength{\belowcaptionskip}{-0.4cm}
    \includegraphics[width=0.95\columnwidth]{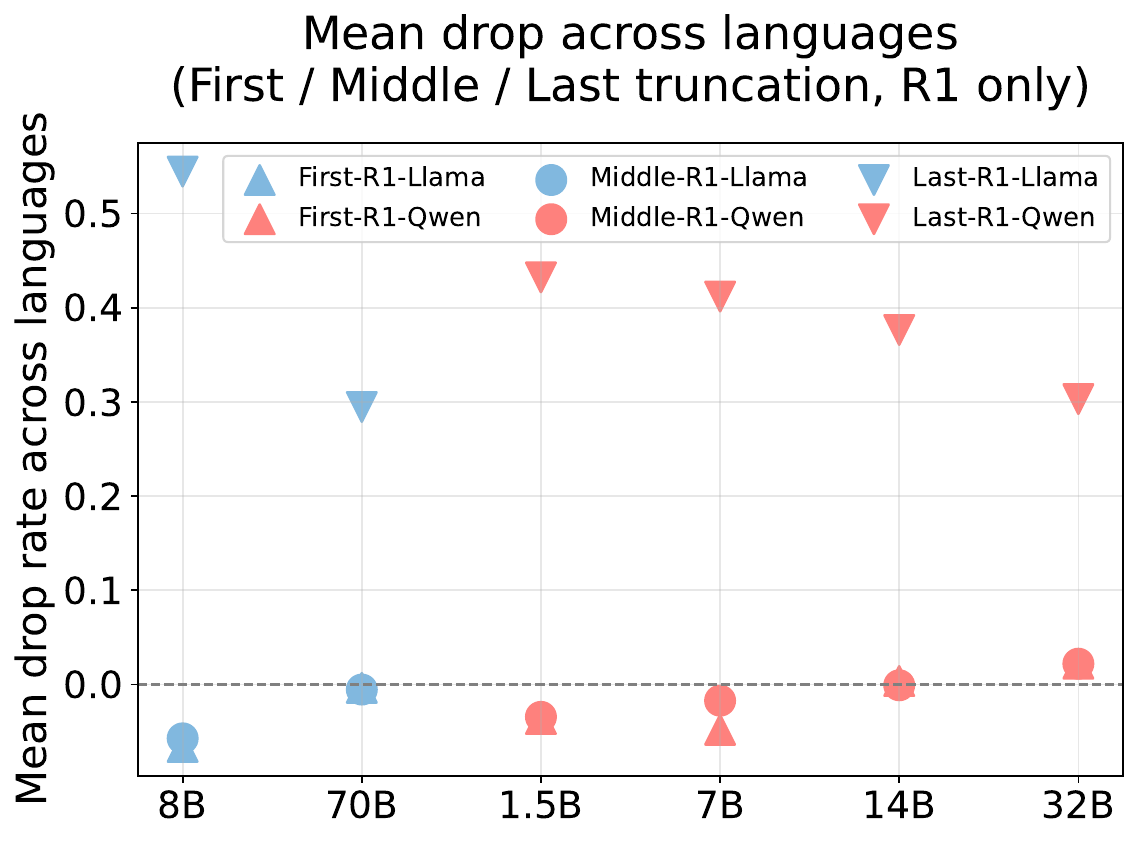}
    \caption{Mean accuracy drop (percentage) across languages for R1 distilled models under truncation of different parts of the thinking trace: first, middle, or last.}
    \label{fig:faithfulness_drop_plot}
\end{figure}

In \secref{cot_consistency}, we observed that thinking traces generated by LRMs are not consistent and not of the same quality across languages.  
In this section, we go one step further to explore the question:  
\emph{Are thinking traces faithful across languages?}  
That is, do the generated reasoning steps reflect the actual reasoning process by which the model arrives at its final answer?
Prior monolingual studies have shown that traces can be \emph{unfaithful}~\citep{lanham2023measuringfaithfulnesschainofthoughtreasoning}. 
However, whether this generalizes to other languages remains largely unexplored.
To address this gap, we perturb the thinking traces and measure how these changes affect model predictions across languages.  
We describe the perturbation strategies in \secref{faithfulness_perturbation} and present results and discussion in \secref{faithfulness_results}.

\subsection{Adding Perturbations to Thinking Traces}
\seclabel{faithfulness_perturbation}

Following \citet{lanham2023measuringfaithfulnesschainofthoughtreasoning}, we adopt two perturbation strategies to evaluate whether the model's final answer depends on its thinking traces.  
The more a model's predictions are influenced by changes to the thinking trace, the more faithfully it appears to use those traces during inference.

\paragraph{Trace Truncation}  
In this setting, we truncate the thinking trace at different points and observe how the final answer changes.  
If the model's answer remains unchanged despite the removal of reasoning steps, this suggests that the original trace may have been post-hoc or ignored during inference.  
Concretely, for each generated trace, we segment the reasoning steps into \emph{three} equal parts and perform targeted truncations: removing the \emph{first} part, the \emph{middle} part, or the \emph{last} part.  
We then compare the model's predictions under each truncated trace to those obtained with the full thinking trace.  
This setup allows us to identify not only whether truncation affects predictions but also \emph{which part} exerts the greatest influence across languages.

\paragraph{Error Injection}  
In this setting, we introduce a small error into the \emph{last sentence} of the thinking trace -- by altering a number involved in the final computation step (e.g., changing it to another number).  
This design specifically targets the final stage of reasoning, where the model is expected to derive or summarize the correct answer.
The goal is to assess whether the model relies on the correctness of the concluding reasoning step.  
If the model's answer changes in response to this minimal perturbation, it suggests that it is faithfully using its own reasoning.
On the other hand, if the answer remains unchanged, despite the final reasoning step being incorrect, this may indicate that the model is ignoring its stated trace or relying on earlier steps, memorized patterns, or even contamination instead.

\subsection{Results and Discussion}\seclabel{faithfulness_results}

Table~\ref{tab:diff_last_corr} reports the change in final-answer accuracy for each language on the \textbf{MGSM} dataset under two perturbation strategies -- \emph{last-part truncation} and \emph{error injection} -- in the \texttt{HackSub} setting.  
Figure~\ref{fig:faithfulness_drop_plot} further visualizes the effect of truncating the first, middle, and last segments of the thinking trace across models in the DeepSeek-R1 distilled series.

\paragraph{Models show varying degrees of faithfulness across languages.}  
We observe diverse sensitivity to perturbations across languages.  
For some low-resource languages like Swahili and Telugu, perturbations have little impact in R1 distilled models -- largely because original performance is already low.  
In contrast, many languages experience substantial accuracy drops, suggesting that models do rely on their thinking traces to varying extents.  
This is further supported by our matching ratio results in Table~\ref{tab:align_by_lang}, which measure how often the final predictions are influenced by the incorrect numbers injected into the trace.  
Notably, English consistently shows lower matching scores (e.g., 0.12 for R1-Qwen-32B) compared with other languages.

\paragraph{Truncating the final part is less disruptive than error injection, especially for R1 distilled models.}  
Across all languages, we find that truncating the final segment of the thinking trace has less impact than injecting an incorrect value, particularly for the R1 distilled series.\footnote{Interestingly, the Qwen3 models show the opposite trend. We hypothesize that this may be due to \emph{data contamination}, as seen in Table~\ref{tab:align_by_lang}, where these models rarely change their answers even when the final sentence is corrupted.}  
This suggests that models may perform \emph{latent-state reasoning}~\citep{zhu2025surveylatentreasoning}, where inference continues internally even after the visible thinking trace is truncated.  
However, when a plausible but incorrect final value is injected, models often copy it directly, revealing that their outputs are sensitive to surface-level reasoning conclusions.

\paragraph{Model scale affects faithfulness behavior.}  
As shown in Figure~\ref{fig:faithfulness_drop_plot}, model scale influences how different parts of the thinking trace affect predictions.  
Smaller models are more reliant on the final portion of the trace, aligning with their higher matching ratios (cf. Table~\ref{tab:align_by_lang}), suggesting a higher degree of surface-level faithfulness.  
Larger models, by contrast, become less dependent on the final segments and more sensitive to earlier reasoning steps.  
This may indicate either (1) reduced faithfulness due to memorization or contamination, or (2) stronger latent-state reasoning capabilities that allow the model to recover from truncated traces or correct surface-level trace errors.

\paragraph{Summary.}  
Our findings reveal that models vary in their faithfulness across languages and model scales.  
Languages other than English show stronger reliance on thinking traces.  
Truncating the final part of the trace is generally less disruptive than injecting incorrect information, especially for R1 distilled models, possibly suggesting latent-state reasoning.  
Larger models are less dependent on surface-level reasoning and more resilient to perturbations, though this may reflect either increased reasoning ability or memorization.

\section{Conclusion}

In this paper, we present the first comprehensive evaluation of multilingual CoT reasoning across a diverse set of LRMs.  
We examine three core dimensions, \emph{performance}, \emph{consistency}, and \emph{faithfulness}, to provide a deeper understanding of how LRMs reason across languages.
We show that LRMs exhibit strong language preferences in reasoning and that final-answer performance varies substantially across languages. 
Through our \emph{crosslingual thinking trace interchanging} method, we show that thinking traces are often inconsistent across languages, with their quality strongly associated with the thinking language.
Finally, our perturbation-based tests reveal that models rely on the traces to varying degrees, suggesting that reasoning faithfulness is uneven across languages.
Our findings highlight the need for more robust and transparent evaluation of multilingual reasoning behavior.

\section*{Limitations}

While our work provides the first comprehensive study of multilingual CoT reasoning, we acknowledge that several limitations remain.

First, although we examine robustness through two perturbation strategies and show that robustness varies across languages, more sophisticated or adversarial perturbations (e.g., paraphrasing, distractor reasoning) remain unexplored and could be incorporated in future work.

Second, while we evaluate and analyze inconsistencies in multilingual reasoning, we do not provide a mechanistic explanation for why these inconsistencies arise.  
Future research could apply mechanistic interpretability methods to investigate model internals to better understand the sources of multilingual inconsistency and faithfulness.

Finally, due to resource constraints, our experiments are limited in the number of models and downstream tasks considered.  
Future work could extend our evaluation framework to a broader set of models, languages, and tasks.

\section*{Acknowledgments}

This research was supported by the Munich Center for Machine Learning (MCML) and German Research Foundation (DFG, grant SCHU 2246/14-1).
We gratefully acknowledge additional support from Google DeepMind through a generous research grant, which enabled our use of the Google Translate API services in this project.

\section*{Ethical Considerations}

\paragraph{Use of AI Assistants}
The authors acknowledge the use of ChatGPT exclusively for grammar correction, improving the clarity and coherence of the draft, and assisting with code implementation.\footnote{\url{https://chatgpt.com/}}

\bibliography{custom}

\appendix

\section{Additional Results}\seclabel{additional_results}

\subsection{Complete Results for Language Controlling}\seclabel{complete_performance}
This section presents the complete compliance results for all languages, as well as additional results on MGSM, covering accuracy, consistency, and compliance.

\begin{table}[]
\centering
\resizebox{\columnwidth}{!}{%
\begin{tabular}{cccc}
\hline
\textbf{Metric} & \textbf{Group}    & \textbf{Mean Value} & \textbf{P-Value}     \\ \hline
\multicolumn{4}{c}{\textbf{Without Low Resource Languages}}                      \\ \hline
\multirow{2}{*}{\begin{tabular}[c]{@{}c@{}}Final-Answer\\ Consistency\end{tabular}} & Indo-European & 0.565  & \multirow{2}{*}{0.034}                        \\
                & Non Indo-European & 0.551               &                      \\ \hline
\multicolumn{4}{c}{\textbf{With Low Resource Languages}}                      \\ \hline
\multirow{2}{*}{\begin{tabular}[c]{@{}c@{}}Final-Answer\\ Consistency\end{tabular}} & Indo-European & 0.5384 & \multicolumn{1}{l}{\multirow{2}{*}{3.88e-20}} \\
                & Non Indo-European & 0.4894              & \multicolumn{1}{l}{} \\ \hline
\end{tabular}%
}
\caption{Consistency comparison between Indo-European and non-Indo-European languages. Reported are mean consistency values for Final-Answer consistency metrics, with corresponding p-values (t-test). We also discard low-resouce languages sw and yo to conduct t-test. Indo-European languages generally achieve higher consistency, and the differences are statistically significant.}
\label{tab:p_value_final_answer}
\end{table}

\begin{table*}[ht]
\centering
\begin{tabular}{llccccccccccc}
\toprule
Method & Model & de & en & es & fr & ja & ru & sw & th & zh & bn & te \\
\midrule
\multirow{8}{*}{Explicit Instruction} & Qwen-14B & .91 & .93 & .91 & .87 & .82 & .91 & .62 & .89 & .83 & .81 & .73 \\
 & Qwen-32B & .89 & .94 & .92 & .88 & .86 & .92 & .80 & .88 & .87 & .84 & .79 \\
 & R1-Qwen-1.5B & .42 & .78 & .50 & .47 & .22 & .47 & .01 & .03 & .66 & .13 & .03 \\
 & R1-Qwen-7B & .69 & .85 & .75 & .74 & .56 & .77 & .10 & .51 & .80 & .49 & .26 \\
 & R1-Qwen-14B & .84 & .93 & .89 & .86 & .80 & .93 & .29 & .85 & .87 & .72 & .40 \\
 & R1-Qwen-32B & .89 & .96 & .91 & .87 & .86 & .93 & .44 & .89 & .87 & .81 & .40 \\
 & R1-Llama-8B & .57 & .78 & .65 & .62 & .45 & .66 & .04 & .39 & .63 & .17 & .18 \\
 & R1-Llama-70B & .88 & .96 & .92 & .89 & .85 & .93 & .85 & .89 & .88 & .84 & .55 \\
\midrule
\multirow{8}{*}{Prompt Hacking} & Qwen-14B & .91 & .95 & .88 & .84 & .84 & .90 & .48 & .90 & .85 & .79 & .72 \\
 & Qwen-32B & .80 & .88 & .68 & .54 & .72 & .74 & .63 & .90 & .80 & .86 & .69 \\
 & R1-Qwen-1.5B & .40 & .79 & .47 & .47 & .26 & .44 & .00 & .01 & .65 & .06 & .04 \\
 & R1-Qwen-7B & .62 & .85 & .76 & .61 & .53 & .72 & .05 & .52 & .80 & .47 & .15 \\
 & R1-Qwen-14B & .76 & .84 & .82 & .78 & .76 & .88 & .25 & .82 & .88 & .67 & .28 \\
 & R1-Qwen-32B & .90 & .96 & .90 & .87 & .86 & .93 & .49 & .90 & .87 & .83 & .45 \\
 & R1-Llama-8B & .56 & .78 & .64 & .62 & .46 & .66 & .07 & .42 & .62 & .24 & .21 \\
 & R1-Llama-70B & .86 & .96 & .86 & .83 & .80 & .92 & .62 & .90 & .84 & .79 & .47 \\
\bottomrule
\end{tabular}
\caption{Final-answer accuracy for different LRMs
across languages on the MGSM task under two language-control strategies: explicit instruction and prompt
hacking.}
\label{tab:mgsm_acc}
\end{table*}
\subsubsection{Accuracy of Final-answer}
Table~\ref{tab:mgsm_acc} reports the final-answer accuracy of different LRMs on the MGSM task under explicit instruction and prompt hacking. The results confirm our findings: explicit instructions are often not strictly followed, and while prompt hacking improves language control, it generally comes at the cost of accuracy. In particular, accuracy drops are most pronounced in low-resource languages (e.g., Swahili, Telugu, Bengali), whereas high-resource languages (e.g., English, German, French, Chinese) show relatively stable performance across both control strategies.

\subsubsection{Consistency of Final-answer}
\begin{figure*}[htbp]
    \centering
    \includegraphics[width=0.24\textwidth]{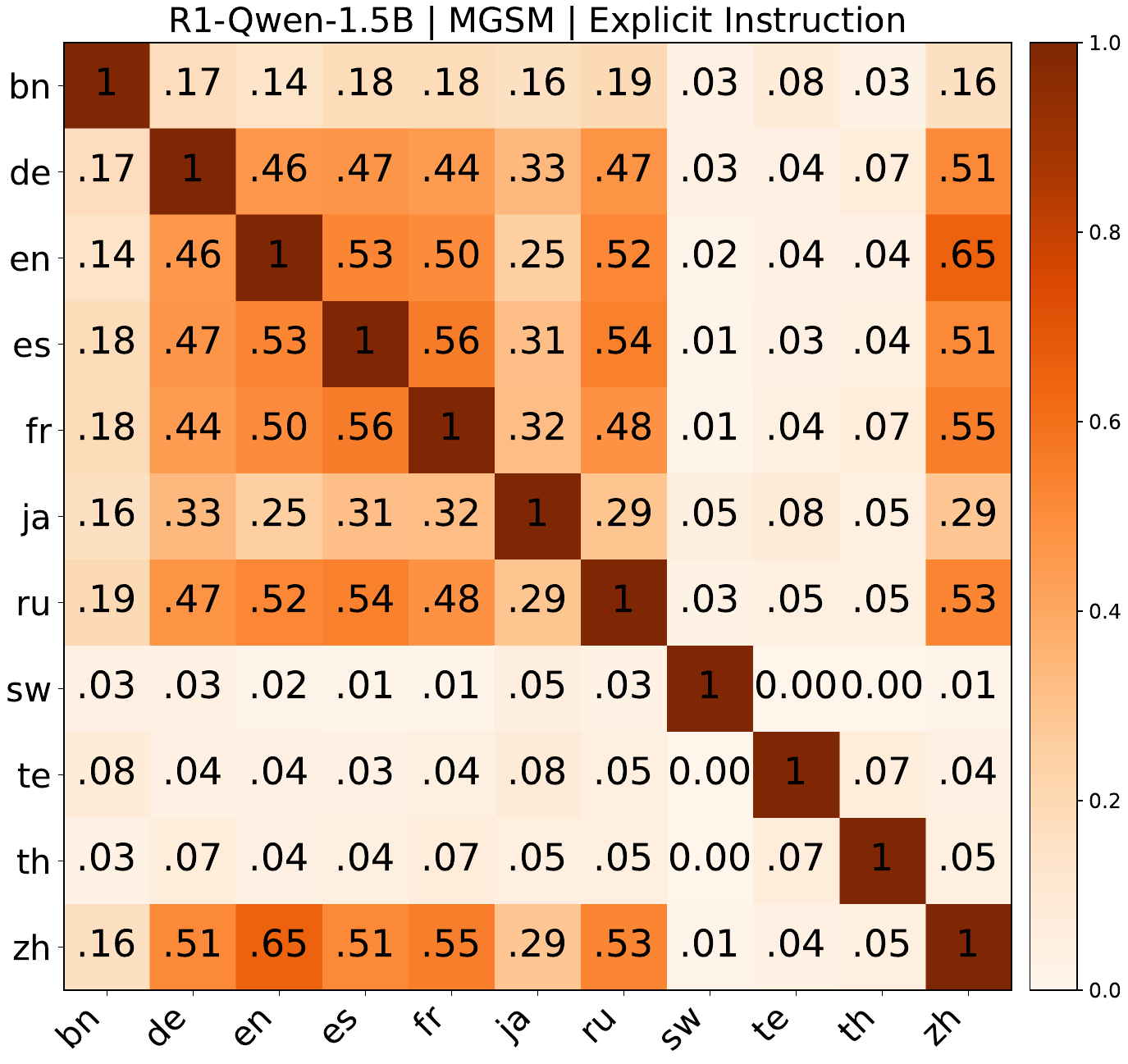}
    \includegraphics[width=0.24\textwidth]{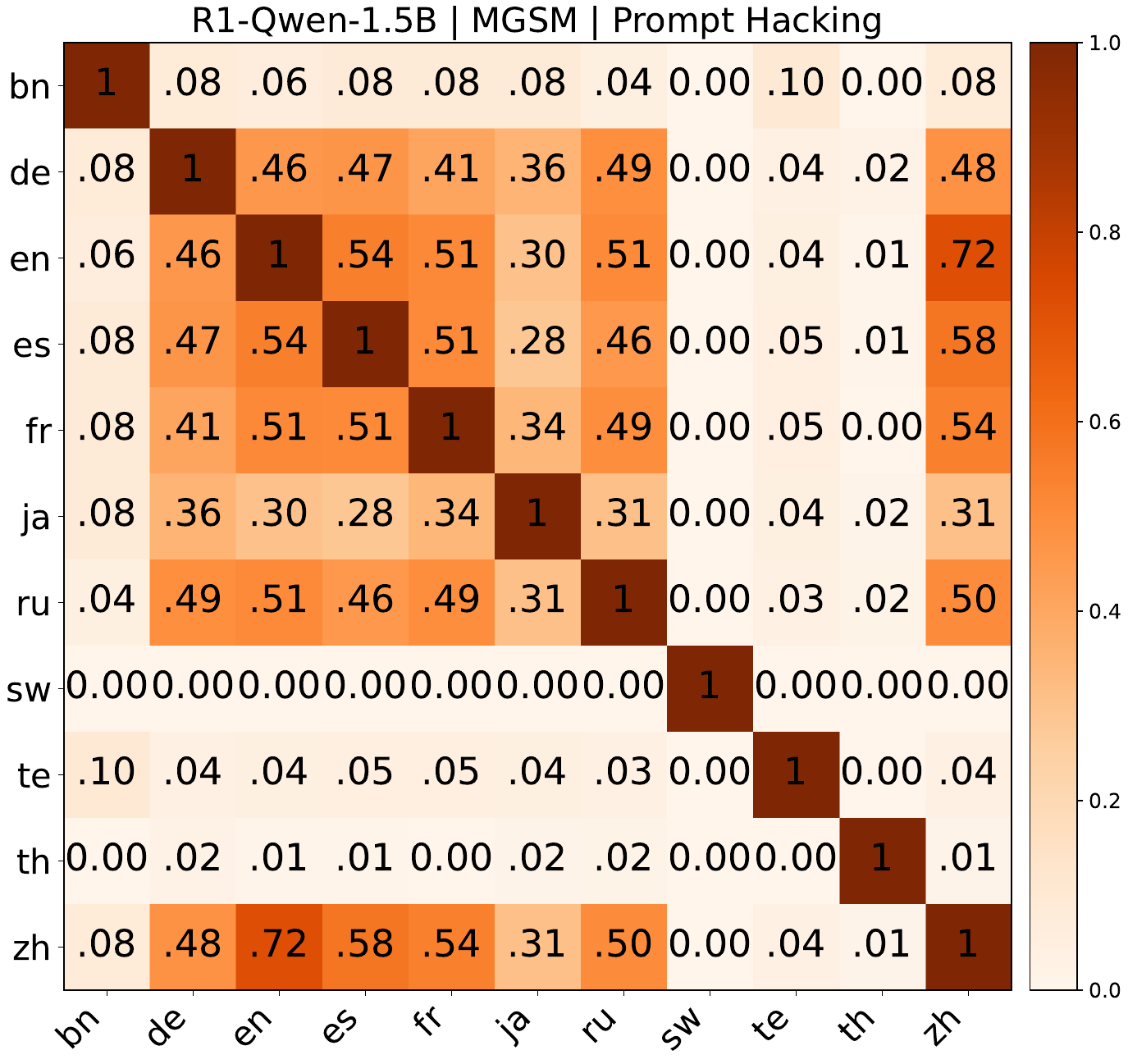}
    \includegraphics[width=0.24\textwidth]{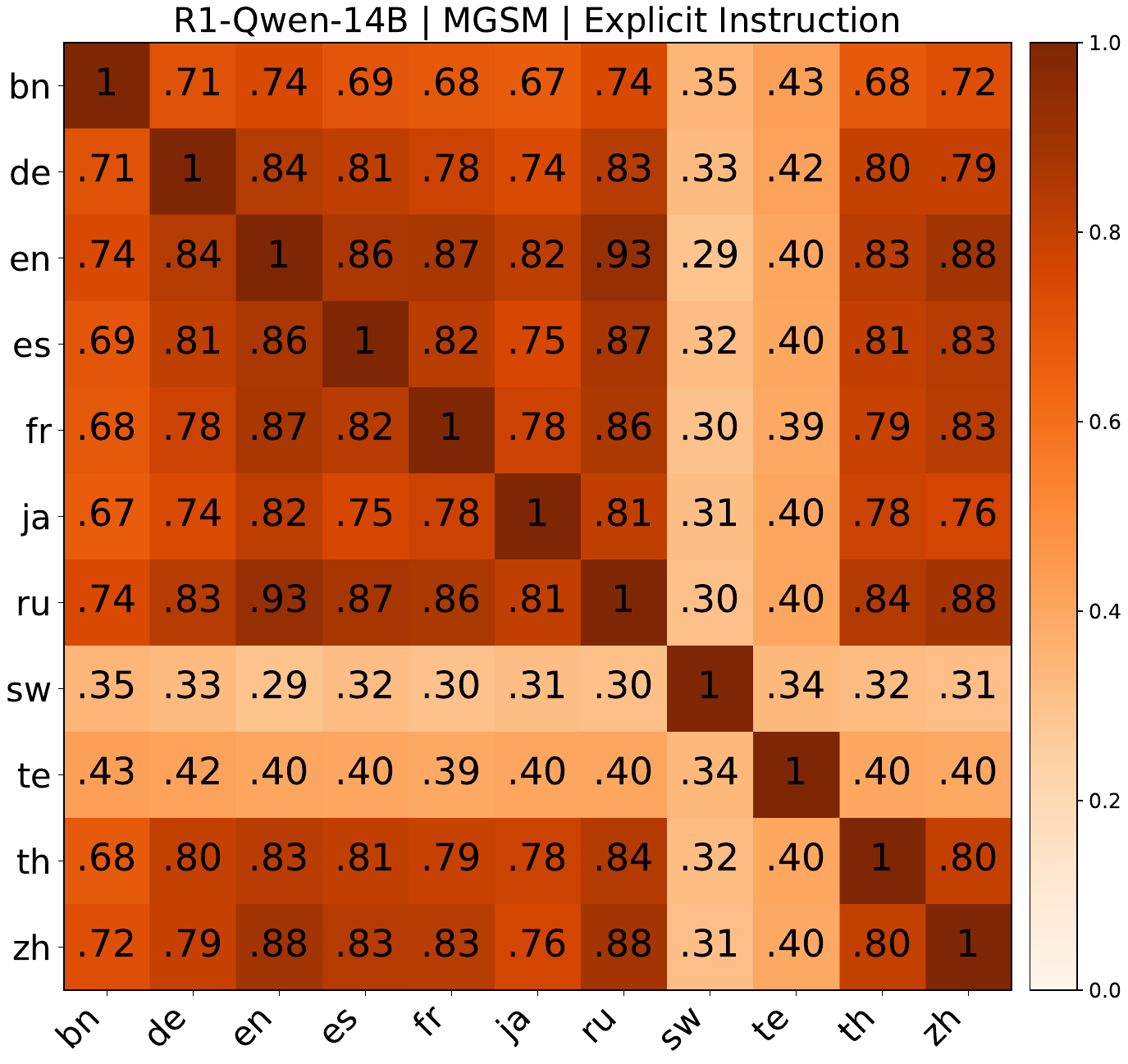}
    \includegraphics[width=0.24\textwidth]{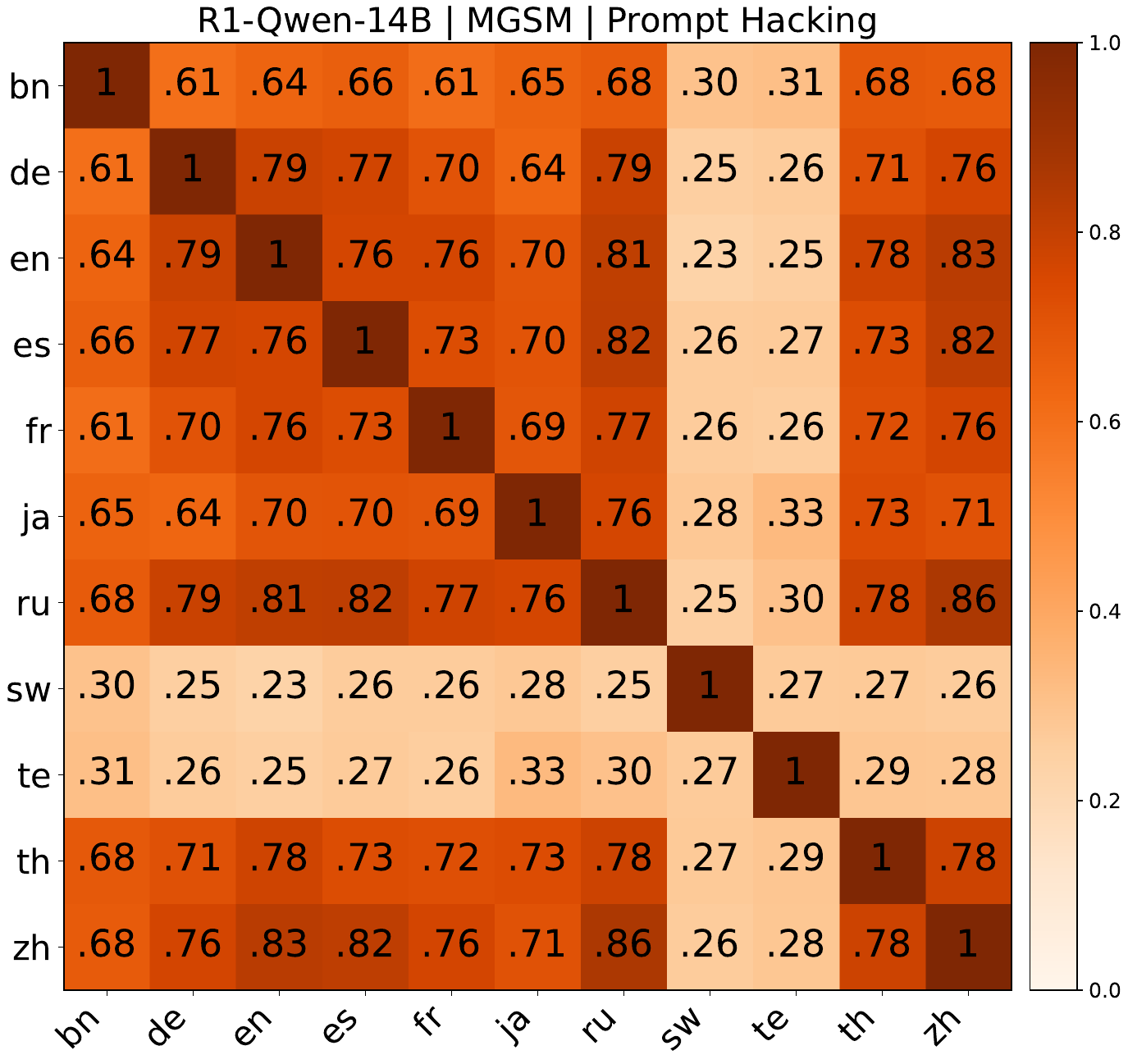}

    \includegraphics[width=0.24\textwidth]{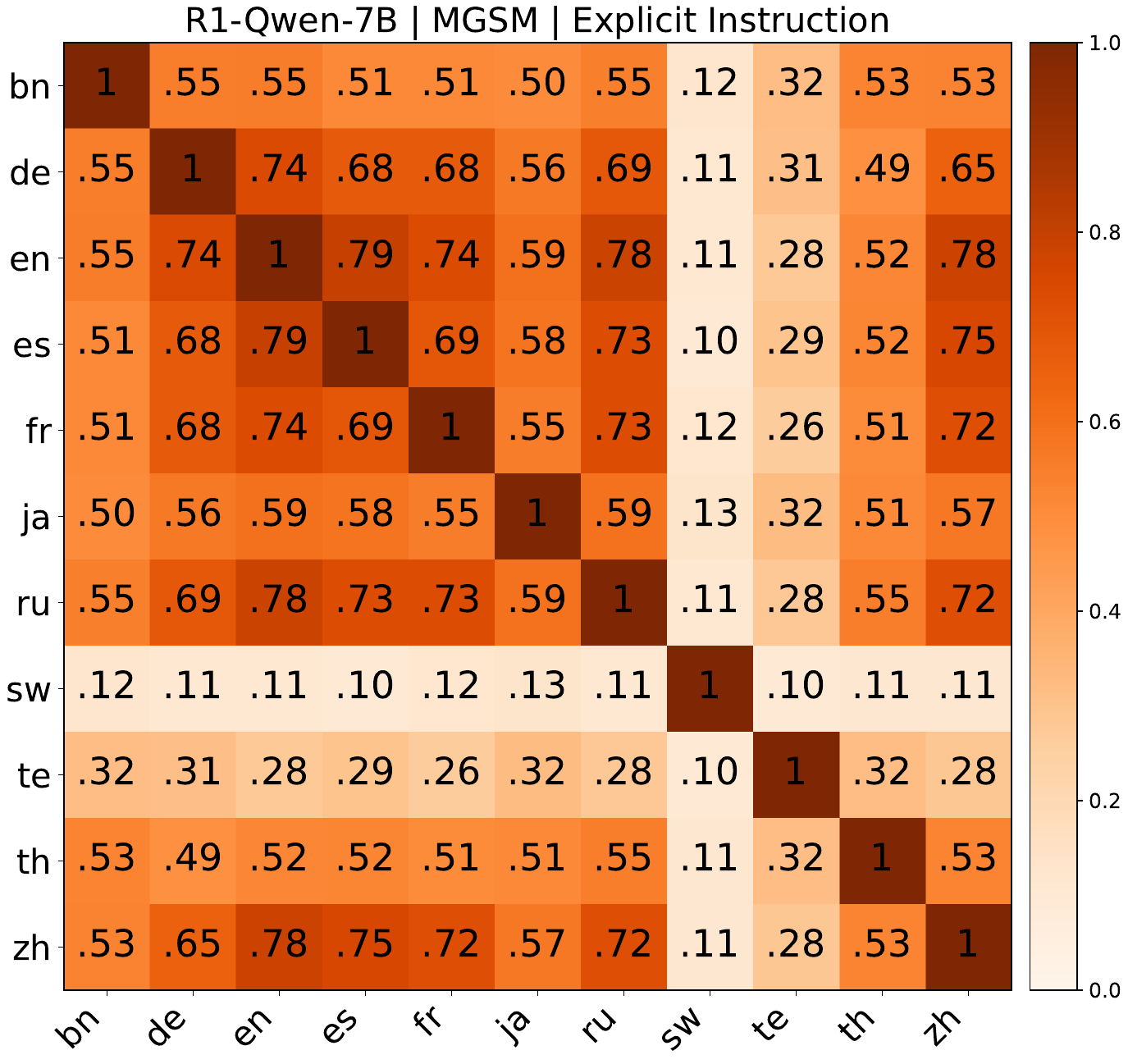}
    \includegraphics[width=0.24\textwidth]{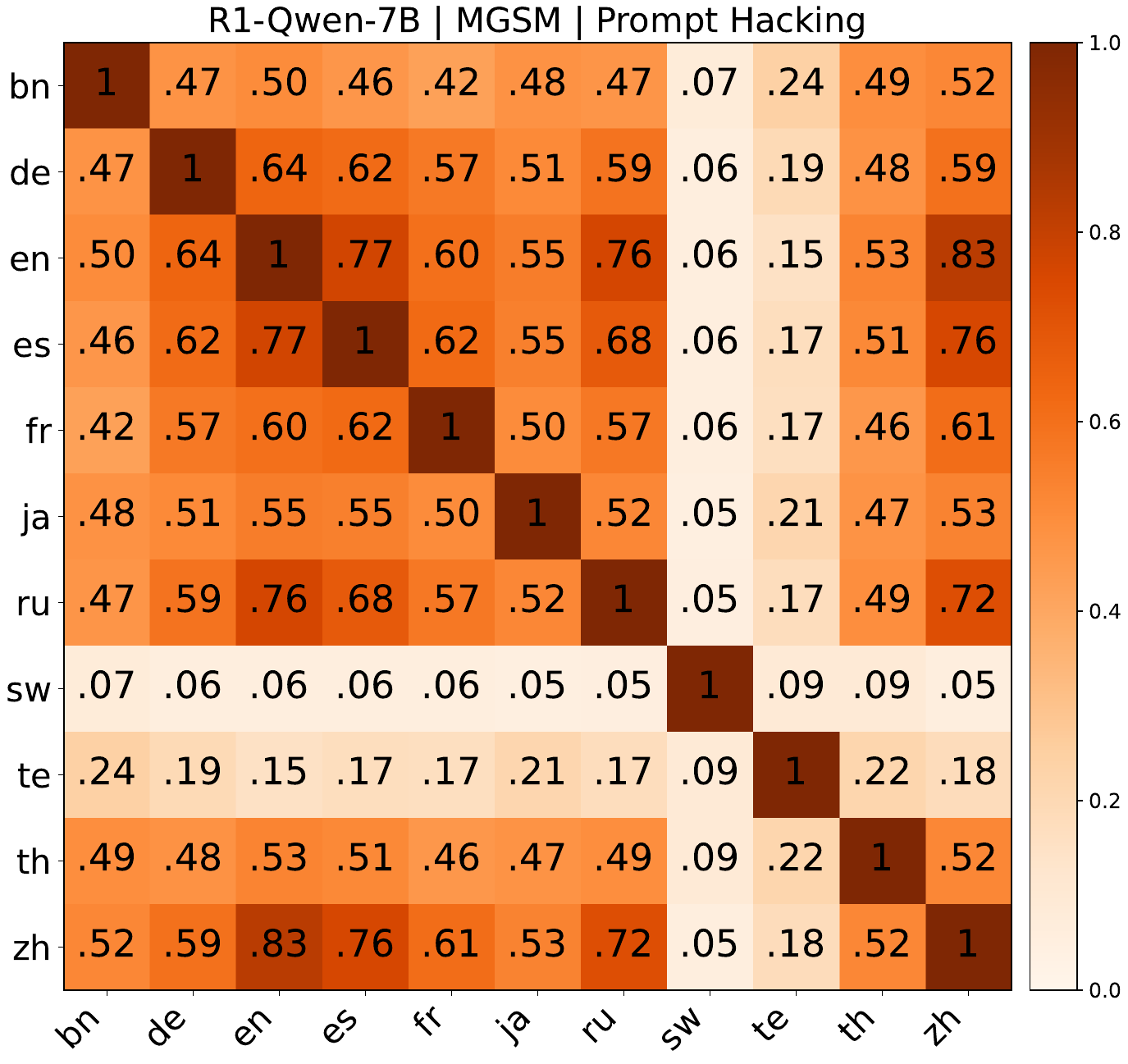}
    \includegraphics[width=0.24\textwidth]{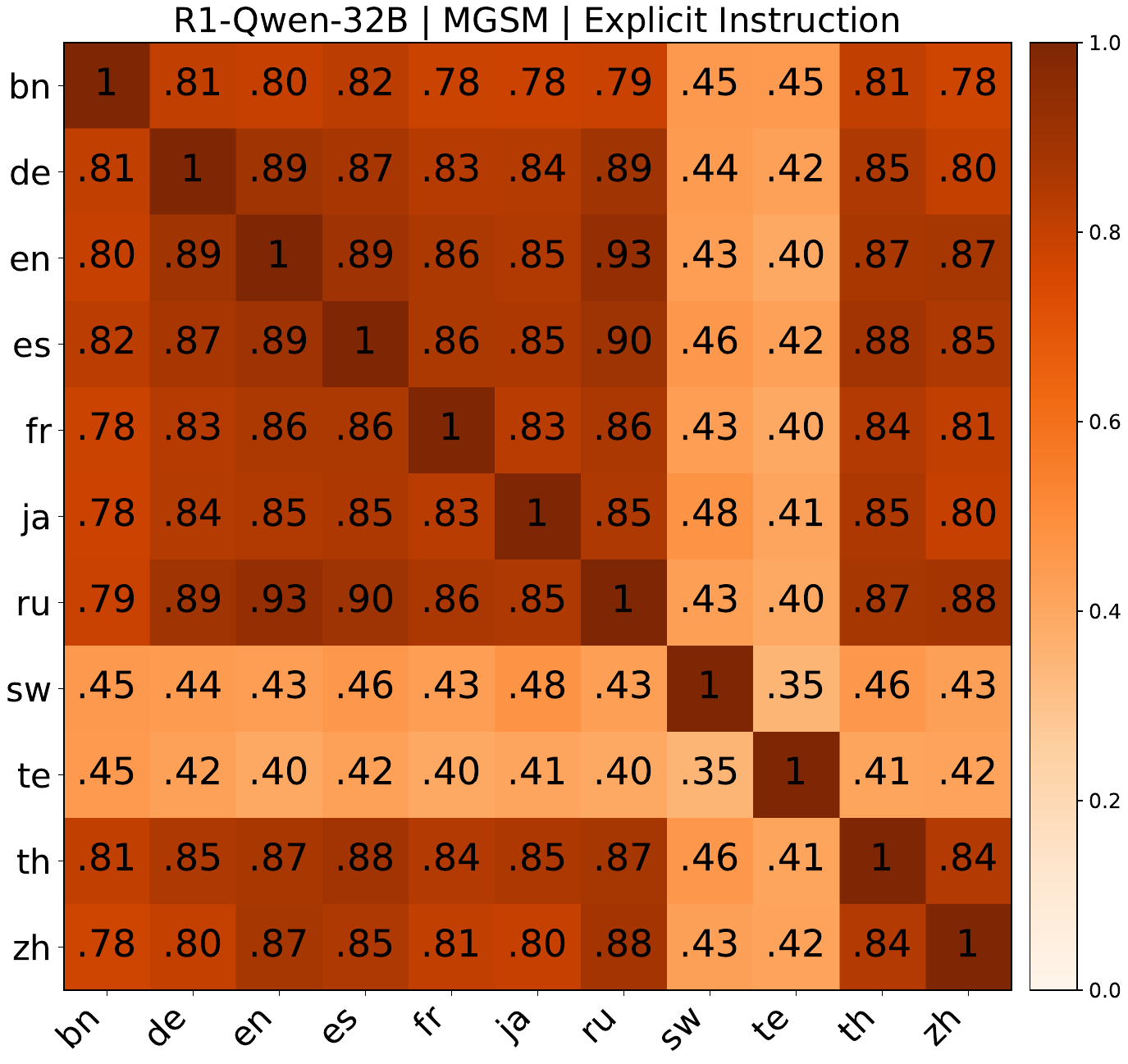}
    \includegraphics[width=0.24\textwidth]{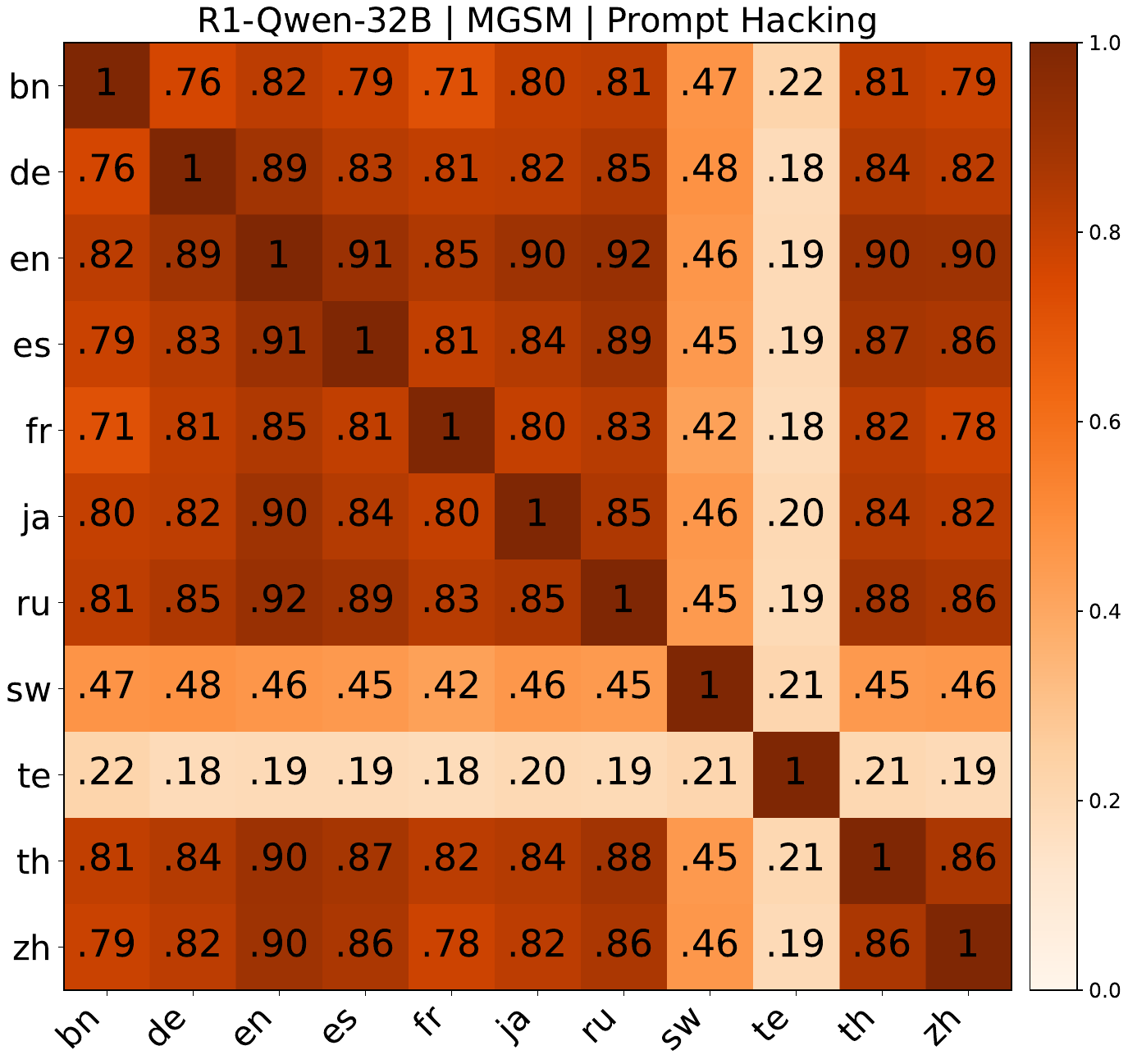}
    
    \includegraphics[width=0.24\textwidth]{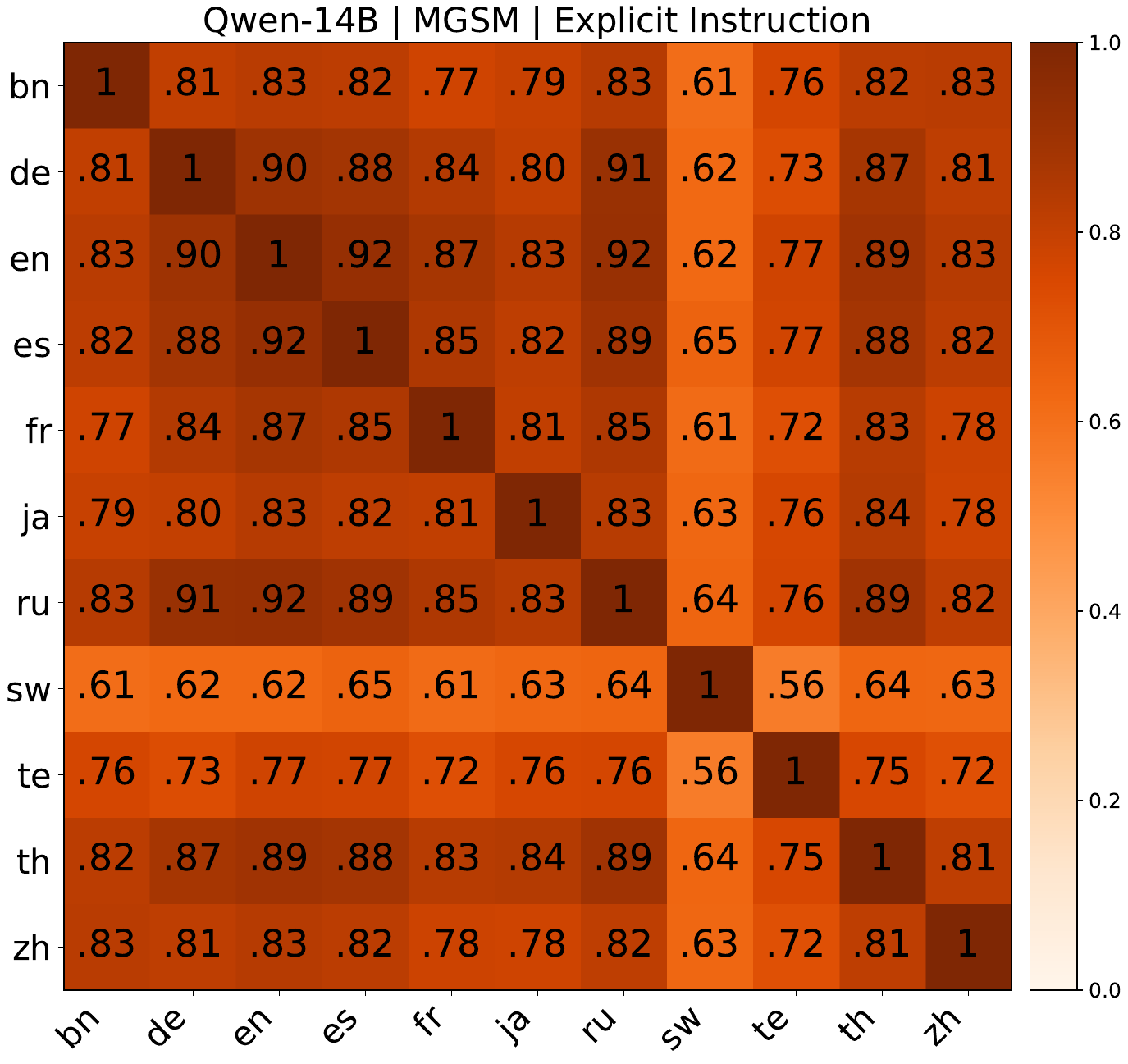}
    \includegraphics[width=0.24\textwidth]{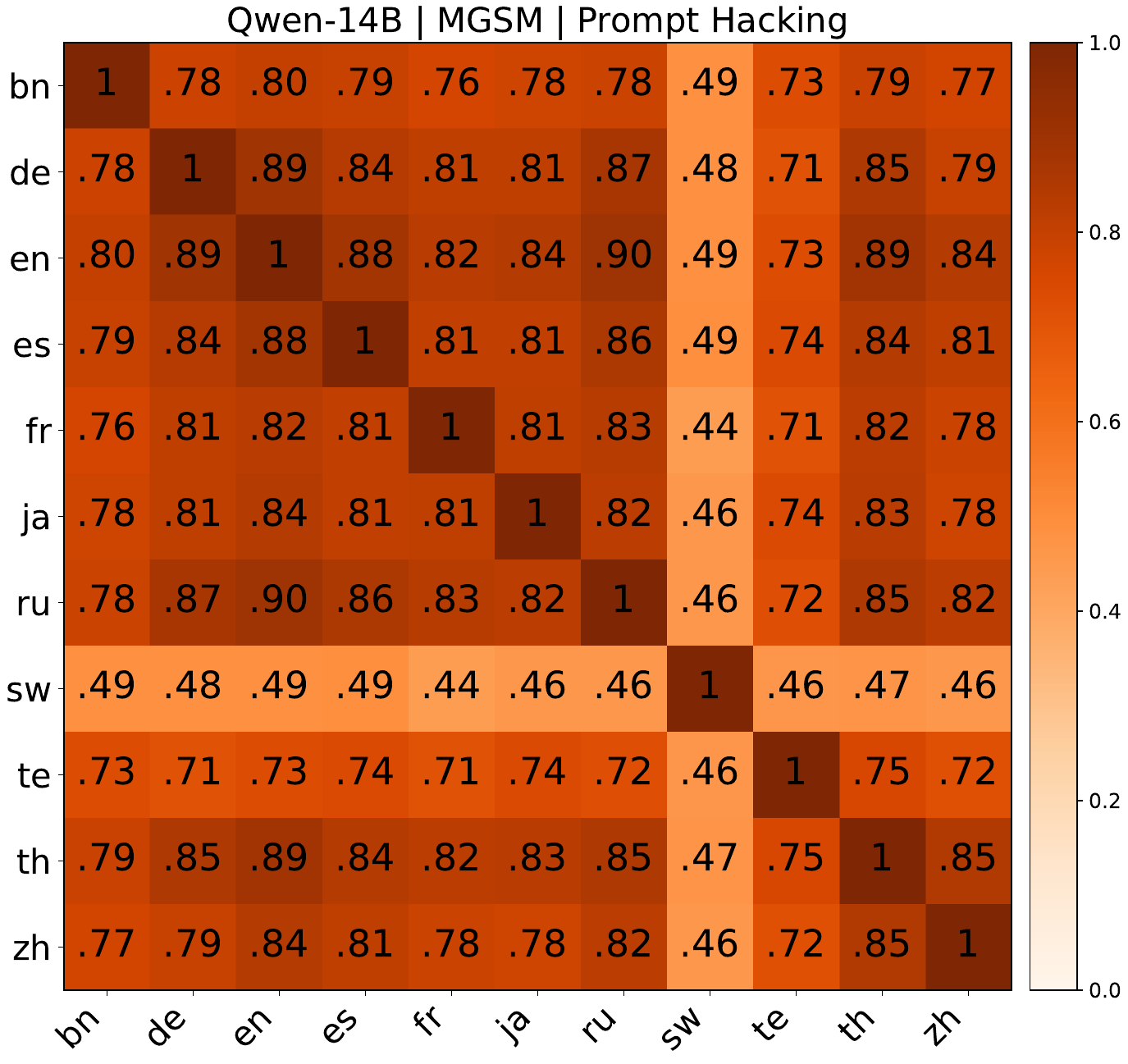}
    \includegraphics[width=0.24\textwidth]{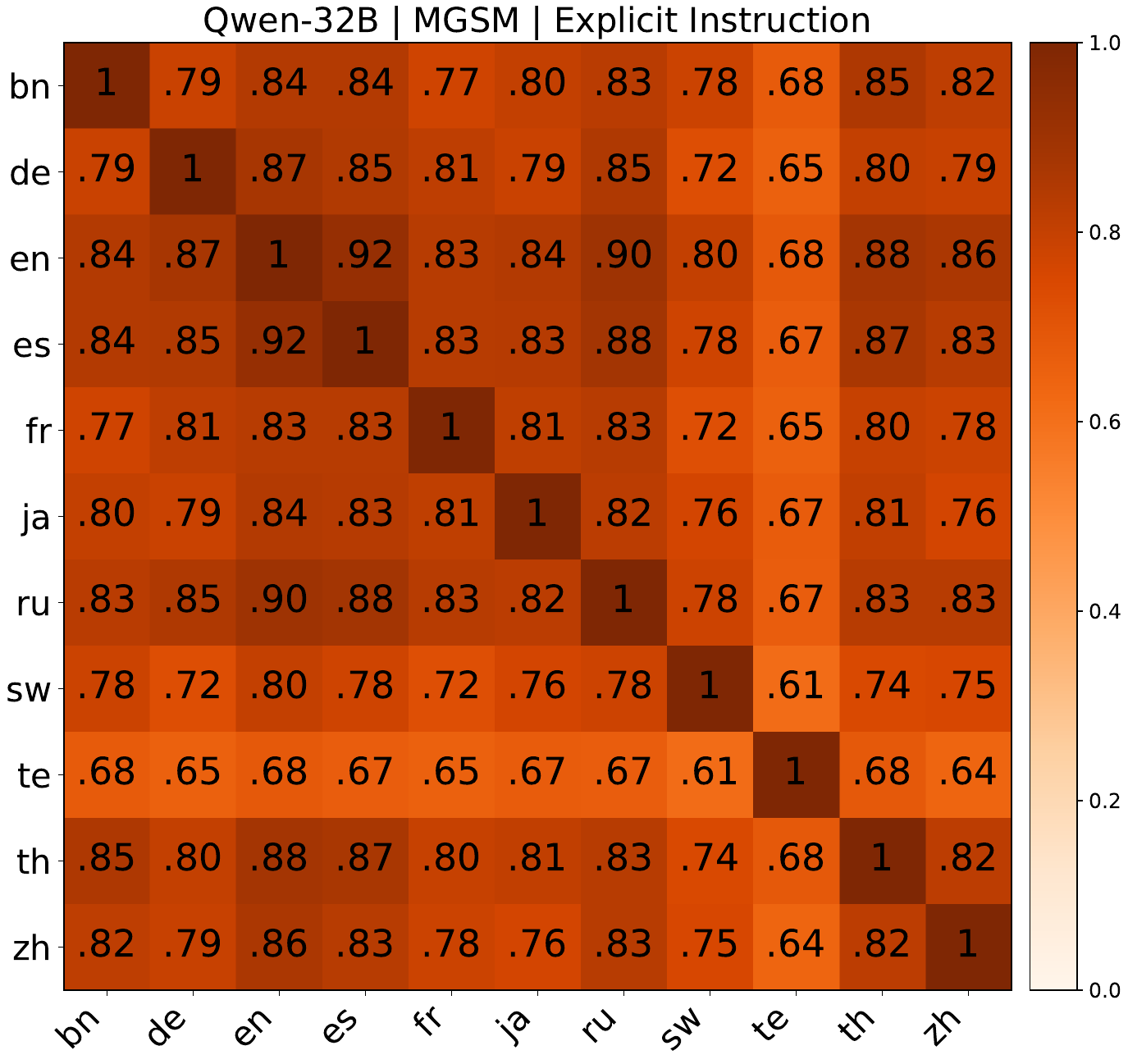}
    \includegraphics[width=0.24\textwidth]{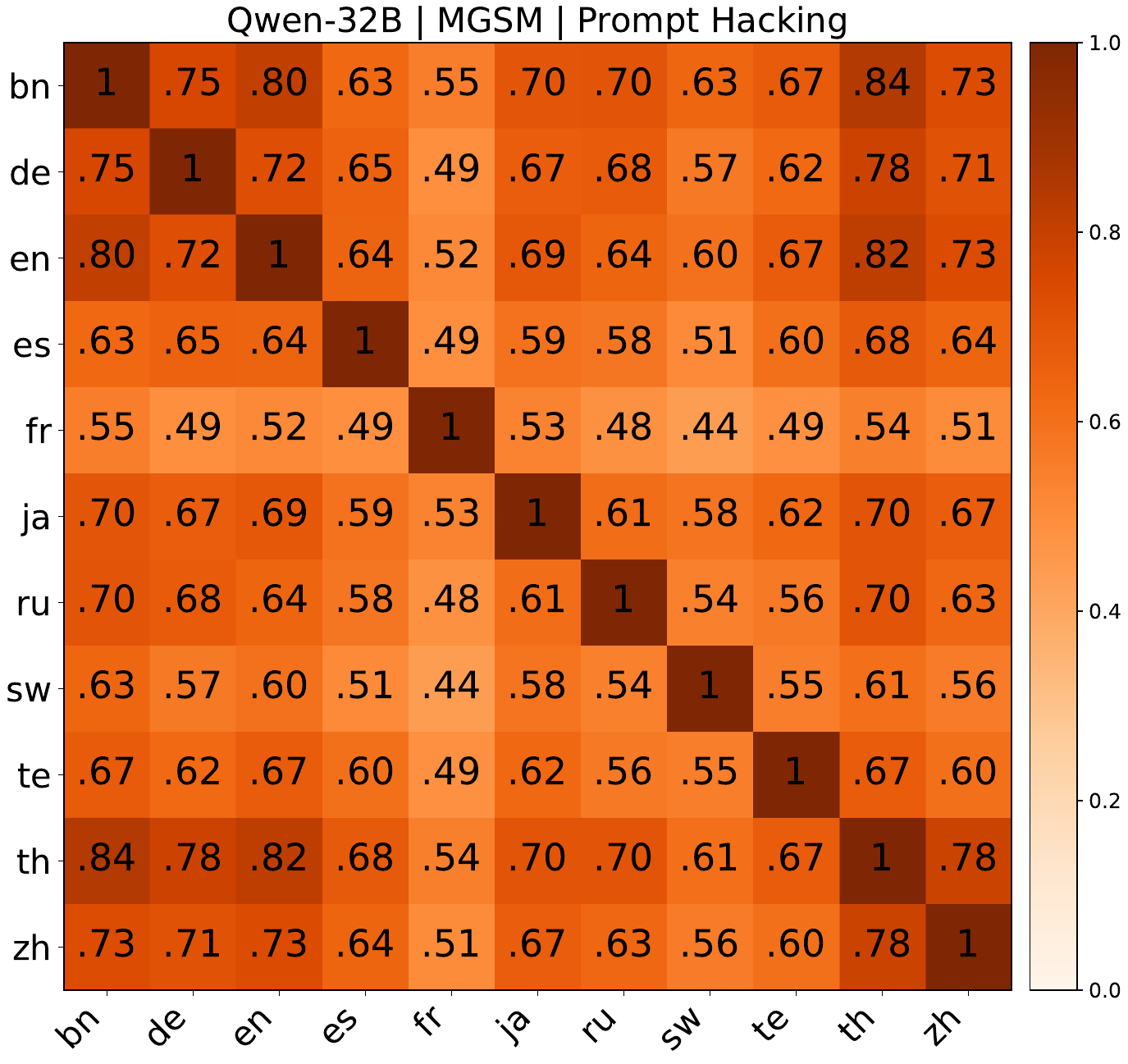}
    
    \includegraphics[width=0.24\textwidth]{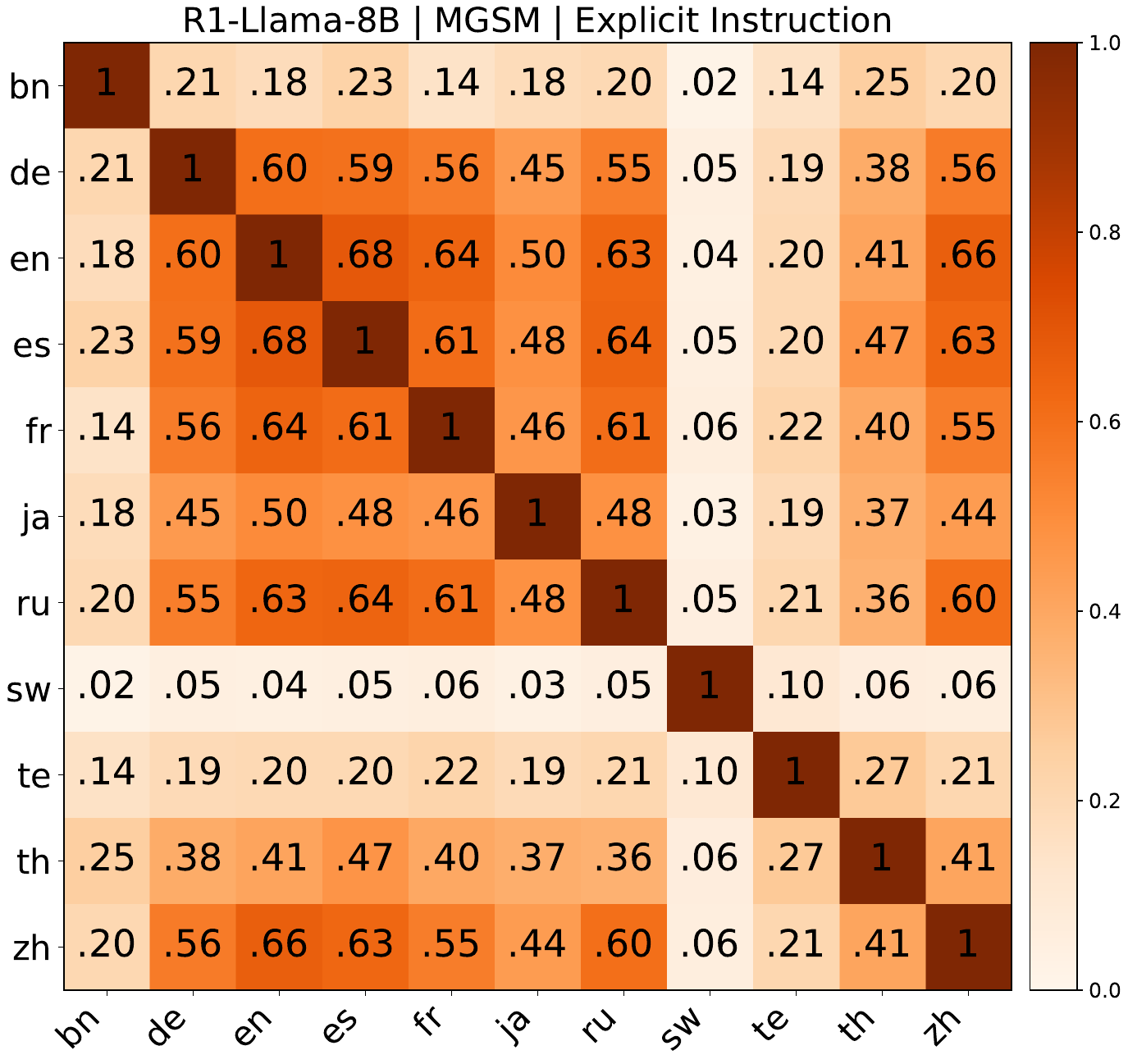}
    \includegraphics[width=0.24\textwidth]{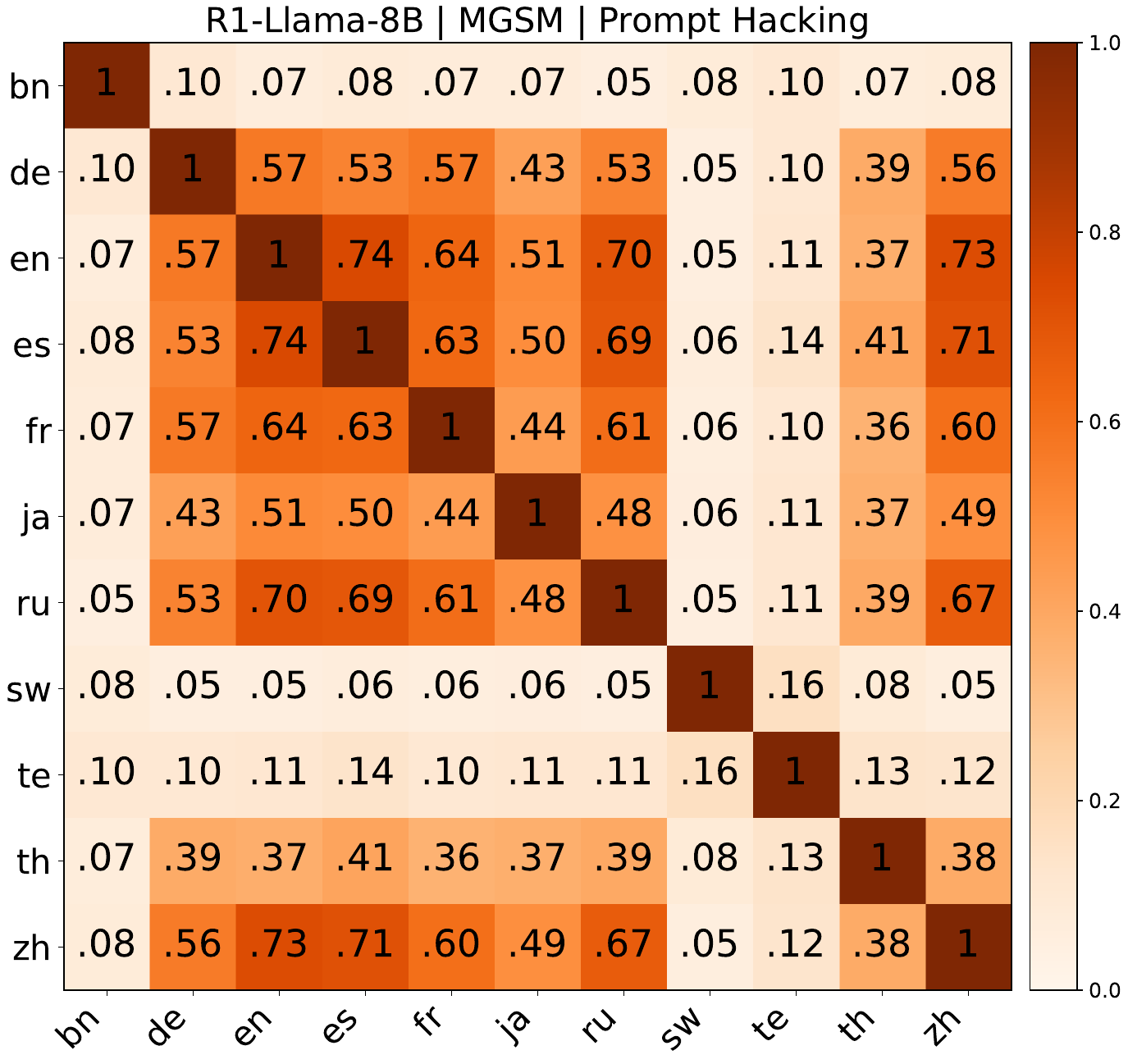}
    \includegraphics[width=0.24\textwidth]{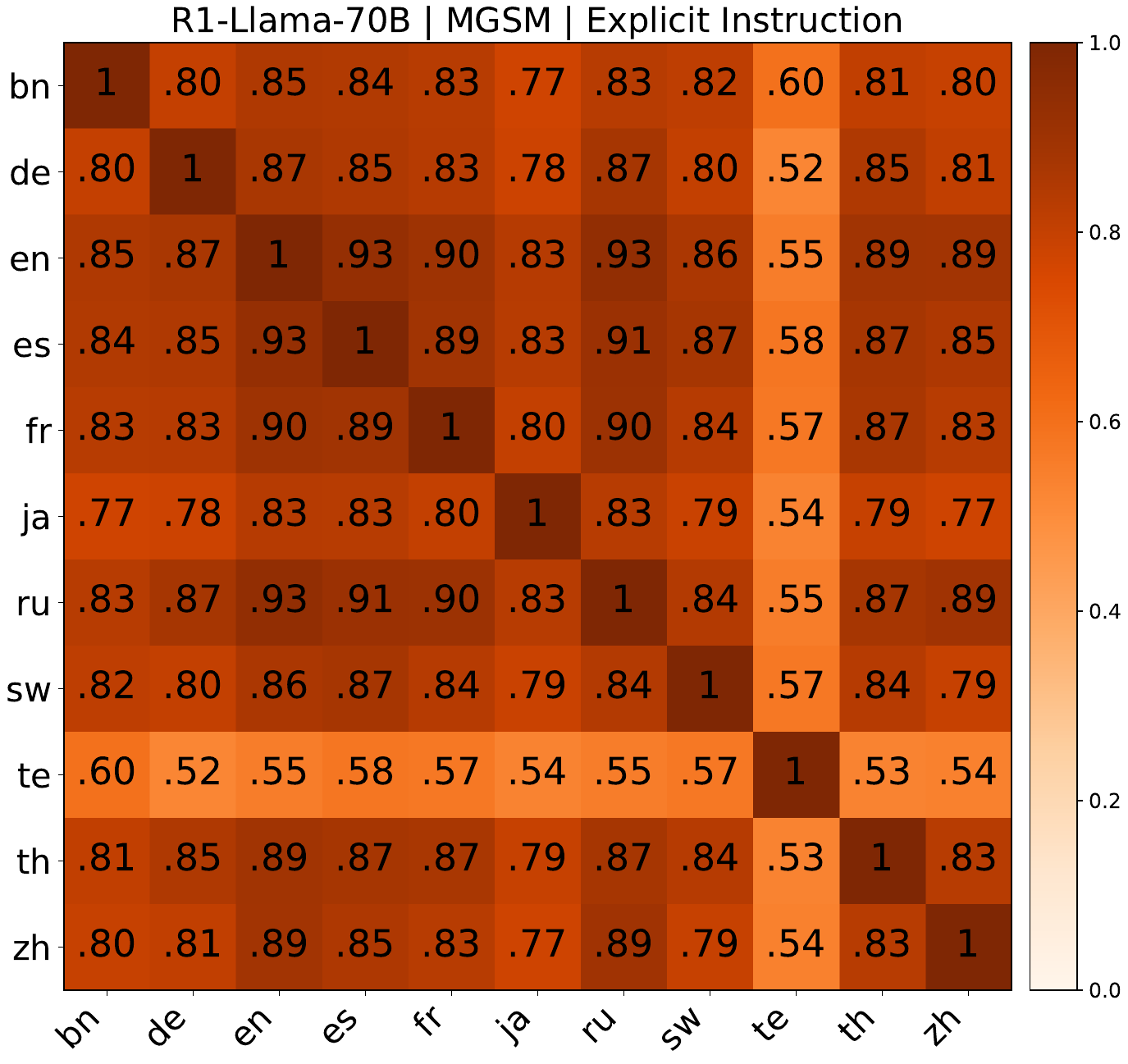}
    \includegraphics[width=0.24\textwidth]{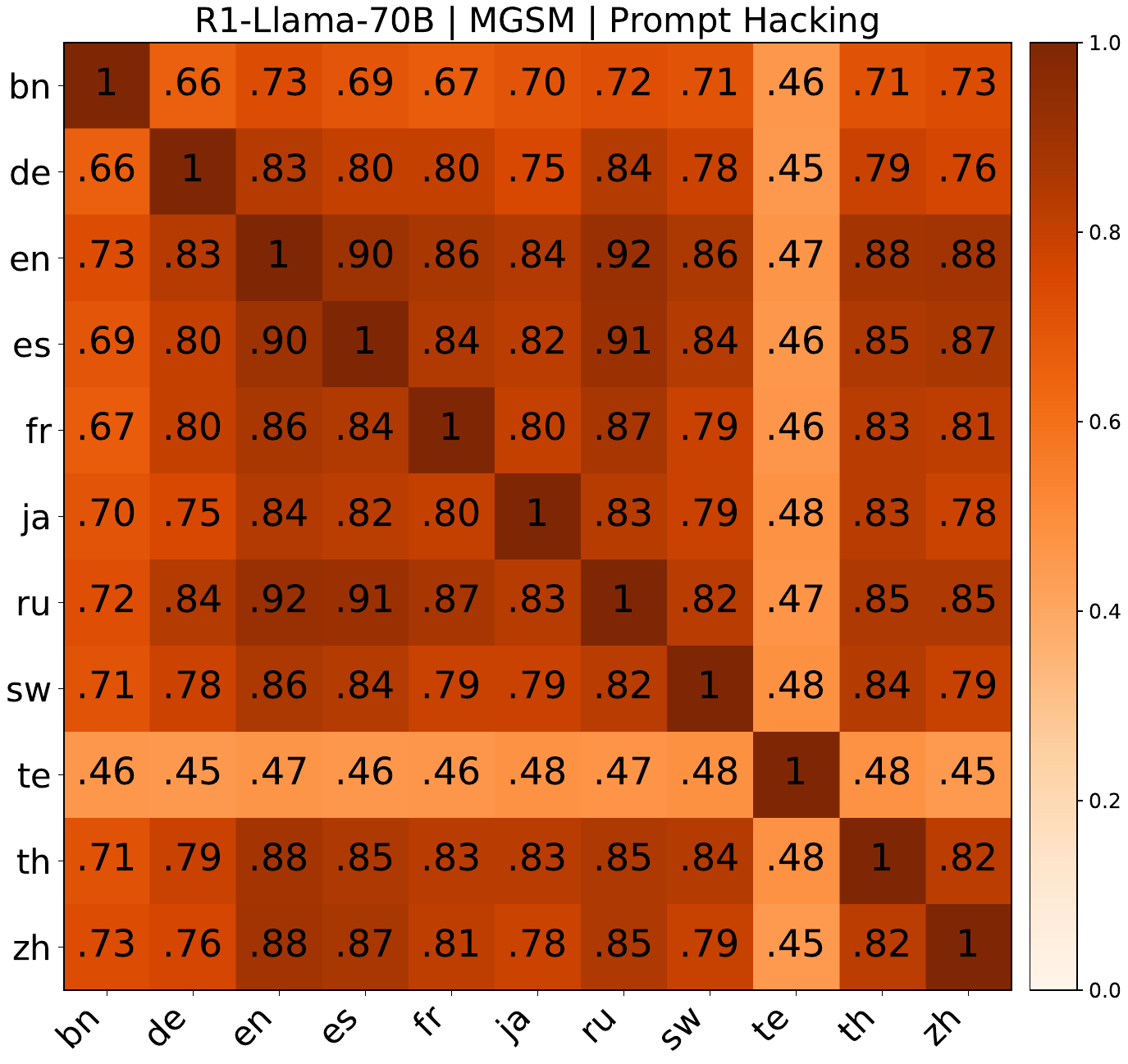}

    \caption{Final-answer consistency heatmaps on the MGSM dataset across different models under explicit instruction and prompt hacking.}
    \label{fig:cons_final_math}
\end{figure*}

\begin{figure*}[htbp]
    \centering
    \includegraphics[width=0.24\textwidth]{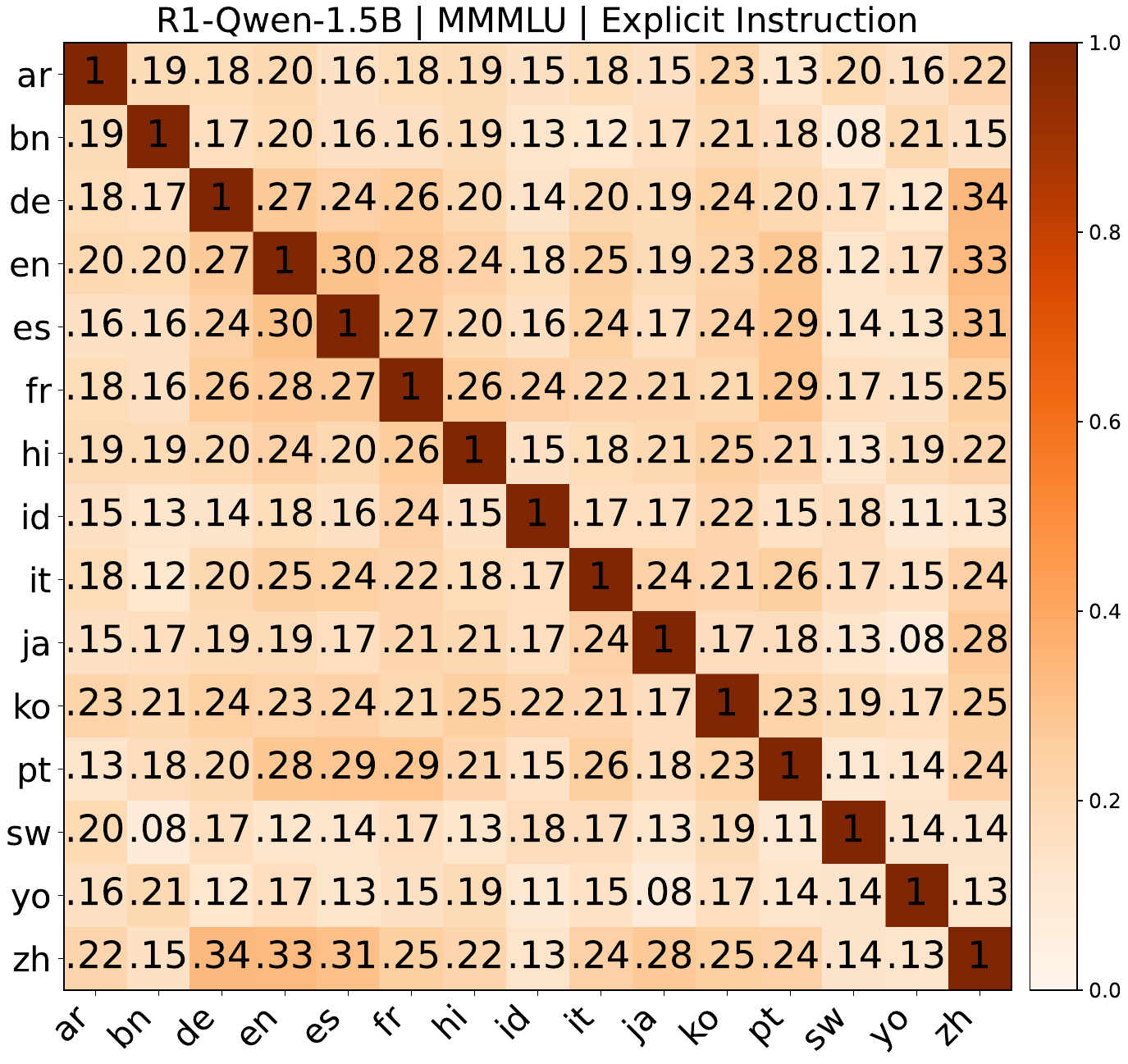}
    \includegraphics[width=0.24\textwidth]{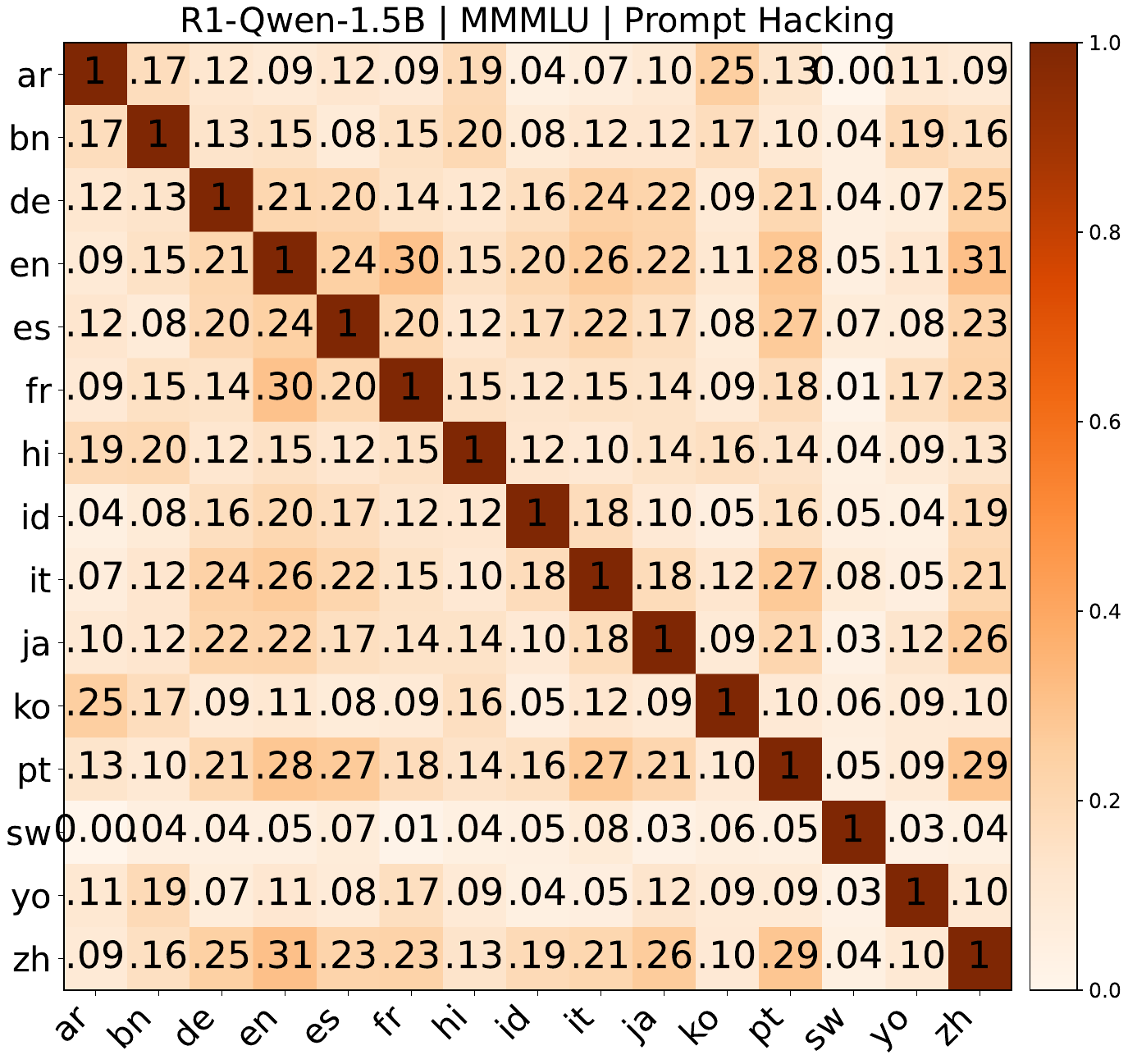}
    \includegraphics[width=0.24\textwidth]{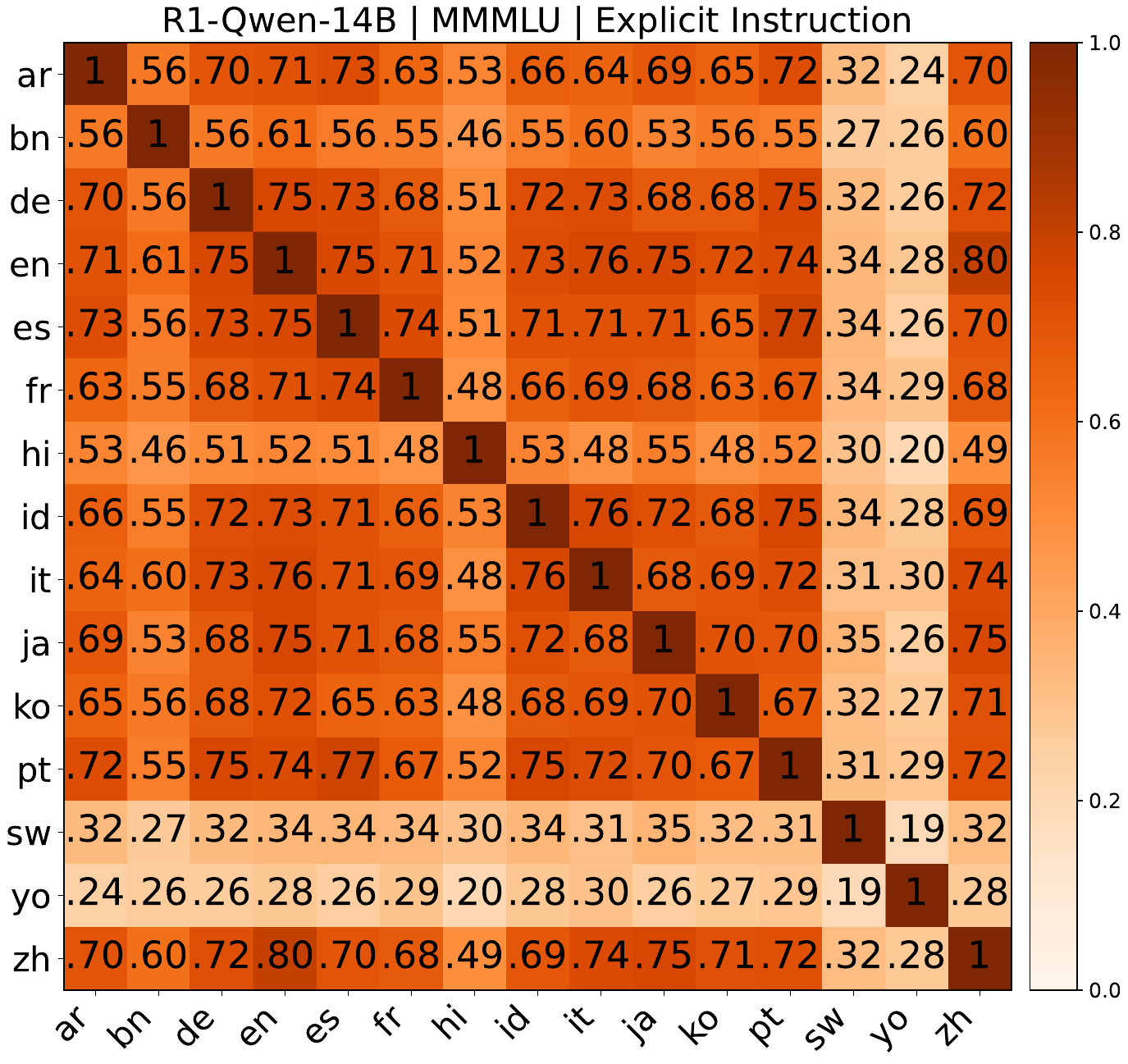}
    \includegraphics[width=0.24\textwidth]{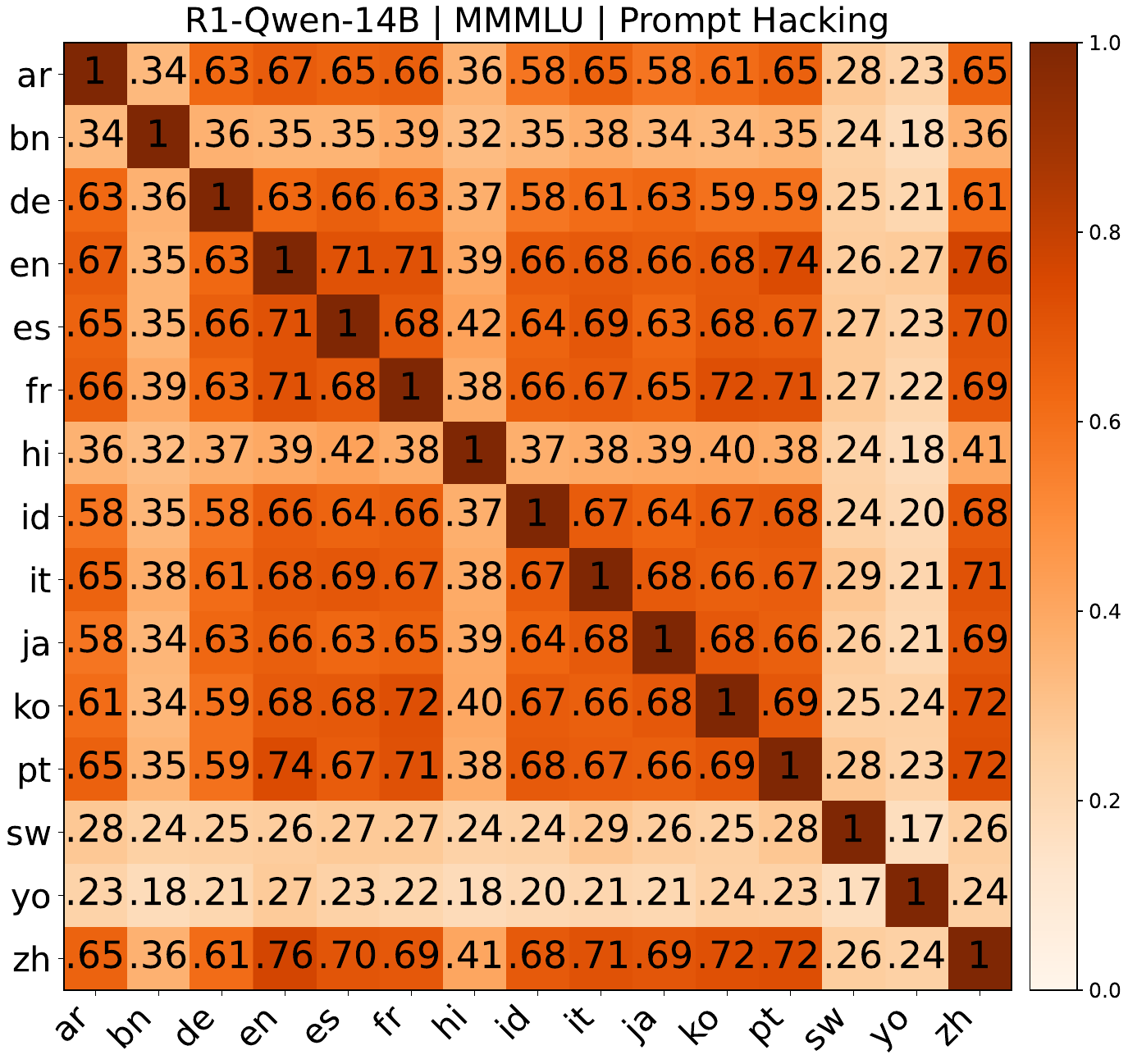}

    \includegraphics[width=0.24\textwidth]{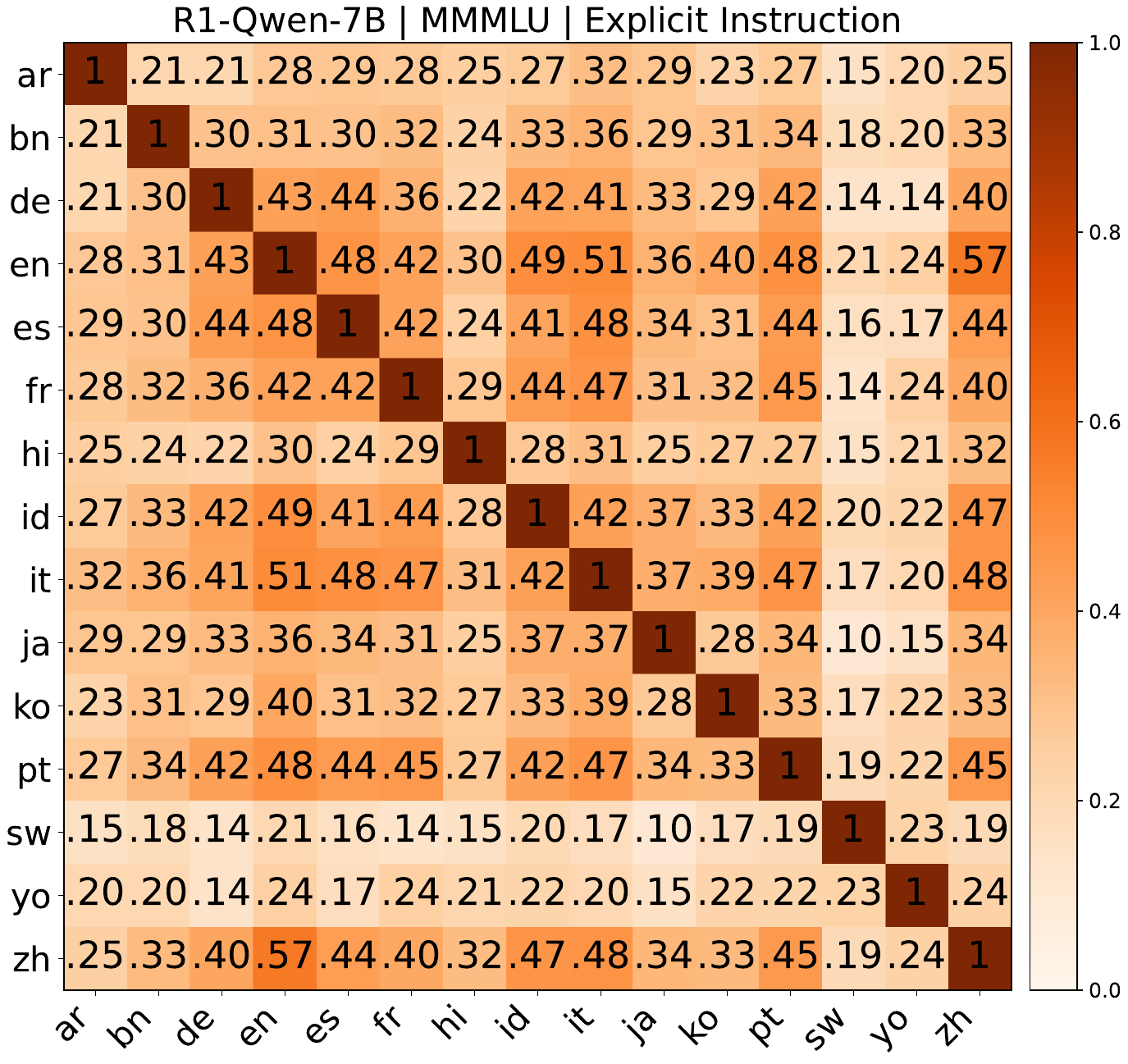}
    \includegraphics[width=0.24\textwidth]{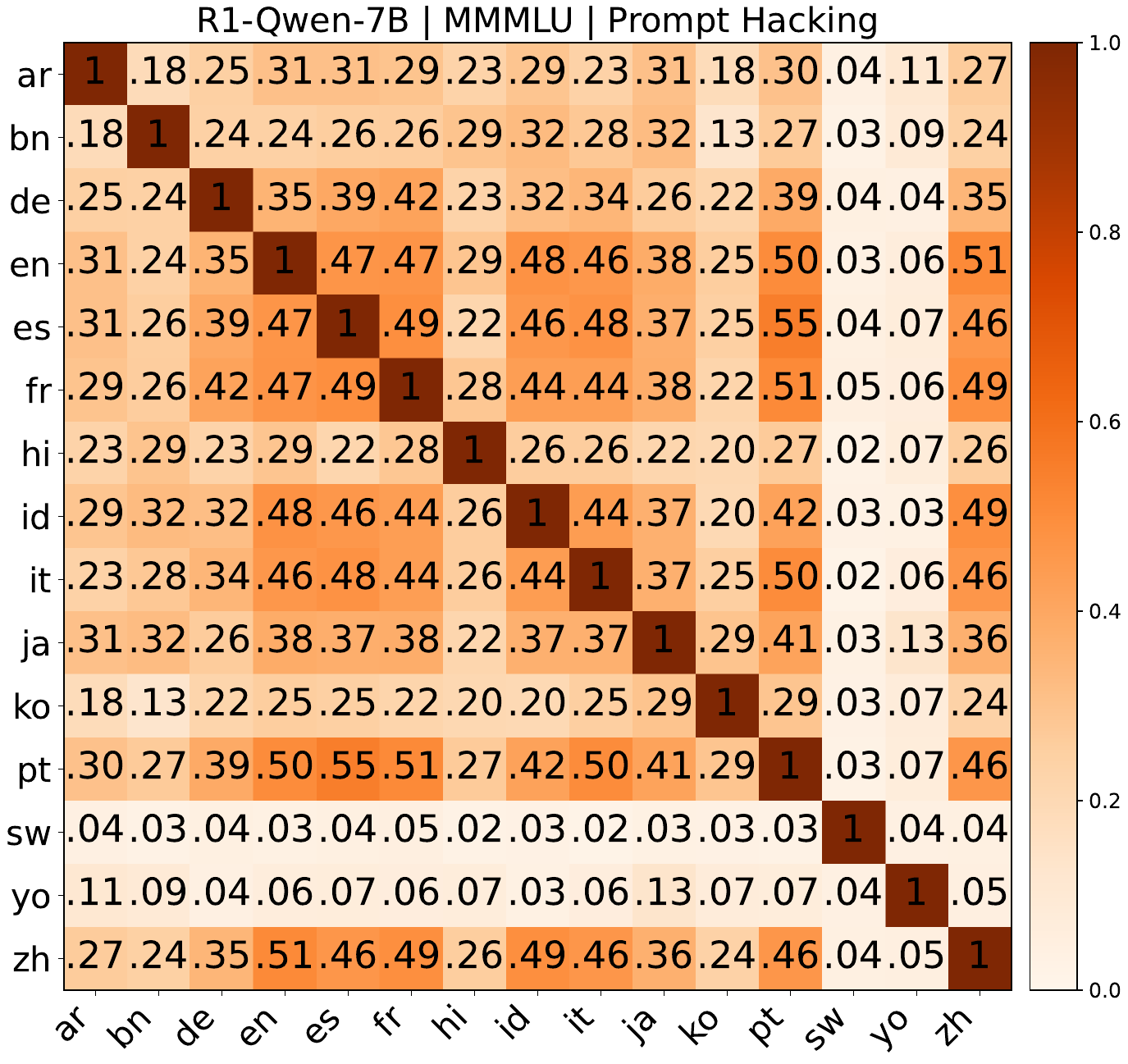}
    \includegraphics[width=0.24\textwidth]{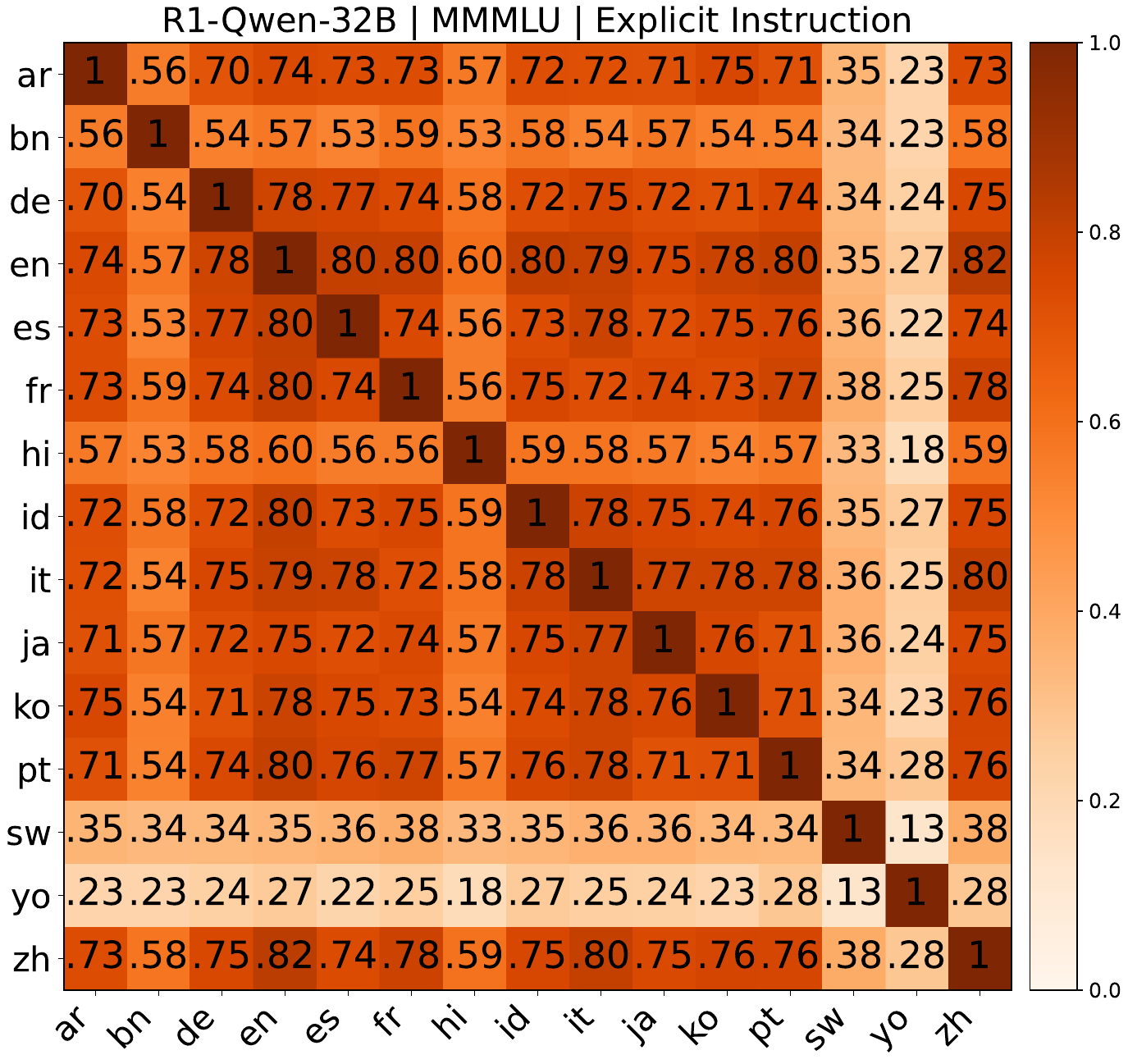}
    \includegraphics[width=0.24\textwidth]{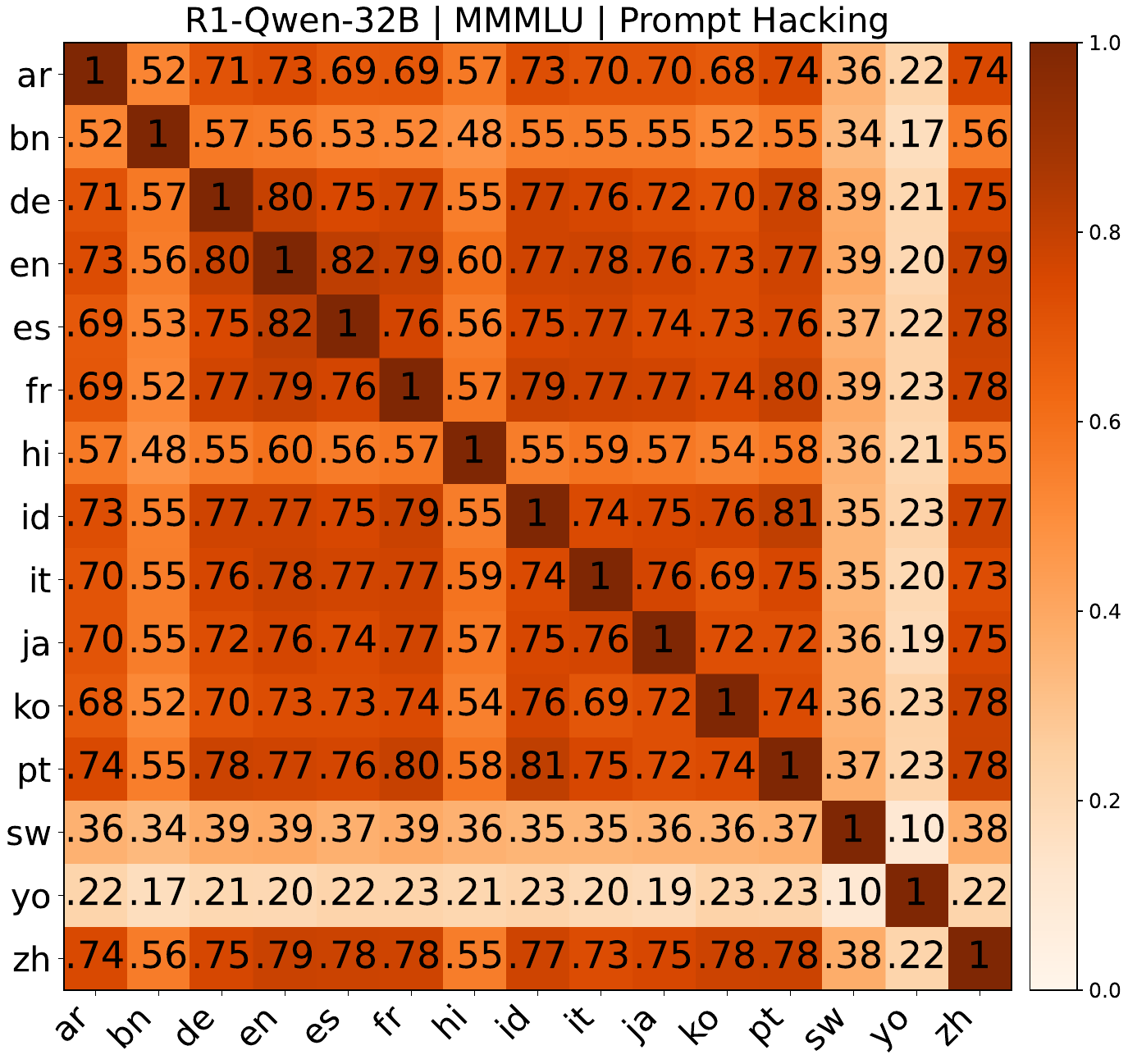}

    \includegraphics[width=0.24\textwidth]{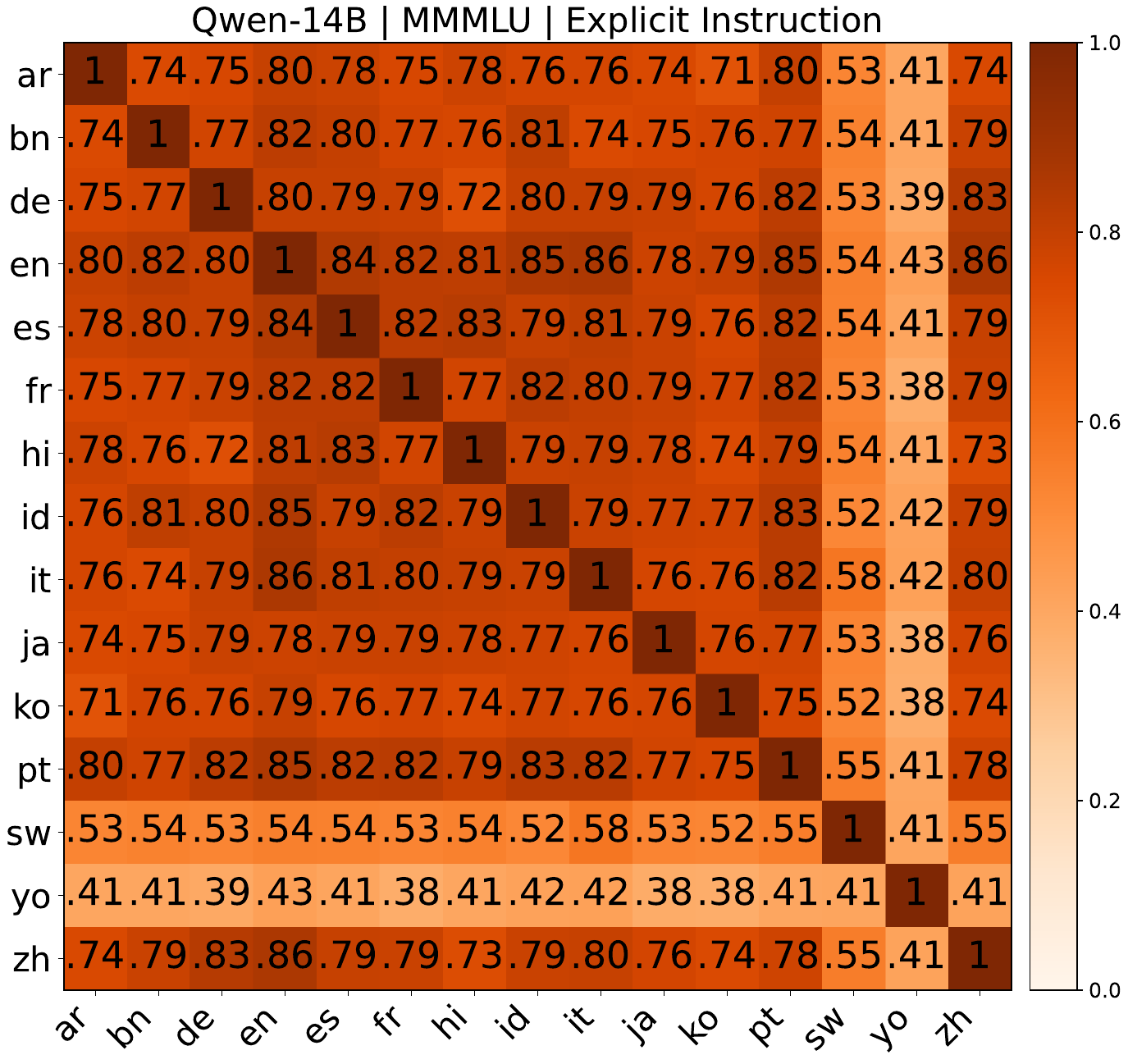}
    \includegraphics[width=0.24\textwidth]{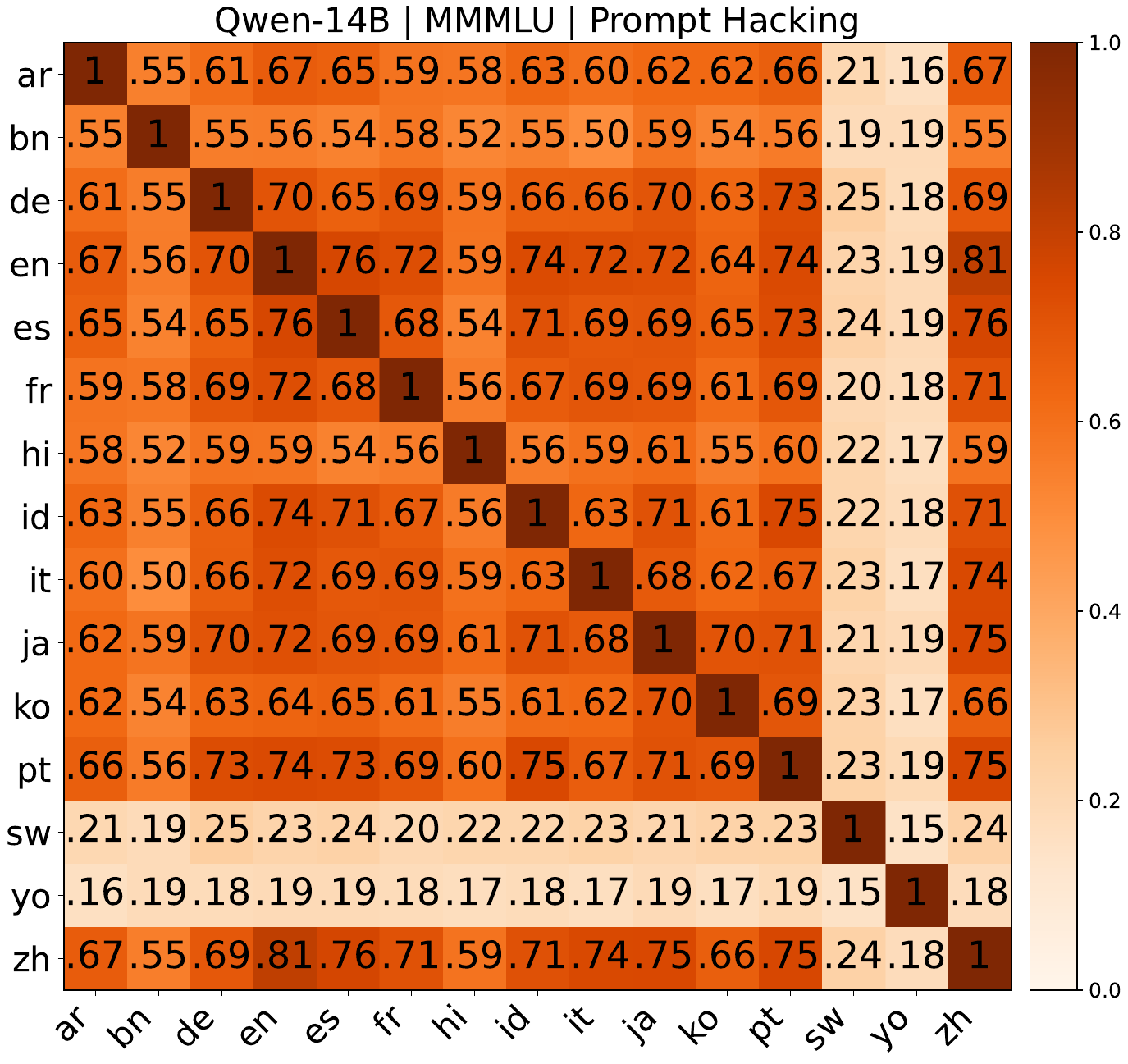}
    \includegraphics[width=0.24\textwidth]{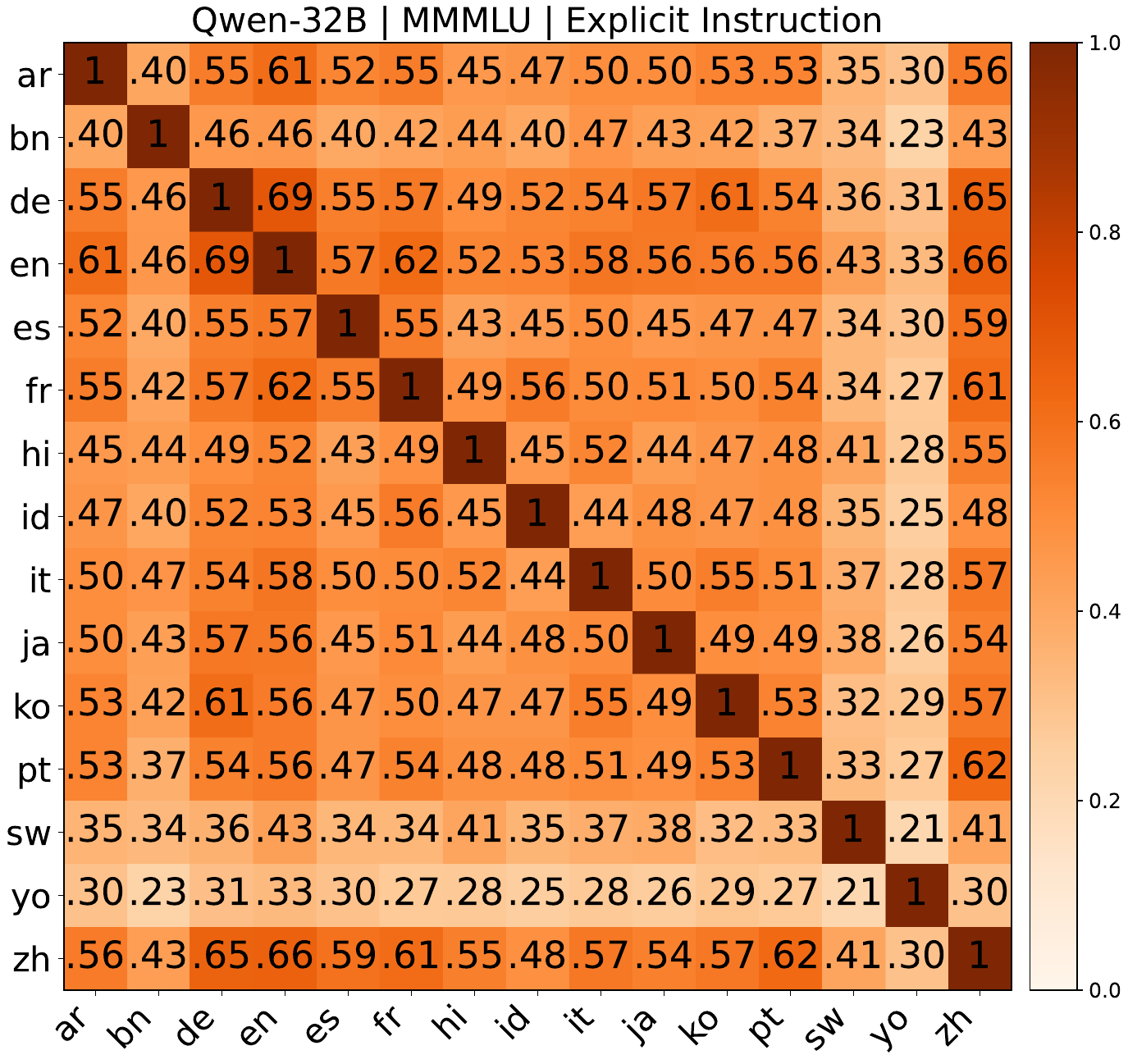}
    \includegraphics[width=0.24\textwidth]{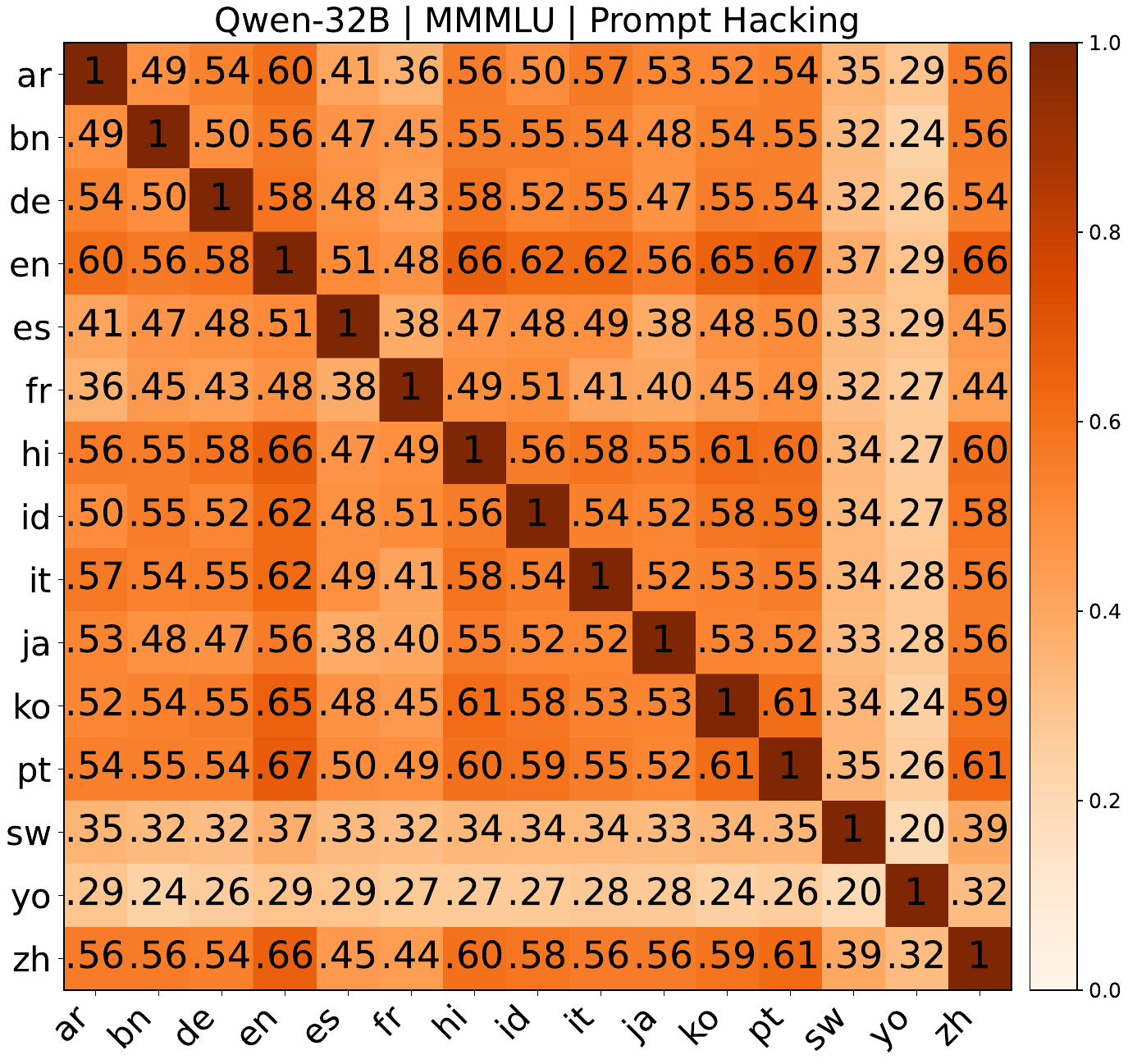}

    \includegraphics[width=0.24\textwidth]{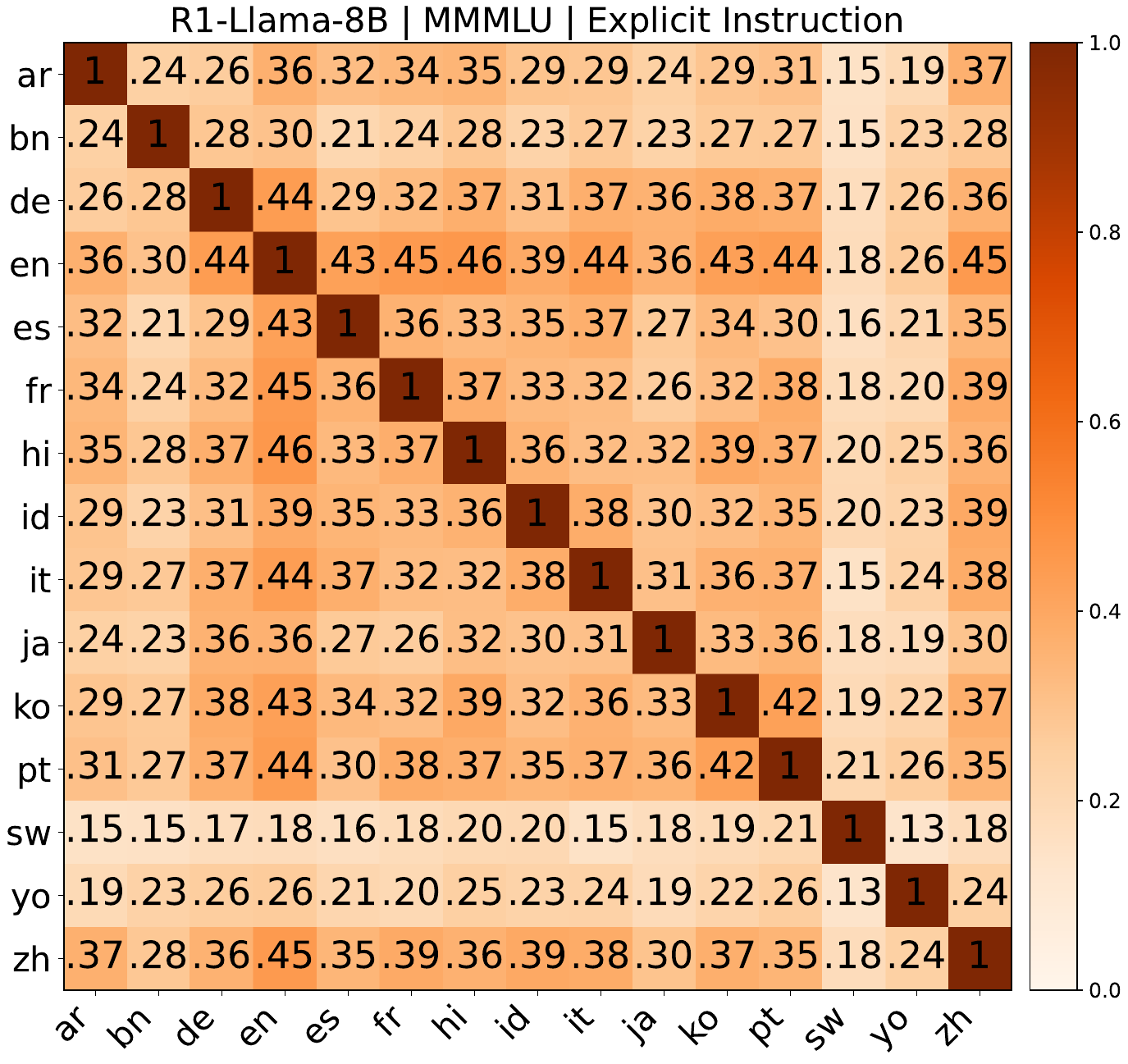}
    \includegraphics[width=0.24\textwidth]{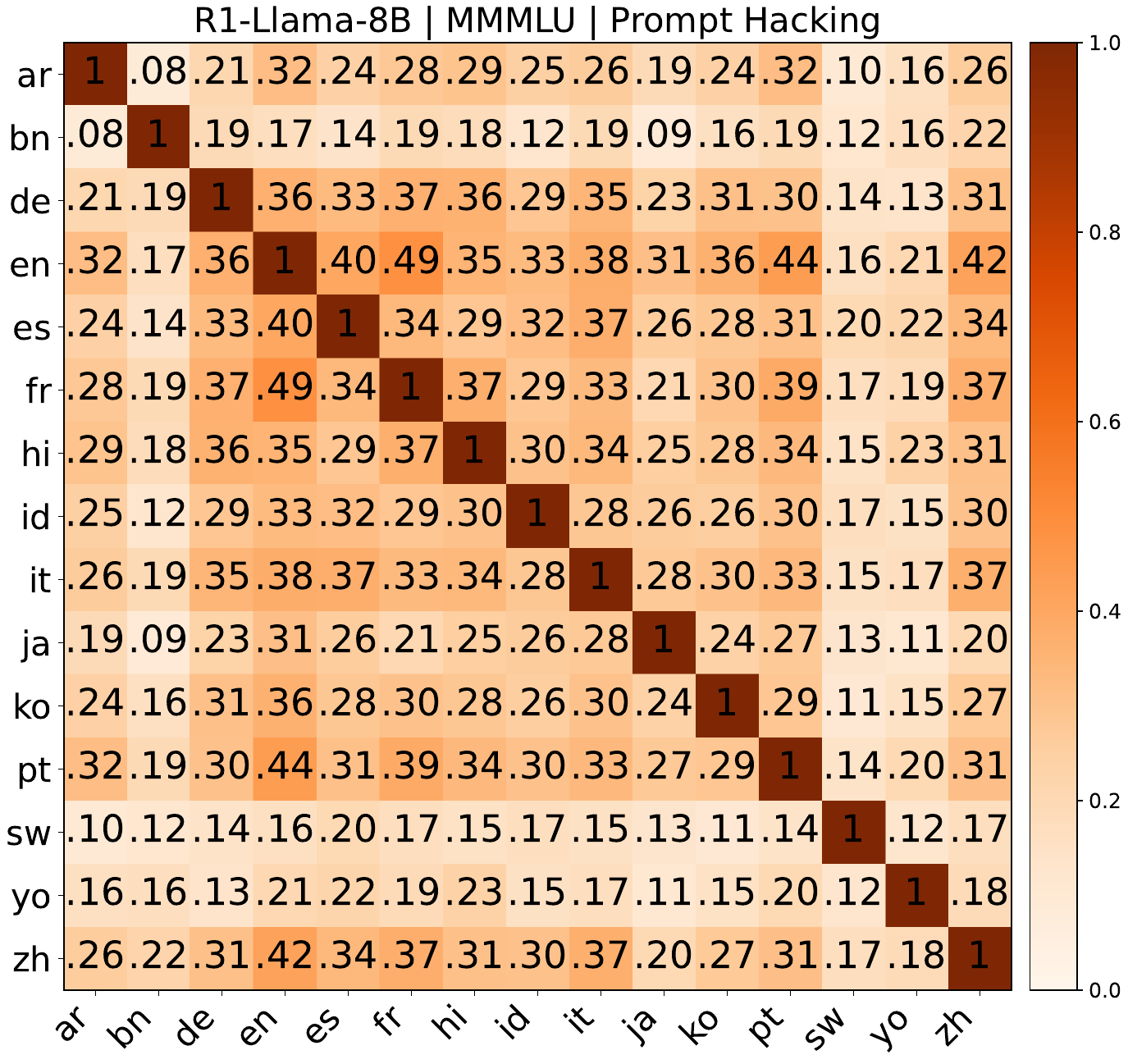}
    \includegraphics[width=0.24\textwidth]{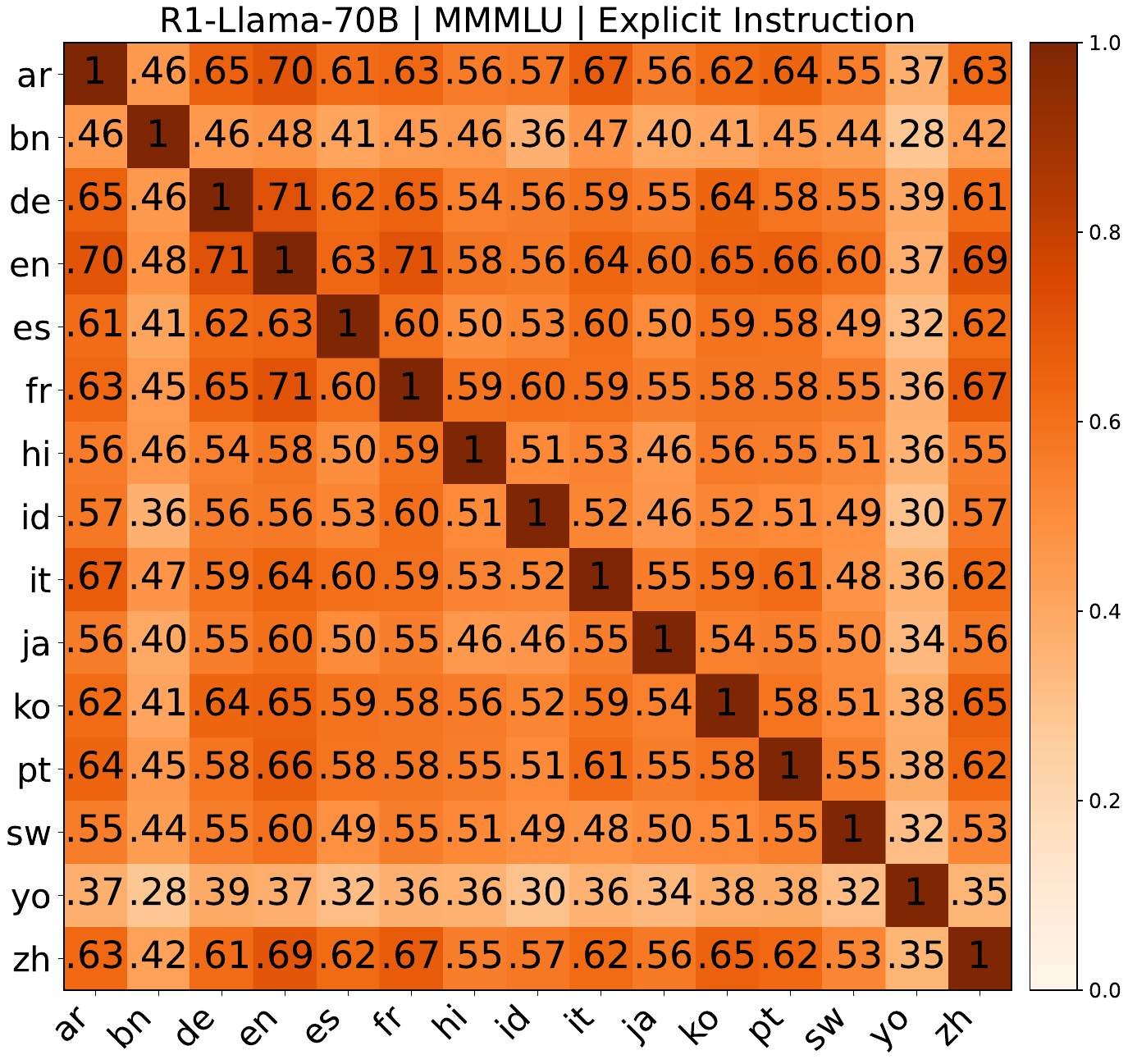}
    \includegraphics[width=0.24\textwidth]{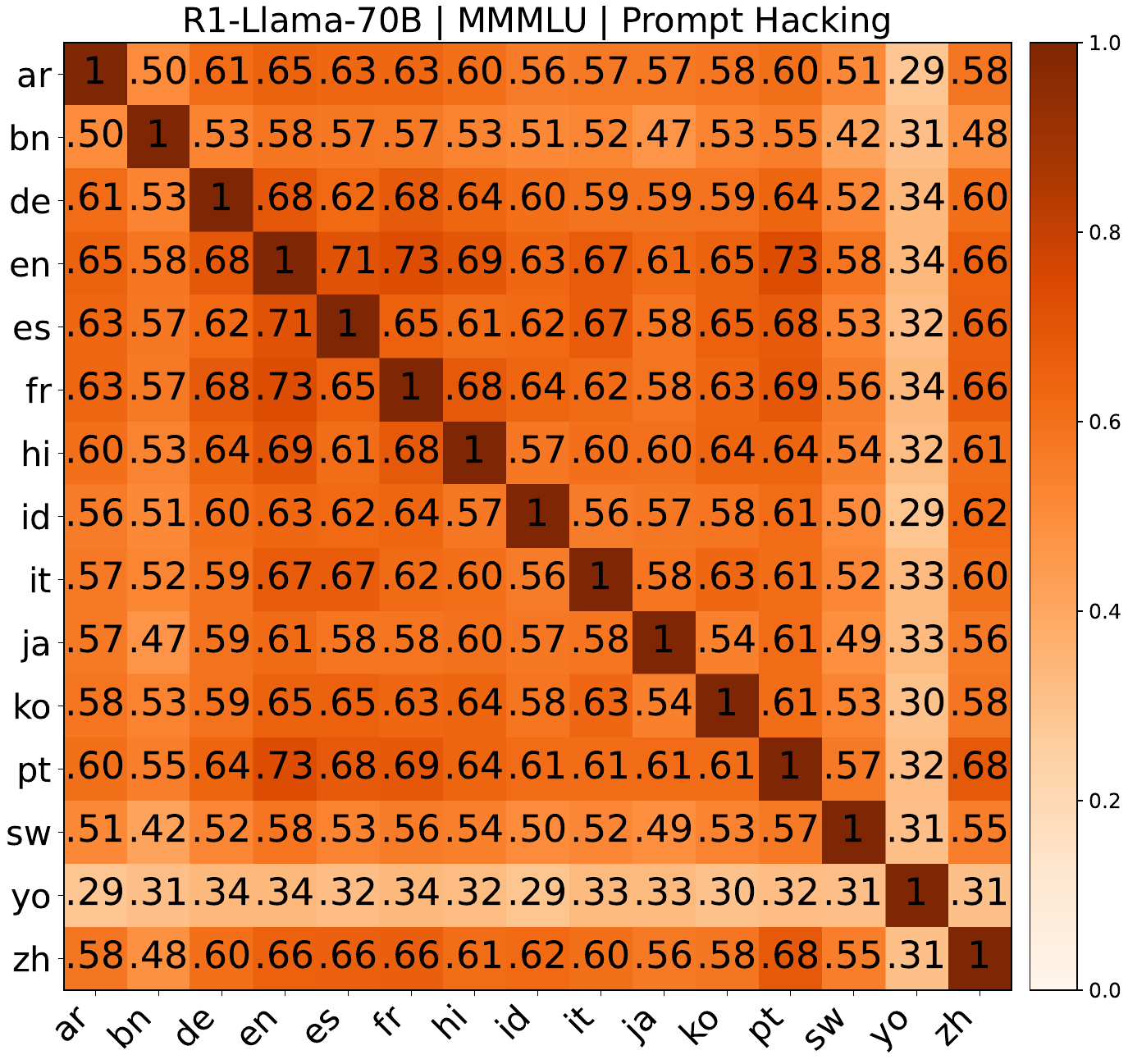}

    \caption{Final-answer consistency heatmaps on the MMLU dataset across different models under explicit instruction and prompt hacking.}
    \label{fig:cons_final_mcq}
\end{figure*}

\seclabel{final_answer_consistency}
Figures~\ref{fig:cons_final_math} and~\ref{fig:cons_final_mcq} show the consistency results on MMMLU and MGSM of LRMs. Across both tasks, the scaling trend remains stable: larger models exhibit higher consistency. However, we also observe that applying prompt hacking tends to affect the internal language consistency of models, sometimes reducing alignment compared to explicit instructions.

\subsubsection{Language Compliance}
Table~\ref{tab:compliance_math} and Table~\ref{tab:compliance_mcq} report sentence-level and token-level compliance statistics on MMMLU and MGSM. Overall, prompt hacking improves alignment with the target language, with stronger effects for low-resource languages. Token-level compliance is consistently lower than sentence-level compliance, likely due to the difficulty of identifying language from individual tokens and the presence of borrowings or quoted words within sentences.

Table~\ref{tab:compliance_en_math} and Table~\ref{tab:compliance_en_mcq} present the proportions of Chinese and English content in the reasoning traces across different models and prompt languages. We observe that, under explicit instruction, models tend to default to English traces even when prompted in another language, with Chinese traces also appearing but less frequently. Prompt hacking can reduce this misalignment by shifting the distribution toward the target language.
\begin{table*}[htbp]
\centering
\resizebox{\textwidth}{!}{
\begin{tabular}{llccccccccccccccc}
\toprule
Method & Model & bn & de & en & es & fr & ja & ru & sw & te & th & zh \\
\midrule
\multirow{8}{*}{Explicit Instruction} 
 & R1-Qwen-1.5B & .01/.00 & .89/.29 & .89/.38 & .88/.27 & .78/.12 & .72/.43 & .92/.42 & .18/.05 & .02/.00 & .71/.82 & .89/.56 \\
 & R1-Qwen-7B   & .60/.04 & .96/.32 & .93/.37 & .93/.28 & .97/.15 & .81/.56 & .97/.47 & .23/.07 & .34/.03 & .47/.34 & .86/.55 \\
 & R1-Qwen-14B  & .71/.06 & .89/.29 & .89/.38 & .92/.28 & .98/.14 & .85/.73 & .97/.48 & .42/.12 & .02/.00 & .84/.50 & .90/.63 \\
 & R1-Qwen-32B  & .78/.05 & .97/.31 & .90/.38 & .95/.28 & .97/.15 & .87/.70 & .98/.48 & .84/.28 & .94/.08 & .85/.52 & .92/.58 \\
 & Qwen-14B     & .01/.00 & .02/.01 & .87/.38 & .02/.01 & .02/.01 & .01/.02 & .97/.47 & .04/.02 & .02/.00 & .00/.02 & .81/.66 \\
 & Qwen-32B     & .01/.00 & .01/.01 & .86/.38 & .01/.01 & .02/.00 & .01/.02 & .96/.47 & .02/.01 & .78/.07 & .01/.01 & .76/.63 \\
 & R1-Llama-8B  & .45/.05 & .93/.31 & .93/.37 & .93/.28 & .97/.15 & .85/.63 & .97/.48 & .92/.32 & .02/.00 & .84/.56 & .90/.56 \\
 & R1-Llama-70B & .80/.06 & .92/.31 & .92/.38 & .93/.28 & .96/.14 & .69/.53 & .97/.49 & .90/.24 & .91/.07 & .82/.52 & .81/.67 \\
\midrule

\multirow{8}{*}{Prompt Hacking}
 & R1-Qwen-1.5B & .86/.06 & .83/.32 & .91/.39 & .82/.28 & .87/.13 & .65/.13 & .94/.42 & .41/.03 & .90/.08 & .97/.58 & .86/.56 \\
 & R1-Qwen-7B   & .92/.07 & .96/.36 & .94/.39 & .87/.30 & .97/.15 & .84/.27 & .94/.43 & .59/.34 & .96/.08 & .82/.45 & .89/.61 \\
 & R1-Qwen-14B  & .85/.07 & .92/.33 & .95/.40 & .67/.21 & .91/.14 & .79/.63 & .96/.47 & .88/.30 & .82/.07 & .81/.52 & .80/.58 \\
 & R1-Qwen-32B  & .83/.06 & .47/.16 & .92/.39 & .84/.27 & .82/.13 & .86/.68 & .97/.47 & .88/.31 & .92/.10 & .84/.56 & .82/.64 \\
 & Qwen-14B     & .89/.05 & .03/.01 & .90/.39 & .82/.26 & .78/.11 & .87/.52 & .96/.42 & .87/.24 & .92/.06 & .88/.51 & .88/.48 \\
 & Qwen-32B     & .81/.04 & .35/.20 & .89/.38 & .33/.14 & .31/.06 & .58/.35 & .92/.43 & .80/.23 & .78/.07 & .70/.47 & .85/.40 \\
 & R1-Llama-8B  & .75/.12 & .90/.33 & .91/.39 & .88/.30 & .93/.14 & .83/.59 & .89/.45 & .67/.38 & .46/.05 & .88/.57 & .83/.55 \\
 & R1-Llama-70B & .86/.06 & .70/.29 & .92/.39 & .71/.26 & .80/.13 & .90/.59 & .96/.49 & .86/.26 & .94/.07 & .84/.53 & .85/.61 \\
\bottomrule
\end{tabular}}
\caption{Sentence/Token Compliance Rate on \textbf{MGSM} across 11 languages.}
\label{tab:compliance_math}
\end{table*}

\begin{table*}[htbp]
\centering
\resizebox{\textwidth}{!}{%
\begin{tabular}{l l ccccccccccccccc}
\toprule
Method & Model & ar & bn & de & es & fr & hi & id & it & ja & ko & pt & sw & yo & zh & en \\
\midrule
\multirow{8}{*}{Explicit Instruction} 
&R1-Qwen-1.5B & .07/.05 & .03/.01 & .24/.08 & .89/.29 & .44/.09 & .05/.06 & .20/.12 & .51/.28 & .14/.16 & .02/.03 & .81/.32 & .40/.34 & .12/.02 & .85/.63 & .95/.45 \\
&R1-Qwen-7B & .69/.38 & .20/.04 & .95/.37 & .96/.30 & .91/.16 & .78/.36 & .92/.32 & .89/.44 & .83/.26 & .16/.10 & .97/.38 & .07/.03 & .16/.03 & .87/.67 & .97/.46 \\
&R1-Qwen-14B & .02/.01 & .02/.01 & .17/.07 & .48/.18 & .17/.03 & .04/.14 & .17/.05 & .24/.12 & .32/.39 & .05/.04 & .12/.05 & .08/.02 & .09/.02 & .85/.69 & .97/.46 \\
&R1-Qwen-32B & .19/.13 & .17/.05 & .87/.38 & .38/.15 & .05/.01 & .19/.27 & .05/.01 & .50/.26 & .37/.37 & .15/.11 & .08/.04 & .40/.17 & .20/.04 & .88/.68 & .96/.45 \\
&Qwen-14B & .00/.00 & .00/.00 & .01/.01 & .01/.01 & .01/.00 & .00/.00 & .01/.00 & .00/.01 & .00/.00 & .00/.01 & .01/.01 & .02/.01 & .07/.02 & .67/.65 & .96/.46 \\
&Qwen-32B & .00/.00 & .00/.00 & .01/.01 & .01/.01 & .01/.00 & .00/.00 & .02/.01 & .01/.01 & .00/.00 & .00/.01 & .02/.01 & .02/.01 & .05/.01 & .82/.69 & .95/.46 \\
&R1-Llama-8B & .52/.33 & .01/.00 & .29/.12 & .53/.20 & .80/.14 & .02/.03 & .20/.06 & .25/.15 & .57/.50 & .06/.06 & .60/.27 & .21/.08 & .15/.02 & .92/.69 & .96/.46 \\
&R1-Llama-70B & .44/.27 & .01/.01 & .05/.02 & .09/.04 & .10/.02 & .04/.11 & .12/.03 & .02/.02 & .29/.20 & .03/.02 & .05/.02 & .06/.01 & .09/.01 & .86/.68 & .95/.46 \\

\midrule
\multirow{8}{*}{Prompt Hacking} 
&R1-Qwen-1.5B & .75/.38 & .97/.07 & .82/.34 & .90/.29 & .96/.16 & .86/.35 & .69/.33 & .94/.47 & .56/.31 & .40/.35 & .65/.38 & .66/.58 & .73/.11 & .89/.56 & .97/.47 \\
&R1-Qwen-7B & .66/.37 & .92/.07 & .95/.39 & .96/.32 & .93/.16 & .84/.36 & .91/.32 & .96/.48 & .74/.25 & .61/.32 & .97/.41 & .76/.56 & .97/.10 & .91/.61 & .97/.47 \\
&R1-Qwen-14B & .73/.43 & .94/.08 & .94/.40 & .95/.32 & .97/.17 & .77/.35 & .98/.33 & .97/.49 & .93/.74 & .97/.86 & .98/.41 & .95/.36 & .75/.16 & .92/.62 & .97/.47 \\
&R1-Qwen-32B & .78/.44 & .91/.07 & .97/.40 & .96/.32 & .97/.17 & .80/.35 & .98/.32 & .96/.49 & .73/.68 & .97/.85 & .98/.42 & .94/.40 & .94/.26 & .85/.62 & .96/.46 \\
&Qwen-14B & .78/.52 & .90/.08 & .89/.38 & .96/.31 & .93/.16 & .76/.34 & .97/.32 & .96/.49 & .86/.71 & .98/.90 & .97/.41 & .96/.35 & .91/.18 & .90/.50 & .96/.46 \\
&Qwen-32B & .70/.49 & .85/.06 & .59/.34 & .42/.18 & .50/.10 & .67/.34 & .80/.27 & .40/.25 & .73/.61 & .86/.80 & .61/.34 & .91/.33 & .91/.19 & .88/.58 & .97/.48 \\
&R1-Llama-8B & .88/.55 & .81/.11 & .87/.40 & .95/.32 & .94/.16 & .87/.35 & .98/.32 & .95/.48 & .89/.65 & .87/.80 & .95/.42 & .89/.38 & .85/.18 & .78/.58 & .94/.46 \\
&R1-Llama-70B & .83/.55 & .91/.07 & .81/.38 & .78/.28 & .92/.17 & .73/.35 & .95/.28 & .88/.46 & .85/.61 & .77/.69 & .94/.43 & .90/.33 & .94/.18 & .87/.59 & .95/.47 \\

\bottomrule
\end{tabular}}
\caption{Compliance rates (sentence/token) for \textbf{MMMLU}, across 15 languages.}
\label{tab:compliance_mcq}
\end{table*}

\begin{sidewaystable*}[htbp]
\centering
\resizebox{\textwidth}{!}{%
\begin{tabular}{l l c c c c c c c c c c c}
\toprule
Model & Setting & bn & de & en & es & fr & ja & ru & sw & te & th & zh \\
\midrule
R1-Qwen-1.5B & Explicit & .86/.37 | .00/.00 & .03/.06 | .00/.00 & .89/.38 | .00/.00 & .00/.02 | .00/.00 & .11/.09 | .00/.00 & .02/.06 | .06/.05 & .02/.04 | .00/.00 & .73/.35 | .00/.00 & .87/.38 | .00/.00 & .20/.03 | .04/.00 & .00/.00 | .89/.56 \\
R1-Qwen-1.5B & Hacking & .03/.00 | .00/.00 & .01/.05 | .00/.00 & .91/.39 | .00/.00 & .01/.03 | .00/.00 & .00/.06 | .00/.00 & .14/.40 | .05/.01 & .04/.03 | .00/.00 & .43/.08 | .00/.00 & .01/.00 | .00/.00 & .01/.01 | .00/.00 & .00/.00 | .86/.56 \\
\midrule
R1-Qwen-7B & Explicit & .20/.04 | .00/.00 & .00/.05 | .00/.00 & .93/.37 | .00/.00 & .00/.02 | .00/.00 & .00/.05 | .00/.00 & .05/.04 | .01/.01 & .02/.02 | .00/.00 & .58/.28 | .00/.00 & .54/.20 | .00/.00 & .36/.11 | .01/.01 & .00/.00 | .86/.55 \\
R1-Qwen-7B & Hacking & .00/.00 | .00/.00 & .00/.05 | .00/.00 & .94/.39 | .00/.00 & .01/.02 | .00/.00 & .00/.05 | .00/.00 & .07/.12 | .01/.01 & .01/.01 | .00/.00 & .02/.04 | .01/.00 & .00/.00 | .00/.00 & .03/.02 | .00/.01 & .00/.01 | .89/.61 \\
\midrule
R1-Qwen-14B & Explicit & .18/.02 | .00/.00 & .08/.07 | .00/.00 & .89/.38 | .00/.00 & .00/.01 | .00/.00 & .00/.05 | .00/.00 & .05/.01 | .00/.01 & .00/.00 | .00/.00 & .44/.20 | .01/.01 & .85/.36 | .00/.00 & .00/.00 | .00/.00 & .00/.00 | .90/.63 \\
R1-Qwen-14B & Hacking & .00/.00 | .00/.00 & .01/.04 | .00/.00 & .95/.40 | .00/.00 & .01/.01 | .21/.17 & .00/.05 | .02/.02 & .12/.05 | .01/.01 & .00/.00 | .00/.00 & .01/.00 | .00/.00 & .03/.01 | .01/.01 & .00/.00 | .00/.00 & .00/.00 | .80/.58 \\
\midrule
R1-Qwen-32B & Explicit & .00/.00 | .00/.00 & .00/.04 | .00/.00 & .90/.38 | .00/.00 & .00/.01 | .00/.00 & .00/.05 | .00/.00 & .00/.00 | .00/.01 & .00/.00 | .00/.00 & .01/.01 | .00/.00 & .00/.00 | .00/.00 & .01/.00 | .00/.00 & .00/.00 | .92/.58 \\
R1-Qwen-32B & Hacking & .01/.00 | .00/.00 & .06/.05 | .34/.31 & .92/.39 | .00/.00 & .03/.02 | .02/.02 & .03/.05 | .07/.07 & .01/.01 | .01/.02 & .00/.00 | .00/.00 & .00/.01 | .00/.00 & .00/.00 | .00/.00 & .01/.00 | .01/.00 & .00/.00 | .82/.64 \\
\midrule
Qwen-14B & Explicit & .87/.37 | .00/.00 & .86/.36 | .00/.00 & .87/.38 | .00/.00 & .86/.36 | .00/.00 & .86/.36 | .00/.00 & .87/.38 | .00/.00 & .00/.00 | .00/.00 & .85/.37 | .00/.00 & .85/.36 | .00/.00 & .88/.37 | .00/.00 & .00/.00 | .81/.66 \\
Qwen-14B & Hacking & .00/.01 | .00/.00 & .89/.38 | .00/.00 & .90/.39 | .00/.00 & .03/.03 | .00/.00 & .15/.10 | .00/.00 & .00/.03 | .02/.02 & .00/.01 | .00/.00 & .00/.01 | .00/.00 & .01/.01 | .00/.00 & .01/.01 | .00/.00 & .00/.00 | .88/.48 \\
\midrule
Qwen-32B & Explicit & .86/.37 | .00/.00 & .85/.36 | .00/.00 & .86/.38 | .00/.00 & .86/.36 | .00/.00 & .84/.36 | .00/.00 & .87/.37 | .00/.00 & .01/.00 | .00/.00 & .86/.37 | .00/.00 & .12/.03 | .00/.00 & .86/.37 | .00/.00 & .00/.00 | .76/.63 \\
Qwen-32B & Hacking & .00/.01 | .00/.00 & .42/.18 | .00/.00 & .89/.38 | .00/.00 & .51/.20 | .00/.00 & .55/.23 | .01/.01 & .29/.15 | .01/.01 & .02/.02 | .00/.00 & .06/.04 | .00/.00 & .12/.03 | .00/.00 & .17/.05 | .00/.00 & .01/.01 | .85/.40 \\
\midrule
R1-Llama-8B & Explicit & .30/.06 | .00/.00 & .02/.04 | .00/.00 & .93/.37 | .00/.00 & .02/.02 | .00/.00 & .00/.05 | .00/.00 & .01/.00 | .03/.04 & .01/.01 | .00/.00 & .01/.01 | .00/.00 & .90/.39 | .00/.00 & .01/.00 | .02/.01 & .00/.00 | .90/.56 \\
R1-Llama-8B & Hacking & .06/.01 | .00/.00 & .03/.05 | .00/.00 & .91/.39 | .00/.00 & .02/.02 | .00/.00 & .02/.06 | .00/.00 & .01/.01 | .02/.04 & .00/.01 | .00/.00 & .01/.01 | .01/.00 & .37/.13 | .00/.00 & .00/.00 | .00/.00 & .00/.00 | .83/.55 \\
\midrule
R1-Llama-70B & Explicit & .01/.00 | .00/.00 & .05/.05 | .00/.00 & .92/.38 | .00/.00 & .00/.02 | .00/.00 & .02/.06 | .00/.00 & .00/.00 | .15/.16 & .00/.00 | .00/.00 & .00/.00 | .00/.00 & .02/.00 | .00/.00 & .00/.00 | .02/.01 & .00/.00 | .81/.67 \\
R1-Llama-70B & Hacking & .01/.00 | .00/.00 & .05/.05 | .13/.07 & .92/.39 | .00/.00 & .11/.05 | .05/.03 & .11/.08 | .00/.00 & .00/.02 | .01/.02 & .00/.00 | .00/.00 & .00/.01 | .01/.00 & .01/.00 | .00/.00 & .00/.01 | .01/.00 & .00/.00 | .85/.61 \\
\bottomrule
\end{tabular}}
\caption{English and Chinese compliance rates (English-sentence-level / English-token-level | Chinese-sentence-level / Chinese-token-level) on \textbf{MGSM}. Explicit instruction vs. Prompt hacking.}
\label{tab:compliance_en_math}
\vspace{1em}
\resizebox{\textwidth}{!}{%
\begin{tabular}{l l c c c c c c c c c c c c c c c}
\toprule
Model & Setting & ar & bn & de & en & es & fr & hi & id & it & ja & ko & pt & sw & yo & zh \\
\midrule
R1-Qwen-1.5B & Explicit & .42/.90 | .00/.00 & .41/.93 | .00/.00 & .36/.72 | .00/.00 & .45/.95 | .00/.00 & .05/.07 | .00/.00 & .25/.38 | .00/.00 & .38/.91 | .01/.00 & .32/.38 | .00/.00 & .20/.43 | .00/.00 & .27/.68 | .12/.10 & .47/.94 | .00/.00 & .12/.15 | .00/.00 & .19/.47 | .00/.00 & .40/.82 | .00/.00 & .63/.85 \\
R1-Qwen-1.5B & Hacking & .12/.11 | .01/.00 & .00/.00 | .00/.00 & .05/.01 | .00/.00 & .47/.97 | .00/.00 & .04/.01 | .00/.00 & .09/.00 | .00/.00 & .00/.00 | .00/.00 & .02/.00 | .00/.00 & .02/.01 | .00/.00 & .21/.32 | .03/.05 & .29/.25 | .00/.01 & .03/.00 | .00/.00 & .00/.00 | .00/.00 & .01/.00 | .00/.00 & .56/.89 \\
\midrule
R1-Qwen-7B & Explicit & .12/.17 | .02/.01 & .16/.76 | .00/.00 & .08/.01 | .00/.00 & .46/.97 | .00/.00 & .06/.00 | .00/.00 & .11/.06 | .00/.00 & .00/.03 | .00/.00 & .05/.06 | .00/.00 & .03/.04 | .00/.00 & .35/.05 | .02/.02 & .36/.69 | .02/.01 & .03/.00 | .00/.00 & .40/.67 | .00/.00 & .40/.79 | .00/.00 & .67/.87 \\
R1-Qwen-7B & Hacking & .12/.18 | .02/.01 & .00/.00 | .00/.00 & .06/.02 | .00/.00 & .47/.97 | .00/.00 & .03/.00 | .00/.00 & .08/.01 | .00/.00 & .00/.00 | .00/.00 & .02/.00 | .00/.00 & .02/.00 | .00/.00 & .29/.16 | .05/.01 & .28/.19 | .02/.01 & .03/.00 | .00/.00 & .01/.00 | .00/.01 & .00/.00 | .00/.00 & .61/.91 \\
\midrule
R1-Qwen-14B & Explicit & .46/.95 | .00/.00 & .42/.96 | .00/.00 & .40/.80 | .00/.00 & .46/.97 | .00/.00 & .23/.50 | .00/.00 & .39/.81 | .00/.00 & .28/.92 | .01/.00 & .40/.82 | .00/.00 & .35/.72 | .00/.00 & .20/.63 | .05/.02 & .45/.92 | .00/.00 & .41/.86 | .01/.00 & .44/.90 | .00/.00 & .41/.85 | .00/.00 & .69/.85 \\
R1-Qwen-14B & Hacking & .03/.04 | .09/.09 & .00/.00 | .00/.00 & .06/.02 | .00/.00 & .47/.97 | .00/.00 & .02/.00 | .00/.00 & .08/.00 | .00/.00 & .00/.01 | .00/.00 & .02/.00 | .00/.00 & .01/.00 | .00/.00 & .00/.00 | .01/.01 & .00/.00 | .02/.01 & .02/.00 | .00/.00 & .01/.01 | .00/.00 & .10/.17 | .00/.00 & .62/.92 \\
\midrule
R1-Qwen-32B & Explicit & .36/.75 | .01/.00 & .15/.80 | .00/.00 & .09/.09 | .00/.00 & .45/.96 | .00/.00 & .27/.60 | .00/.00 & .44/.93 | .00/.00 & .11/.75 | .00/.00 & .44/.92 | .00/.00 & .22/.47 | .00/.00 & .13/.45 | .14/.12 & .41/.82 | .01/.00 & .43/.89 | .00/.00 & .27/.57 | .00/.00 & .36/.73 | .00/.00 & .68/.88 \\
R1-Qwen-32B & Hacking & .03/.05 | .05/.03 & .00/.00 | .00/.00 & .05/.00 | .00/.00 & .46/.96 | .00/.00 & .02/.00 | .00/.00 & .08/.00 | .00/.00 & .00/.00 | .00/.00 & .02/.00 | .00/.00 & .01/.00 | .00/.00 & .01/.00 | .02/.13 & .00/.00 | .03/.02 & .02/.00 | .00/.00 & .01/.00 | .00/.00 & .01/.00 | .00/.00 & .62/.85 \\
\midrule
Qwen-14B & Explicit & .45/.96 | .00/.00 & .45/.97 | .00/.00 & .44/.95 | .00/.00 & .46/.96 | .00/.00 & .45/.95 | .00/.00 & .45/.95 | .00/.00 & .45/.96 | .00/.00 & .46/.95 | .00/.00 & .45/.96 | .00/.00 & .46/.95 | .00/.00 & .45/.96 | .00/.00 & .45/.95 | .00/.00 & .44/.94 | .00/.00 & .41/.89 | .00/.00 & .65/.67 \\
Qwen-14B & Hacking & .04/.10 | .02/.01 & .00/.00 | .00/.00 & .09/.09 | .00/.00 & .46/.96 | .00/.00 & .02/.00 | .00/.00 & .09/.05 | .00/.00 & .00/.00 | .00/.00 & .02/.01 | .00/.00 & .02/.02 | .00/.00 & .01/.00 | .02/.01 & .00/.00 | .00/.00 & .02/.00 | .00/.00 & .00/.00 | .00/.00 & .00/.00 | .00/.00 & .50/.90 \\
\midrule
Qwen-32B & Explicit & .47/.98 | .00/.00 & .47/.97 | .00/.00 & .45/.95 | .00/.00 & .46/.95 | .00/.00 & .45/.96 | .00/.00 & .45/.94 | .00/.00 & .47/.97 | .00/.00 & .46/.95 | .00/.00 & .46/.96 | .00/.00 & .47/.96 | .00/.00 & .47/.96 | .00/.00 & .46/.95 | .00/.00 & .47/.93 | .00/.00 & .46/.92 | .00/.00 & .69/.82 \\
Qwen-32B & Hacking & .04/.08 | .02/.02 & .01/.04 | .00/.00 & .11/.16 | .00/.00 & .48/.97 | .00/.00 & .20/.41 | .01/.01 & .20/.40 | .01/.01 & .01/.08 | .00/.04 & .04/.03 | .00/.00 & .23/.50 | .00/.00 & .06/.16 | .02/.01 & .04/.04 | .02/.02 & .11/.21 | .00/.00 & .02/.01 | .00/.01 & .01/.00 | .00/.00 & .58/.88 \\
\midrule
R1-Llama-8B & Explicit & .19/.40 | .03/.01 & .46/.97 | .00/.00 & .34/.68 | .00/.00 & .46/.96 | .00/.00 & .20/.44 | .00/.00 & .14/.18 | .00/.00 & .41/.94 | .04/.02 & .40/.79 | .00/.00 & .34/.72 | .00/.00 & .09/.30 | .08/.06 & .40/.88 | .07/.04 & .19/.38 | .00/.00 & .38/.76 | .00/.00 & .42/.81 | .00/.00 & .69/.92 \\
R1-Llama-8B & Hacking & .01/.01 | .01/.00 & .00/.10 | .00/.00 & .05/.02 | .00/.00 & .46/.94 | .00/.00 & .02/.00 | .00/.00 & .09/.03 | .00/.00 & .00/.00 | .00/.00 & .02/.00 | .00/.00 & .02/.01 | .00/.00 & .01/.01 | .04/.02 & .02/.02 | .01/.01 & .03/.00 | .00/.00 & .01/.00 | .00/.01 & .08/.13 | .00/.00 & .58/.78 \\
\midrule
R1-Llama-70B & Explicit & .22/.45 | .06/.03 & .44/.96 | .00/.00 & .44/.92 | .00/.00 & .46/.95 | .00/.00 & .42/.89 | .00/.00 & .43/.87 | .00/.00 & .33/.94 | .00/.00 & .43/.86 | .00/.00 & .46/.95 | .00/.00 & .03/.07 | .48/.51 & .46/.94 | .02/.01 & .45/.92 | .00/.00 & .46/.90 | .00/.00 & .43/.89 | .00/.00 & .68/.86 \\
R1-Llama-70B & Hacking & .02/.02 | .02/.02 & .00/.00 | .00/.00 & .06/.03 | .02/.06 & .47/.95 | .00/.00 & .06/.10 | .02/.02 & .09/.02 | .00/.00 & .00/.00 | .00/.00 & .02/.00 | .00/.00 & .03/.04 | .01/.00 & .01/.00 | .05/.02 & .00/.00 | .13/.16 & .03/.00 | .00/.00 & .01/.00 | .00/.01 & .01/.01 | .00/.00 & .59/.87 \\
\bottomrule
\end{tabular}}
\caption{English and Chinese compliance rates (English-sentence-level / English-token-level | Chinese-sentence-level / Chinese-token-level) on \textbf{MMMLU}. Explicit instruction vs. Prompt hacking.}
\label{tab:compliance_en_mcq}
\end{sidewaystable*}

\subsection{Complete Results for Interchanging Thinking traces}
\begin{figure*}[htbp]
    \centering
    \includegraphics[width=0.44\textwidth]{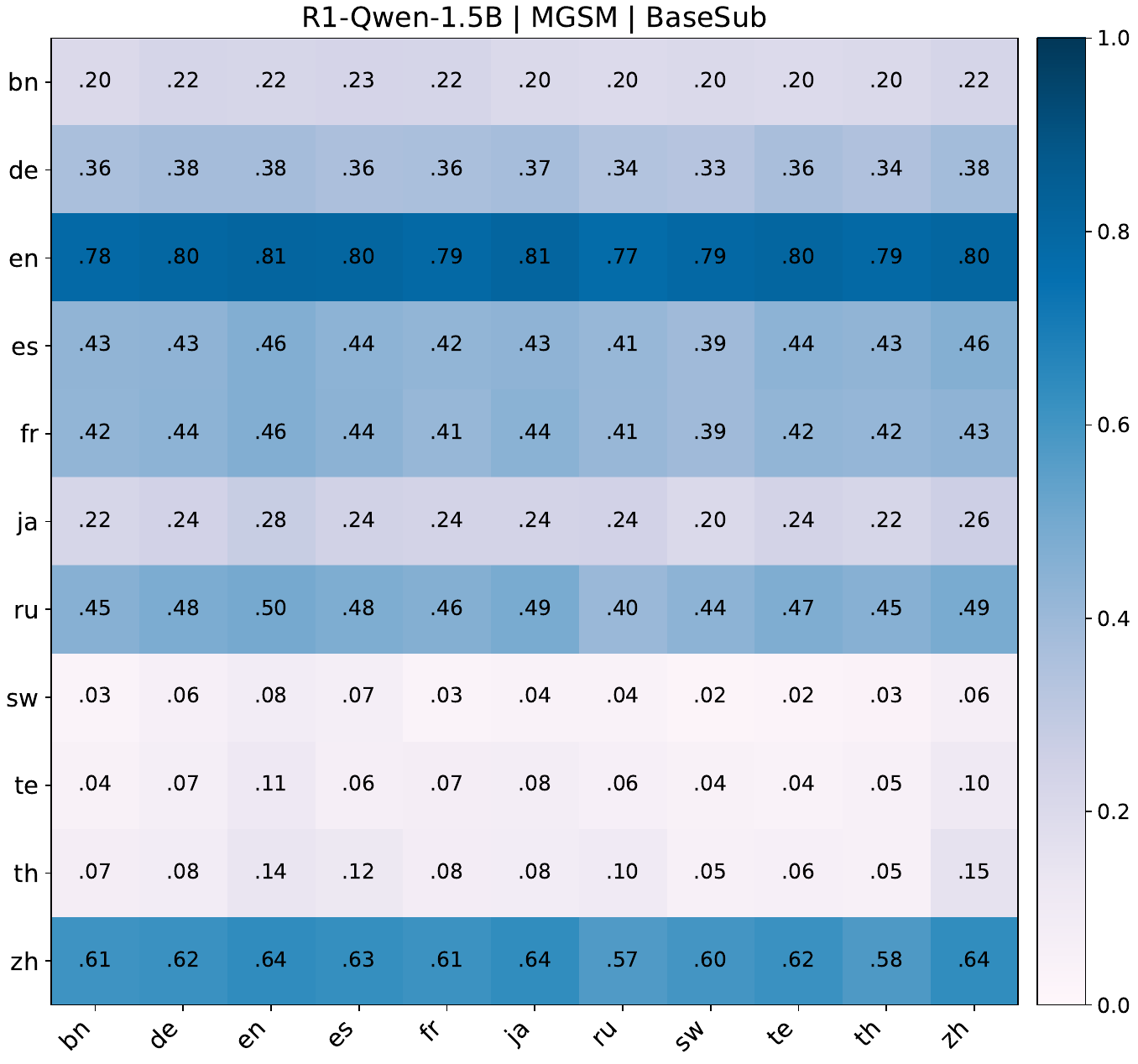}
    \includegraphics[width=0.44\textwidth]{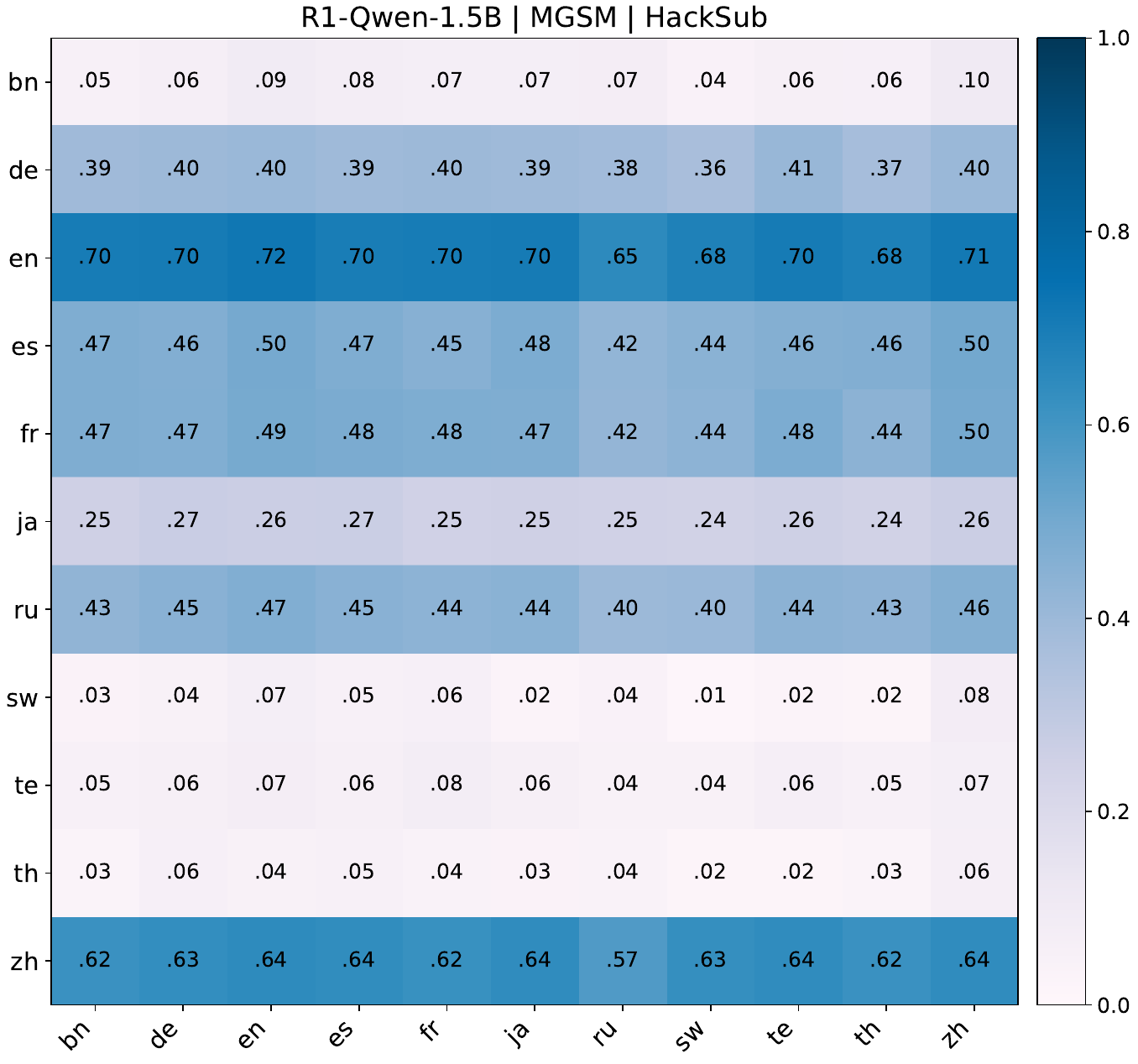}
    
    \includegraphics[width=0.44\textwidth]{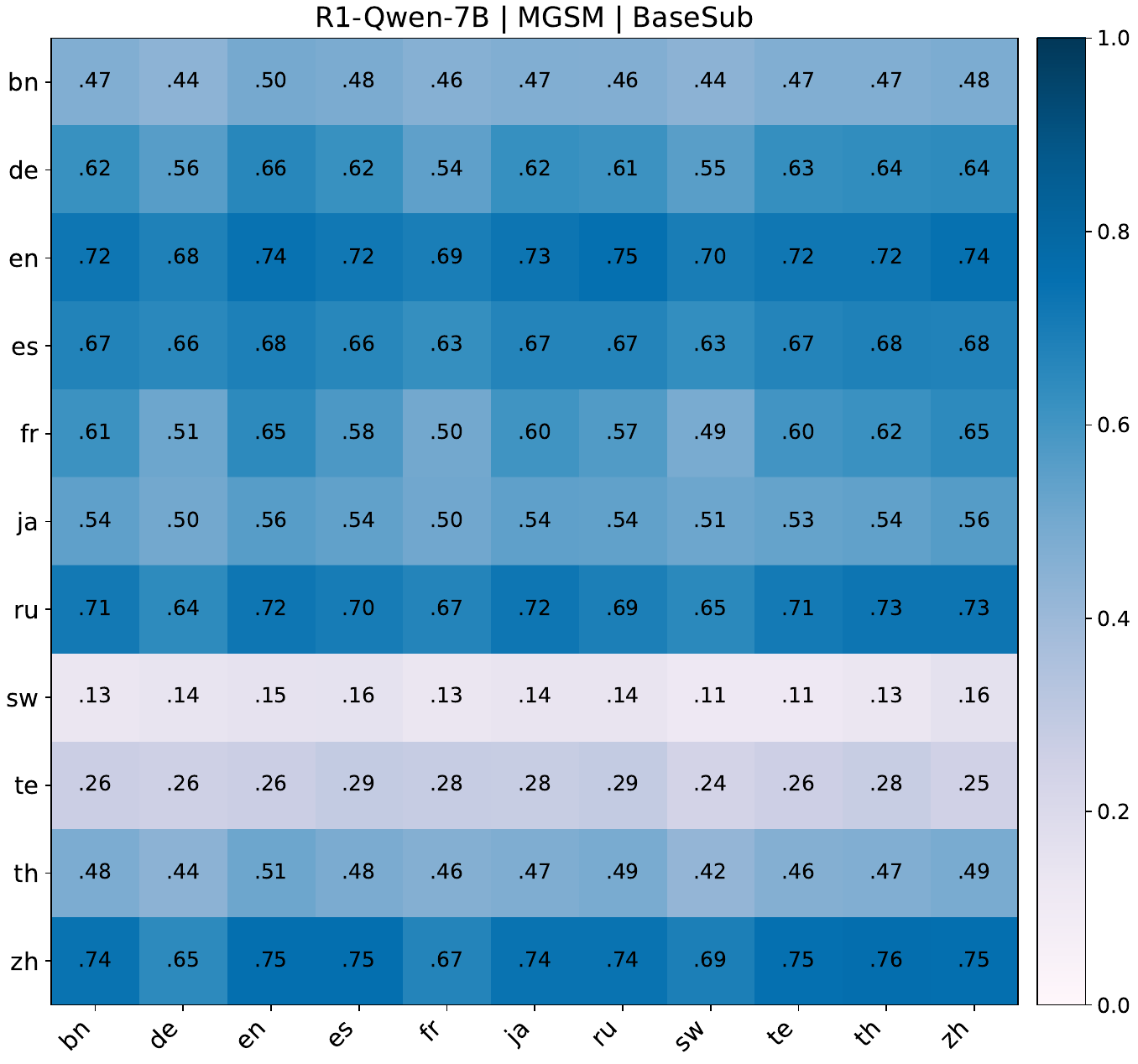}
    \includegraphics[width=0.44\textwidth]{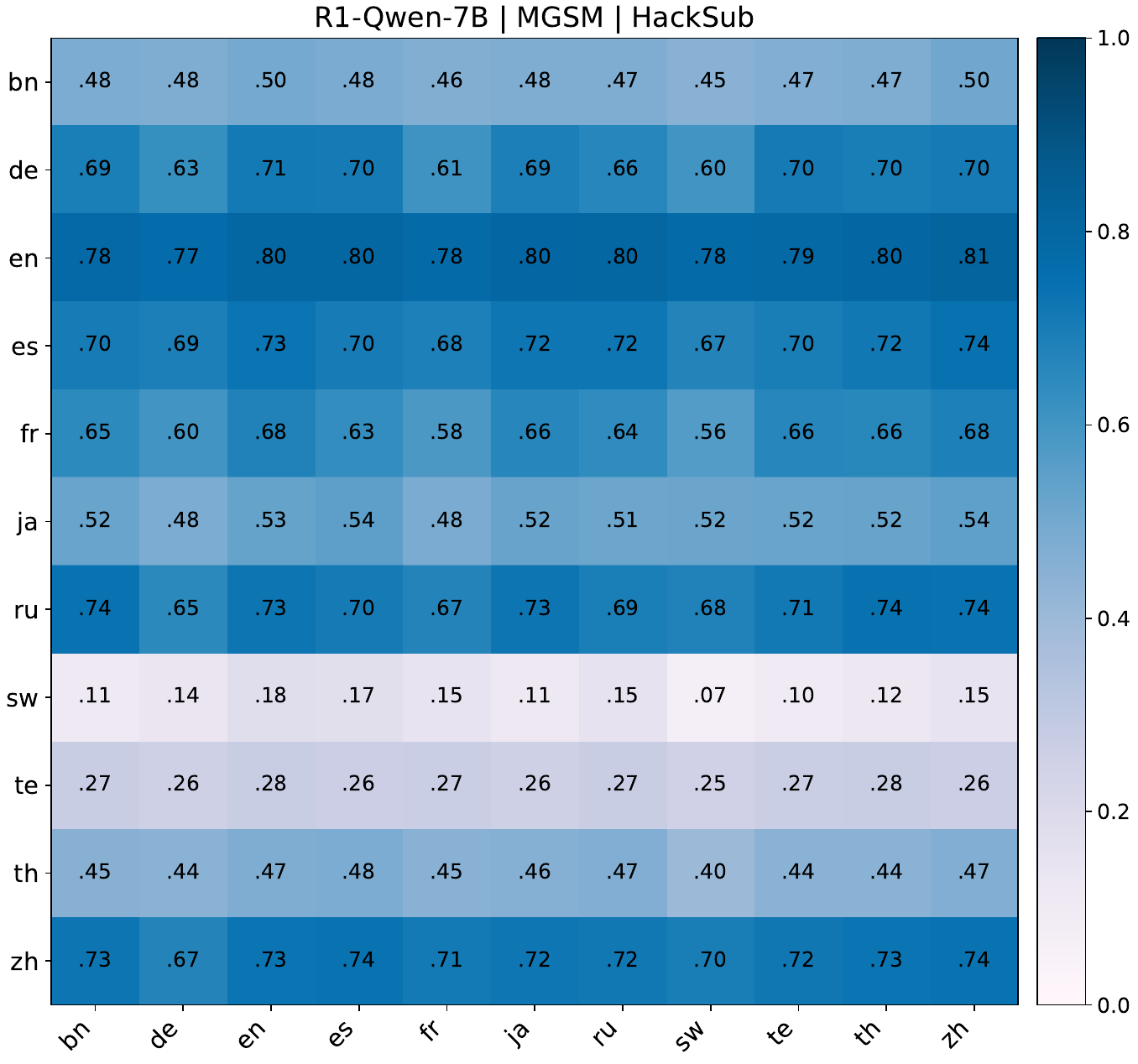}
    
    \includegraphics[width=0.44\textwidth]{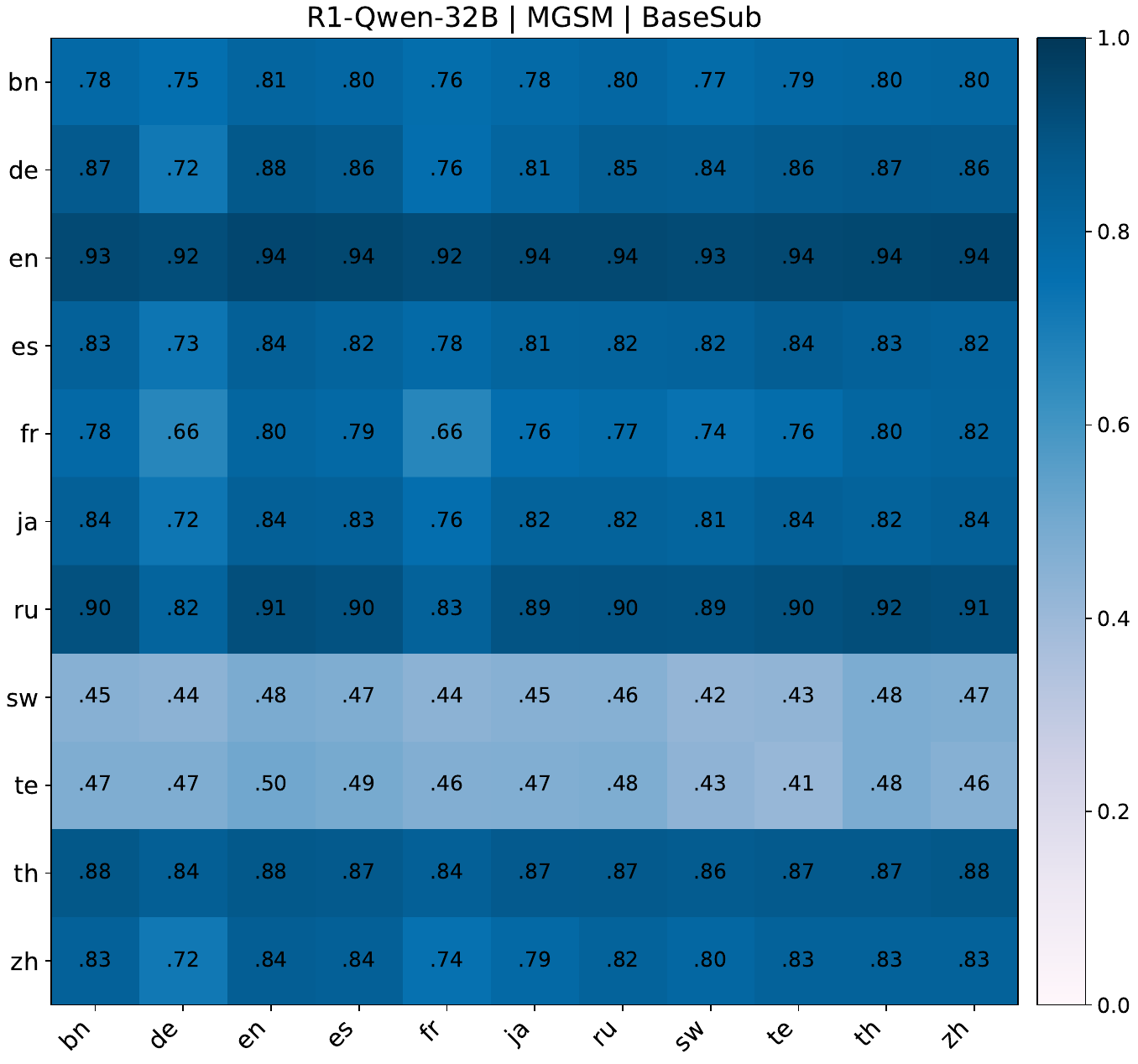}
    \includegraphics[width=0.44\textwidth]{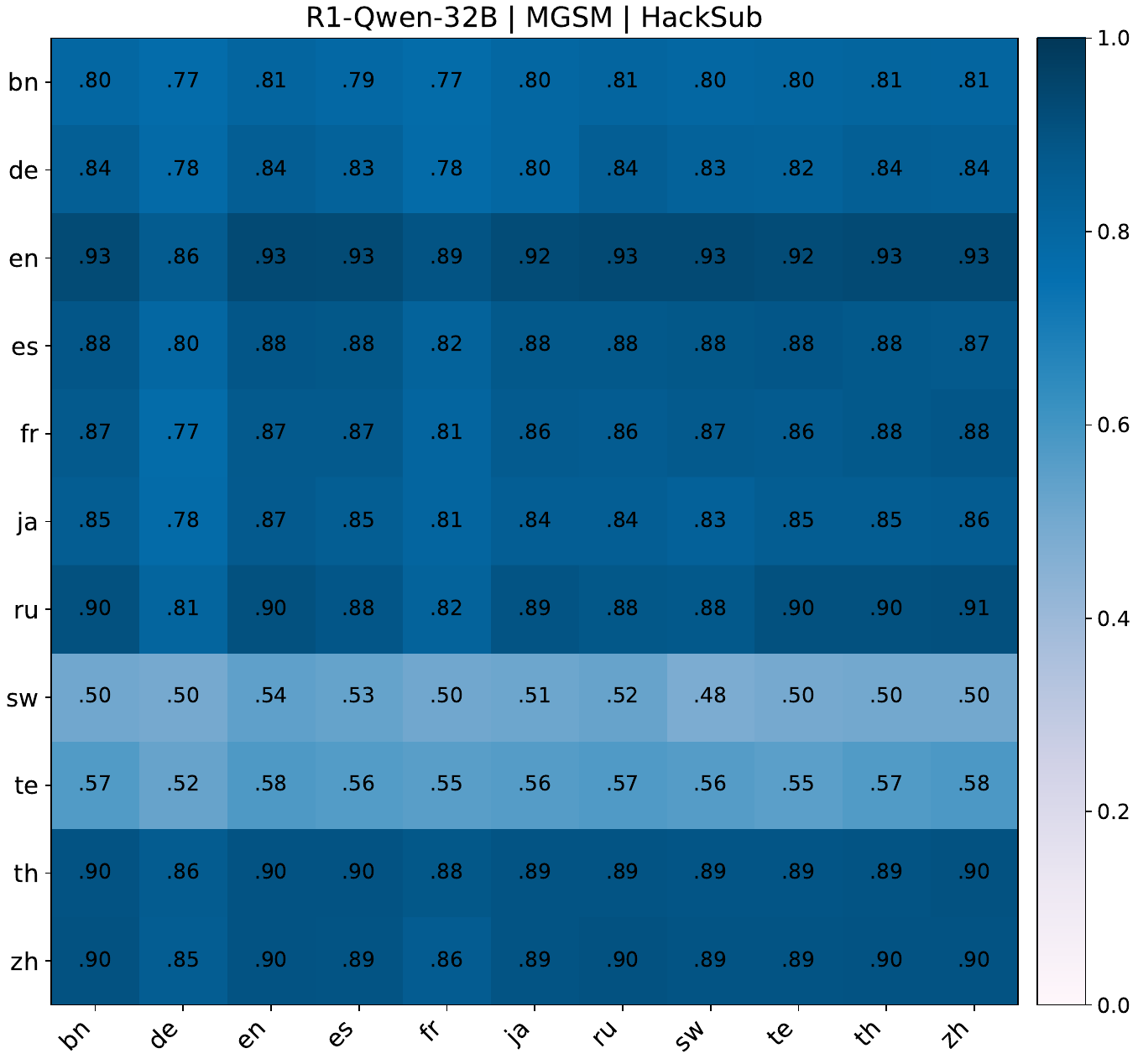}

    \caption{Final-answer accuracy of LRMs under two thinking trace substitutions: \texttt{BaseSub}, \texttt{HackSub}. Each cell shows the accuracy when injecting thinking traces from a language on the y-axis into a language on the y-axis. Performance disparities indicate that thinking trace quality varies across languages.}
    \label{fig:swap_acc}
\end{figure*}

\begin{figure*}[htbp]
    \centering
    \includegraphics[width=0.44\textwidth]{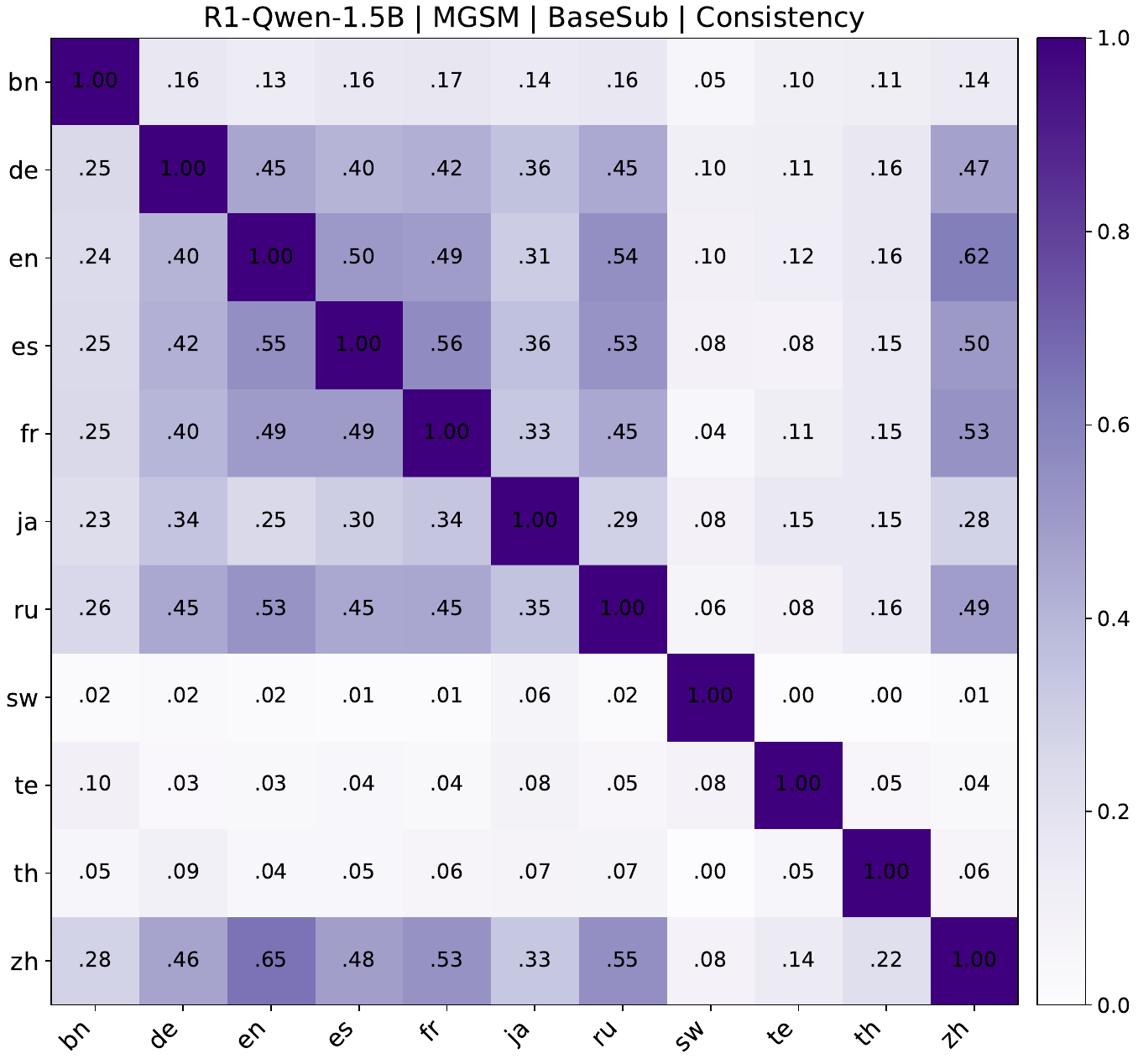}
    \includegraphics[width=0.44\textwidth]{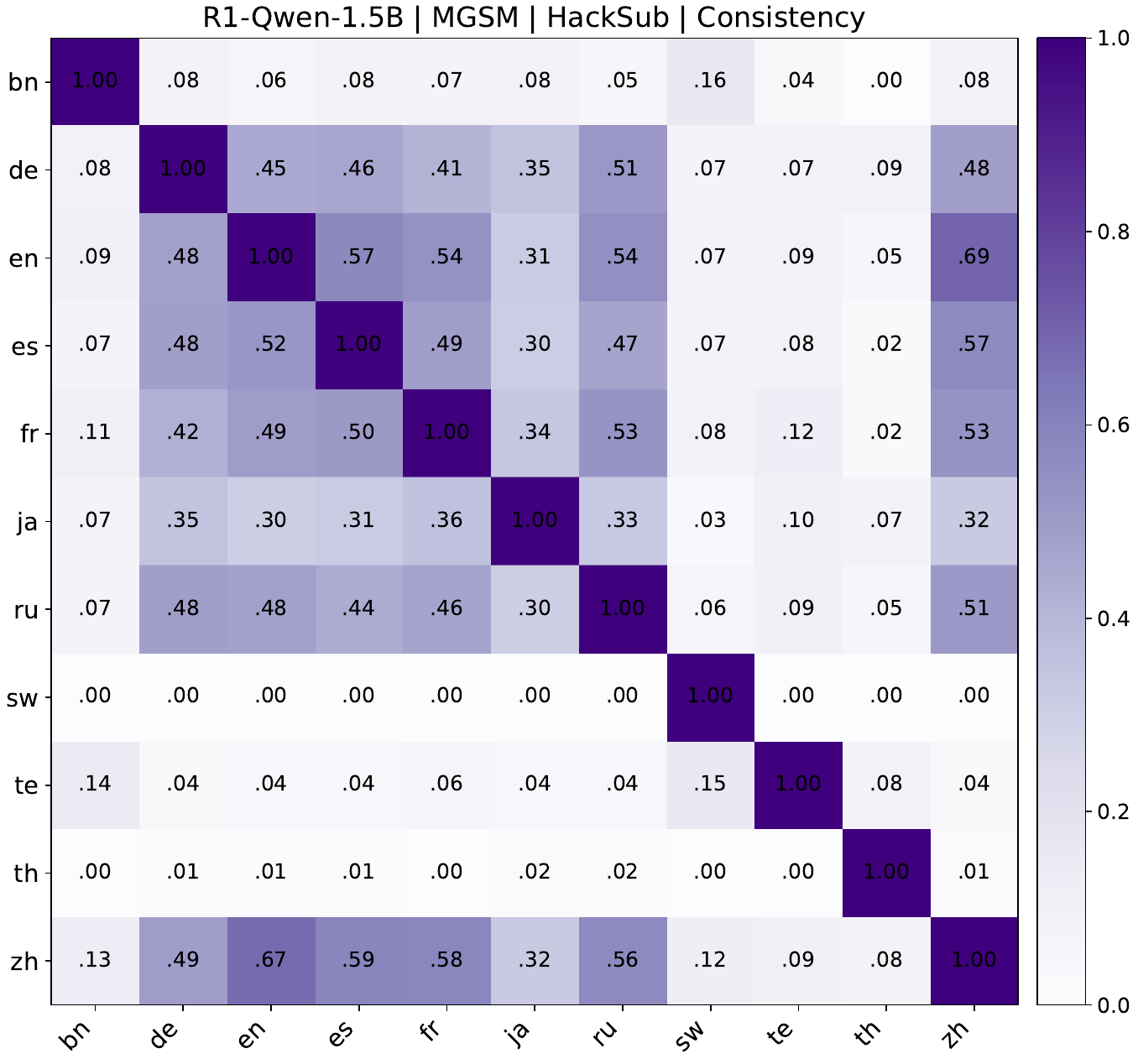}

    \includegraphics[width=0.44\textwidth]{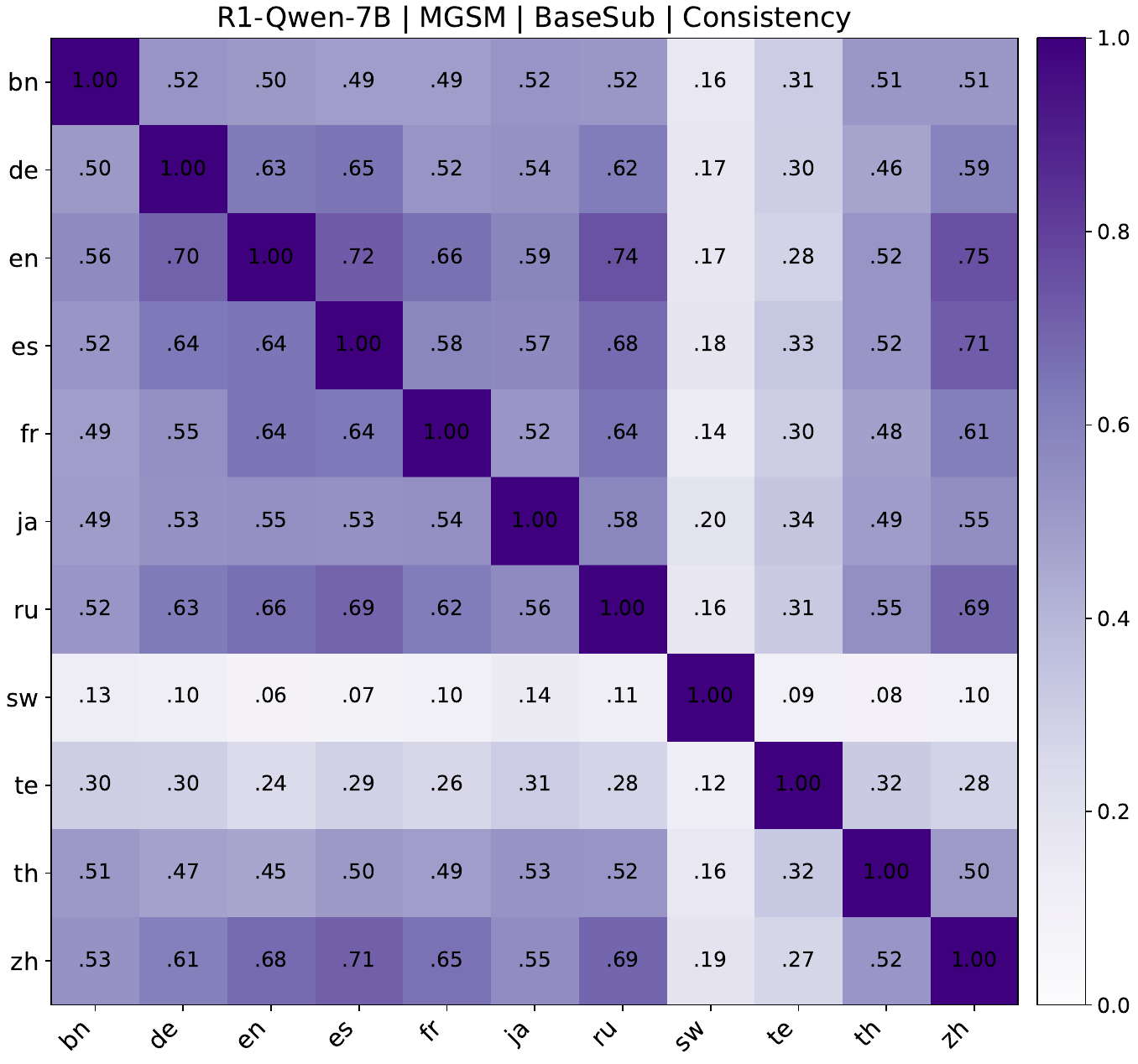}
    \includegraphics[width=0.44\textwidth]{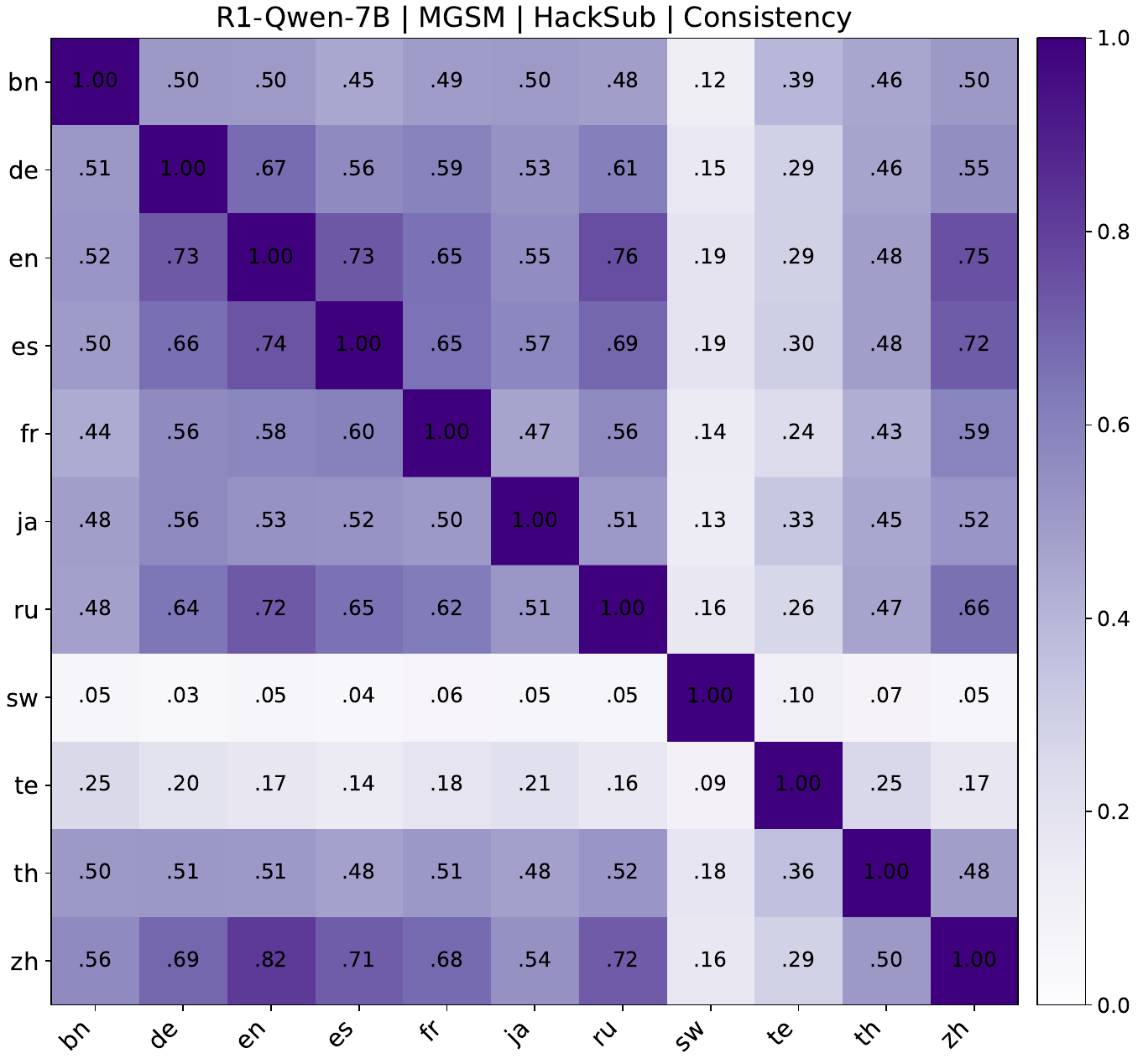}

    \includegraphics[width=0.44\textwidth]{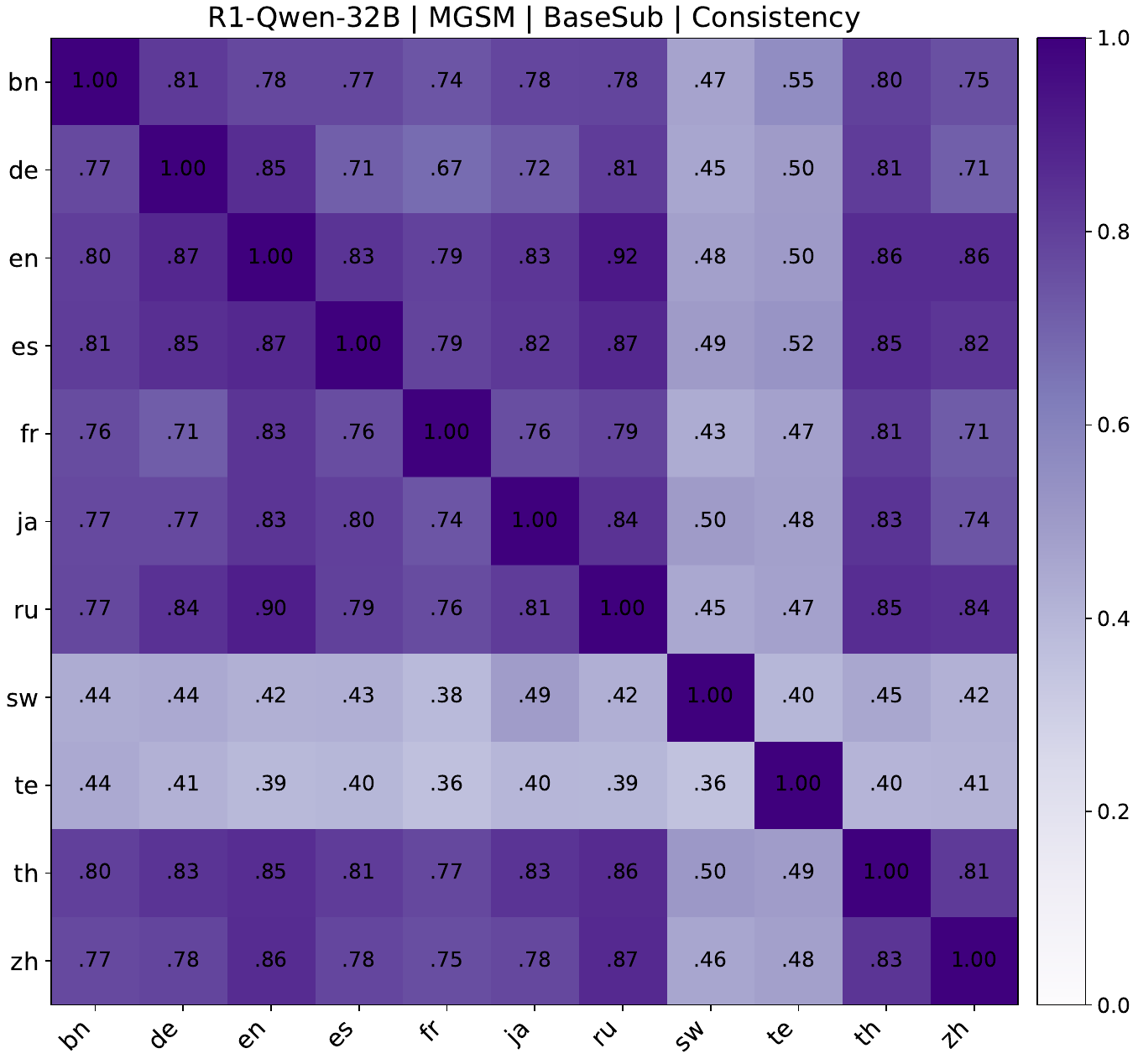}
    \includegraphics[width=0.44\textwidth]{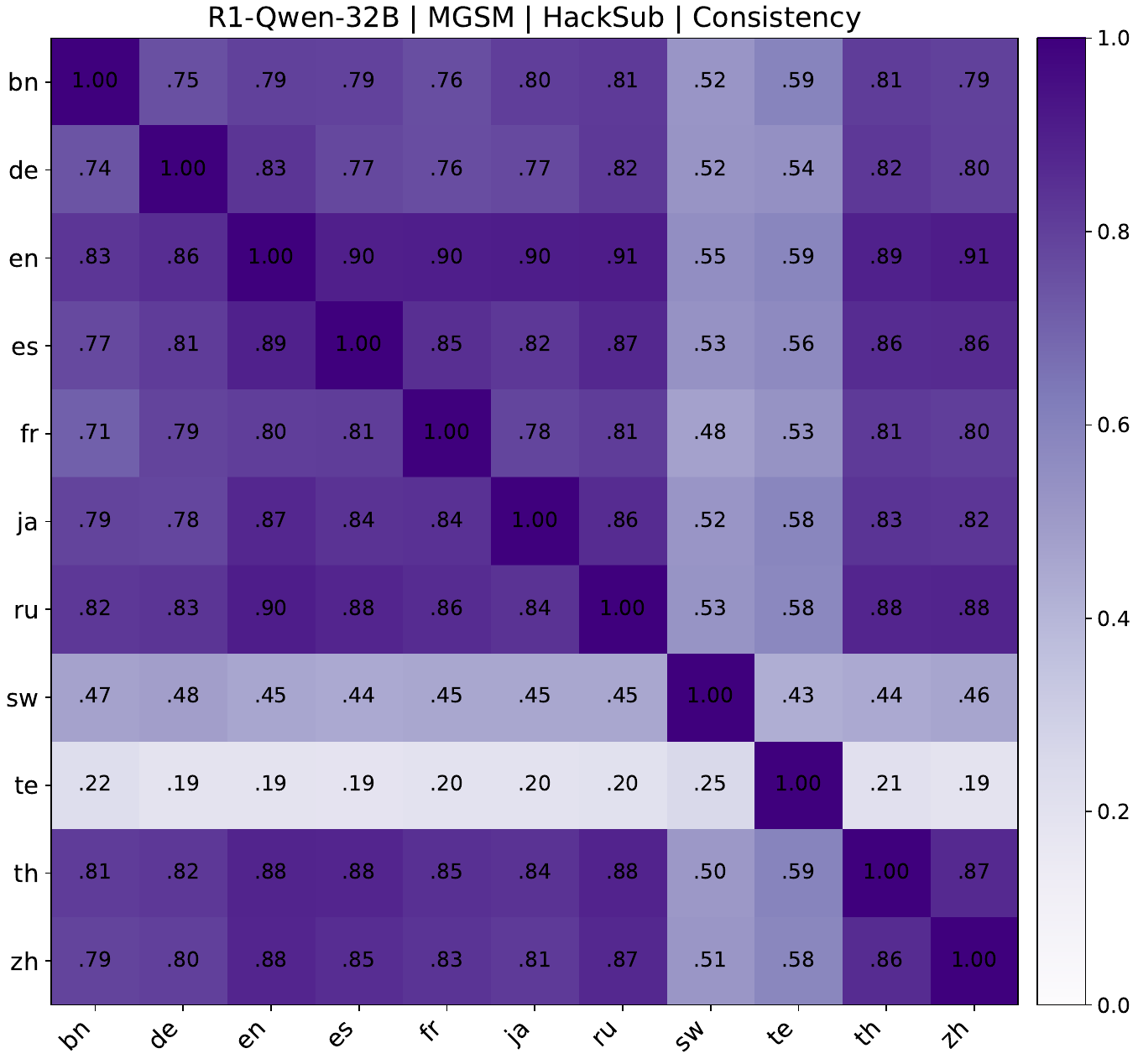}
    \caption{Substitution consistency of LRMs under two thinking trace substitutions: \texttt{BaseSub}, \texttt{HackSub}. Each cell indicates the consistency between the original predictions in the language on the x-axis and the predictions after injecting thinking traces from the language on the y-axis. Higher consistency is observed when traces are substituted between similar languages.}
    \label{fig:swap_cons}
\end{figure*}

\seclabel{complete_consistency}
Figure~\ref{fig:swap_acc} shows the accuracy of R1-Qwen models on MGSM with interchanged reasoning traces. Injecting traces from low-resource languages into high-resource prompts lowers performance, while high-resource traces can boost low-resource prompts. This confirms the strong influence of reasoning trace quality on final accuracy. 
Figure~\ref{fig:swap_cons} shows that consistency with original setup after substitution is generally higher between similar languages. HackSub follows the same trend as BaseSub but sometimes lowers cross-language consistency, especially for low-resource languages.

\subsection{Complete Results for Faithfulness}\seclabel{complete_faithfulness}
Table~\ref{tab:drop_first_middle_last} reports accuracy changes when truncating the first, middle, or last part of the reasoning traces. We observe that truncating the middle or the beginning generally has smaller impact, while removing the final part leads to larger performance drops, highlighting the importance of the concluding reasoning steps.

\begin{table}[]
\centering
\resizebox{\columnwidth}{!}{%
\begin{tabular}{cccc}
\hline
\textbf{Metric} & \textbf{Group}    & \textbf{Mean Value} & \textbf{P-Value}     \\ \hline
\multicolumn{4}{c}{\textbf{Without Low Resource Languages}}                      \\ \hline
\multirow{2}{*}{\begin{tabular}[c]{@{}c@{}}Trace- Substitution\\ Consistency\end{tabular}} & Indo-European & 0.7208 & \multirow{2}{*}{9.05e-03}                     \\
                & Non Indo-European & 0.7108              &                      \\ \hline
\multicolumn{4}{c}{\textbf{With Low Resource Languages}}                      \\ \hline
\multirow{2}{*}{\begin{tabular}[c]{@{}c@{}}Trace- Substitution\\ Consistency\end{tabular}} & Indo-European & 0.7108 & \multicolumn{1}{l}{\multirow{2}{*}{1.99e-52}} \\
                & Non Indo-European & 0.5452              & \multicolumn{1}{l}{} \\ \hline
\end{tabular}%
}
\caption{Consistency comparison between Indo-European and non-Indo-European languages. Reported are mean consistency values for Consistency of Trace-Substitution metrics, with corresponding p-values (t-test). We also discard low-resouce languages sw and te to conduct t-test. Indo-European languages generally achieve higher consistency, and the differences are statistically significant.}
\label{tab:p_value_swap}
\end{table}

\begin{table*}[!ht]
\centering
\small
\resizebox{\textwidth}{!}{%
\begin{tabular}{cc c c c c c c c c c c c}
\toprule
\textbf{Operation} & \textbf{Model} & \textbf{de} & \textbf{en} & \textbf{es} & \textbf{fr} & \textbf{ja} & \textbf{ru} & \textbf{sw} & \textbf{th} & \textbf{zh} & \textbf{bn} & \textbf{te} \\
\midrule
\multirow{8}{*}{\shortstack{Truncation \\ (First)}} 
&Qwen-14B     & .46 (.51) & .54 (.57) & .40 (.46) & .32 (.39) & .46 (.55) & .32 (.36) & .28 (.60) & .25 (.28) & .56 (.67) & .51 (.64) & .36 (.49) \\
&Qwen-32B     & .53 (.66) & .62 (.71) & .43 (.64) & .31 (.58) & .46 (.64) & .37 (.50) & .51 (.80) & .60 (.67) & .38 (.48) & .46 (.53) & .26 (.38) \\
&R1-Qwen-1.5B & .05 (.13) & .08 (.11) & .04 (.08) & -.00 (-.01) & .03 (.12) & .02 (.05) & -.00 (NaN) & -.02 (-2.00) & .03 (.04) & -.02 (-.29) & -.02 (-.56) \\
&R1-Qwen-7B   & .08 (.12) & .06 (.07) & .06 (.07) & .06 (.09) & .03 (.05) & .05 (.07) & -.02 (-.50) & .09 (.18) & .08 (.10) & .00 (.00) & -.12 (-.78) \\
&R1-Qwen-14B  & .08 (.10) & -.02 (-.02) & -.02 (-.02) & .06 (.07) & .03 (.04) & .02 (.02) & .00 (.02) & .04 (.05) & .02 (.02) & .04 (.06) & -.08 (-.30) \\
&R1-Qwen-32B  & .11 (.12) & .02 (.03) & .01 (.01) & -.01 (-.01) & .02 (.02) & .01 (.01) & .00 (.01) & .01 (.01) & .01 (.01) & .01 (.01) & -.36 (-1.98) \\
&R1-Llama-8B  & .10 (.20) & .13 (.16) & .11 (.15) & .16 (.26) & .04 (.09) & .04 (.06) & -.02 (-.55) & -.02 (-.05) & .04 (.05) & -.01 (-.18) & -.10 (-.92) \\
&R1-Llama-70B & .08 (.10) & .03 (.03) & .07 (.08) & .07 (.08) & .06 (.07) & .06 (.07) & .04 (.04) & .03 (.04) & .17 (.19) & -.02 (-.02) & -.32 (-.72) \\
\midrule
\multirow{8}{*}{\shortstack{Truncation \\ (Middle)}} 
&Qwen-14B     & .46 (.51) & .56 (.58) & .46 (.52) & .39 (.47) & .42 (.51) & .30 (.34) & .29 (.61) & .33 (.37) & .52 (.61) & .56 (.71) & .38 (.53) \\
&Qwen-32B     & .54 (.67) & .64 (.73) & .44 (.65) & .31 (.58) & .48 (.67) & .36 (.48) & .50 (.78) & .61 (.68) & .39 (.49) & .53 (.62) & .36 (.53) \\
&R1-Qwen-1.5B & .03 (.07) & .09 (.11) & .04 (.08) & .02 (.03) & .02 (.08) & .03 (.07) & -.00 (NaN) & -.01 (-1.50) & .01 (.02) & -.01 (-.21) & -.02 (-.56) \\
&R1-Qwen-7B   & .09 (.15) & .06 (.07) & .06 (.08) & .08 (.13) & .02 (.04) & .06 (.08) & -.01 (-.17) & .07 (.13) & .07 (.09) & -.00 (-.01) & -.12 (-.78) \\
&R1-Qwen-14B  & .08 (.11) & .00 (.00) & -.02 (-.02) & .05 (.06) & .04 (.05) & .02 (.03) & -.02 (-.06) & .04 (.04) & .02 (.02) & .04 (.06) & -.08 (-.30) \\
&R1-Qwen-32B  & .10 (.12) & .04 (.04) & .02 (.03) & -.02 (-.03) & .02 (.03) & .02 (.02) & -.01 (-.03) & .01 (.01) & .01 (.01) & .02 (.02) & -.36 (-1.93) \\
&R1-Llama-8B  & .09 (.17) & .13 (.15) & .13 (.18) & .15 (.24) & .04 (.08) & .04 (.05) & -.02 (-.36) & -.03 (-.09) & .03 (.04) & -.02 (-.24) & -.09 (-.85) \\
&R1-Llama-70B & .10 (.12) & .02 (.02) & .09 (.10) & .08 (.10) & .06 (.07) & .07 (.08) & .04 (.05) & .03 (.04) & .13 (.15) & -.04 (-.06) & -.33 (-.73) \\
\midrule
\multirow{8}{*}{\shortstack{Truncation \\ (Last)}} 
&Qwen-14B     & .68 (.75) & .75 (.79) & .69 (.79) & .64 (.77) & .69 (.82) & .68 (.75) & .40 (.84) & .65 (.72) & .67 (.79) & .66 (.83) & .63 (.87) \\
&Qwen-32B     & .66 (.83) & .76 (.86) & .55 (.81) & .39 (.72) & .62 (.87) & .60 (.81) & .56 (.89) & .77 (.86) & .60 (.75) & .77 (.89) & .60 (.88) \\
&R1-Qwen-1.5B & .22 (.55) & .51 (.65) & .31 (.67) & .23 (.50) & .14 (.56) & .28 (.64) & -.01 (NaN) & -.02 (-2.50) & .26 (.40) & .01 (.14) & -.01 (-.22) \\
&R1-Qwen-7B   & .28 (.45) & .43 (.51) & .41 (.54) & .32 (.53) & .22 (.42) & .36 (.50) & .03 (.67) & .22 (.43) & .29 (.36) & .06 (.12) & .00 (.00) \\
&R1-Qwen-14B  & .40 (.52) & .32 (.38) & .29 (.36) & .35 (.44) & .24 (.31) & .39 (.44) & .12 (.48) & .30 (.37) & .24 (.27) & .24 (.37) & .06 (.20) \\
&R1-Qwen-32B  & .30 (.35) & .26 (.27) & .35 (.40) & .21 (.26) & .30 (.34) & .29 (.32) & .20 (.43) & .22 (.25) & .14 (.16) & .20 (.25) & -.22 (-1.22) \\
&R1-Llama-8B  & .38 (.71) & .59 (.70) & .54 (.77) & .45 (.71) & .28 (.61) & .48 (.72) & .01 (.18) & .16 (.43) & .44 (.60) & .01 (.18) & .04 (.38) \\
&R1-Llama-70B & .36 (.44) & .22 (.23) & .38 (.42) & .35 (.41) & .28 (.34) & .39 (.43) & .32 (.37) & .26 (.31) & .29 (.33) & .11 (.15) & -.09 (-.19) \\

\bottomrule
\end{tabular}
} 
\caption{Absolute drop and relative drop rate compared to baseline (First, Middle and Last truncation).}
\label{tab:drop_first_middle_last}
\end{table*}

\section{Details of Datasets}\seclabel{appendix_datasets}

\subsection{Language Coverage}

Table~\ref{tab:language_coverage} summarizes the language coverage of the two datasets used in our experiments: MMMLU and MGSM.

\begin{table}
\footnotesize
  \centering
    \begin{tabular}{p{1.5cm}p{5cm}} 
    \midrule
    \textbf{Dataset} & \textbf{Languages} \\
    \midrule
    MMMLU & Arabic, Bengali, German, Spanish, French, Hindi, Indonesian, Italian, Japanese, Korean, Portuguese, Swahili, Yoruba, Chinese \\
    MGSM & Bengali, Chinese, French, German, Japanese, Russian, Spanish, Swahili, Telugu, Thai \\
    \midrule
    \end{tabular}
  \caption{Languages covered in MMMLU and MGSM.}
  \label{tab:language_coverage}
\end{table}

\subsection{Test Instances}

To reduce computational costs, we limit each dataset to a maximum of 250 test instances per language.
For MMMLU, we randomly sample 250 examples from the full 14K test set.
This sampling is applied consistently across all parallel language versions to ensure comparability.
For MGSM, we use the official test set, which consists of 250 parallel examples available across all supported languages.

\subsection{Prompt Templates}\seclabel{prompt_templates}

\begin{table}[!htbp]
\centering
\includegraphics[width=\linewidth]{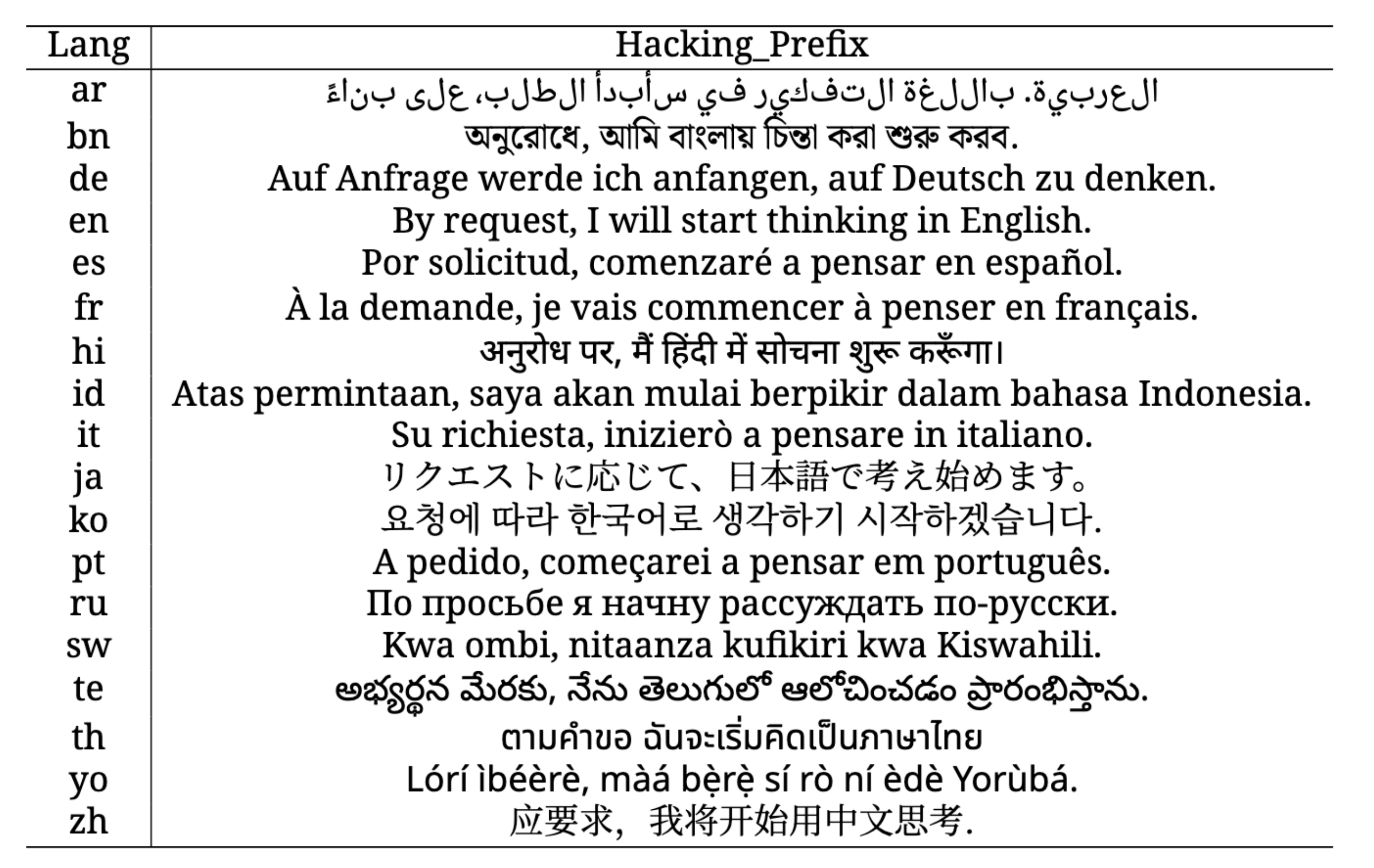}
\caption{Language-specific answer instructions.}
\label{tab:prompt_hacking}
\end{table}

\begin{figure*}[htbp]
\centering
\includegraphics[width=\textwidth]{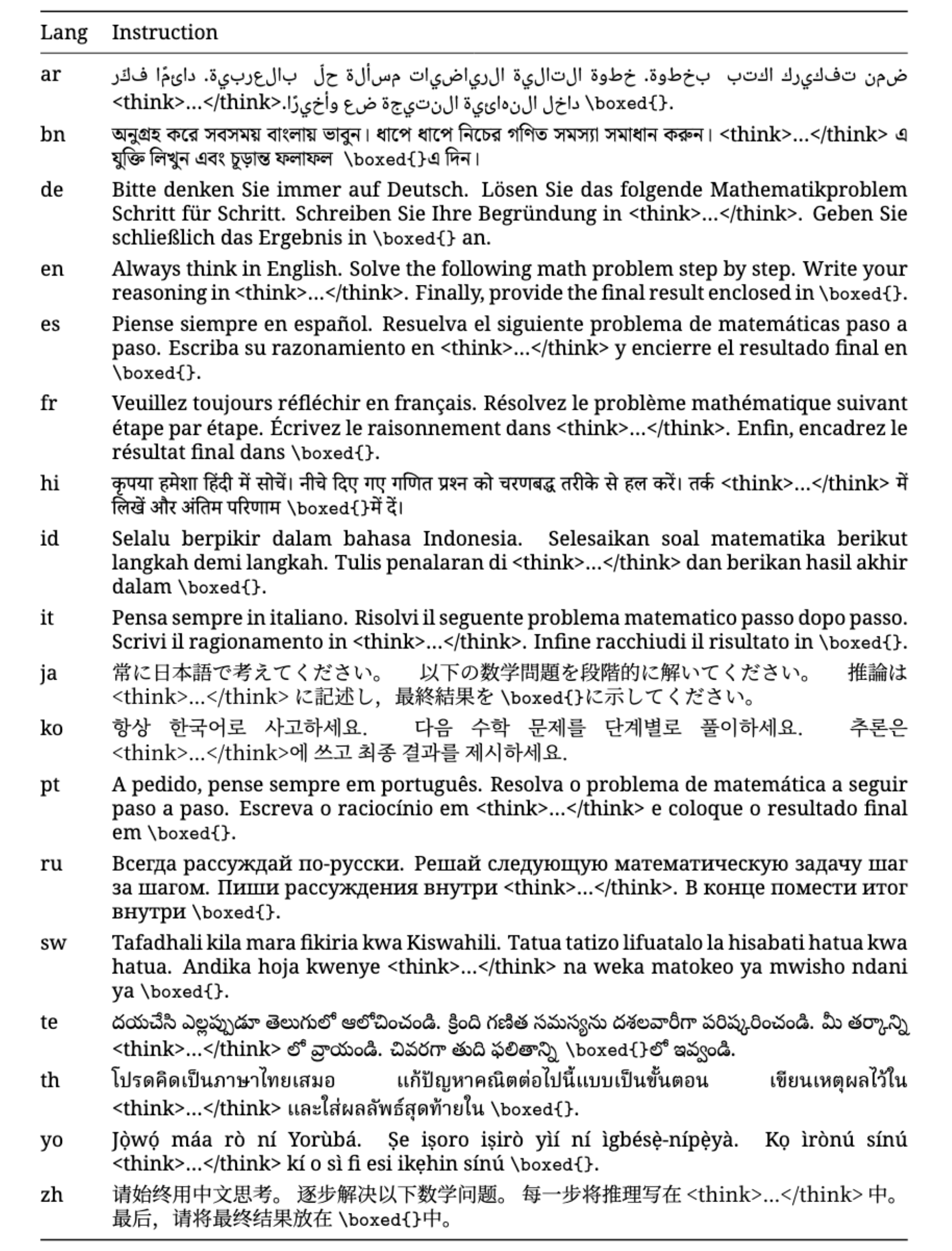}
\caption{Prompts for \textbf{MGSM} task in different languages.}
\label{tab:prompts}
\end{figure*}

Table~\ref{tab:prompts} provides an overview of the prompts designed for model instruction. However, we observed that, even when explicit prompts were provided, the model frequently conducted its intermediate reasoning in a language different from that of the input prompt. To mitigate this inconsistency, we appended the language-specific instructions listed in Table~\ref{tab:prompt_hacking} after the query and the \texttt{<think>} tag. This strategy effectively enforces the model to align its reasoning language with the prompt language. To avoid redundancy, we present prompt templates for the MGSM task in Table~\ref{tab:prompts} only, while templates for MMMLU share the same structural format and are available in the released code.

\section{Experimental Environment and Hyperparameters}
\seclabel{environment}

We set the maximum generation length to 8192 tokens for all models. 
We use the recommended configurations provided on \href{https://huggingface.co}{HuggingFace} for all models.
Specifically, we set the temperature to 0.6 and top-$p$ to 0.95 for distilled versions of \texttt{DeepSeek R1}.
For Qwen3 models, we set the temperature to 0.6, top-$p$ to 0.95, and top-$k$ to 20.

Experiments are primarily conducted on NVIDIA A100 GPUs.  
For larger models, such as \texttt{DeepSeek-R1-Distill-Llama-70B}, we use NVIDIA H200 GPUs for inference.

To evaluate final-answer correctness, we adopt an exact matching strategy.  
Following prior work~\citep{qi2025modelsreasonlanguagecontrolling}, we prompt the model to wrap its final answer in \texttt{\textbackslash boxed\{\}}, and extract the boxed content for comparison against the gold answer.

\end{document}